\documentclass[letterpaper,twoside]{article}
\usepackage{preprint}
\usepackage[round]{natbib}

\usepackage{amsmath,amssymb}
\usepackage{graphicx}
\usepackage{booktabs,longtable,multirow}
\usepackage{balance}
\usepackage{xurl}
\usepackage[hidelinks]{hyperref}
\hypersetup{
  pdftitle={Learning What to Share: Expert Specialization for Time-Series Forecasting},
  pdfauthor={Bashar Khoury; Shadi Omari; Assaf Schuster},
  pdfsubject={Expert specialization for time-series forecasting}
}
\usepackage{xcolor}
\usepackage{needspace,float}
\AddToHook{env/longtable/before}{\Needspace{12\baselineskip}}

\begin{document}
\twocolumn[{%
\preprinttitle{Learning What to Share: Expert Specialization for Time-Series Forecasting}
\begin{center}
{\large\bfseries Bashar Khoury\textsuperscript{*}\qquad
Shadi Omari\textsuperscript{*}\qquad Assaf Schuster}\par
\vspace{0.6em}
Department of Computer Science\\
Technion -- Israel Institute of Technology\\[0.3em]
{\small\{\texttt{bashar.kh,shadi-omari}\}\texttt{@cs.technion.ac.il},
\texttt{assaf@technion.ac.il}}
\end{center}
\vspace{0.5em}
}]
\thispagestyle{plain}
\begingroup
\renewcommand{\thefootnote}{\fnsymbol{footnote}}
\footnotetext[1]{These authors contributed equally.}
\endgroup

\begin{abstract}
A central challenge in multivariate forecasting is deciding what to share across channels and where to specialize. Mixture-of-experts models offer flexible sharing, yet it remains unclear how much of their forecasting gains comes from added capacity or expert allocation. Through controlled comparisons across eleven datasets and three backbones, we show that informative channel assignments improve forecasting beyond matched shared capacity and random allocation. On frozen transformer backbones, persistent channel-specific mixtures outperform input-dependent routing on average. To examine when and how much specialization helps, we introduce \emph{GalaxyTS}, a controlled synthetic dataset with exact shared and channel-specific forecasting references. It reveals how persistent channel mixtures exploit predictive differences concealed by short input windows and measures gains over matched shared corrections. These insights motivate \emph{Persistent Closed-Form Routing (PCR)}, a plug-and-play method that alternates exact conditional estimation of channel-specific mixture weights with gradient-based training of low-rank residual experts around a frozen pretrained forecaster. Averaged over four forecasting horizons, PCR-K4 achieves dataset-balanced mean squared error reductions of 3.86-14.25\% relative to the three frozen backbones. In matched comparisons, PCR outperforms competing specialization methods on average while requiring substantially fewer trainable parameters.
\end{abstract}

\section{Introduction}
\label{sec:introduction}

Effective multivariate forecasting requires sharing information across related time series while accommodating different predictive relationships \citep{monteromanso2021global,shao2022stid}. Traffic sensors and electricity meters may exhibit similar cycles yet follow different forecasting rules. Mixture-of-experts (MoE) architectures enable specialization by allowing channels to combine shared experts differently \citep{ni2024mole,chen2024ccm,wang2024ve}. This raises a central question: \emph{when does specialization improve forecasting, and how can effective sharing be learned efficiently?}

Gains over a shared predictor can reflect added capacity, informative allocation, or optimization differences. Routing can exploit channel identity \citep{wang2024ve}, input features \citep{shazeer2017sparselygated,shi2025timemoe}, or both. Controlled comparisons must therefore assess allocation quality, routing information, and the accuracy and cost of backbone adaptation.

We investigate which sharing decisions matter across eleven datasets using DLinear, PatchTST, and iTransformer \citep{zeng2023transformers,nie2023patchtst,liu2024itransformer}, matching expert budgets and backbone checkpoints in core comparisons. Informative assignments outperform shared corrections and random allocation on average. On the frozen transformers, persistent channel mixtures, whose weights remain fixed across input windows, outperform mixtures conditioned on identity and input on average; DLinear favors input conditioning. Joint adaptation improves accuracy at higher observed transformer fitting cost. These comparisons leave open which predictive differences make specialization useful and how much can be gained.

To investigate which sharing patterns exploit stable predictive differences, we introduce a synthetic benchmark dataset, \emph{GalaxyTS} (Figure~\ref{fig:synthetic-galaxy}). It varies forecasting relationships within and across overlapping families and provides exact errors for optimal shared and channel-specific predictors using one channel's history. Their difference quantifies the cost of imposing one common forecasting rule. In matched correction experiments, persistent mixtures and ridge-based grouping both outperform random and input-dependent routing; persistent mixtures achieve larger gains over Global correction. This leaves a practical question: how can effective persistent mixtures be fitted efficiently while keeping the backbone frozen?

\begin{figure*}[t]
\centering
\includegraphics[width=\textwidth]{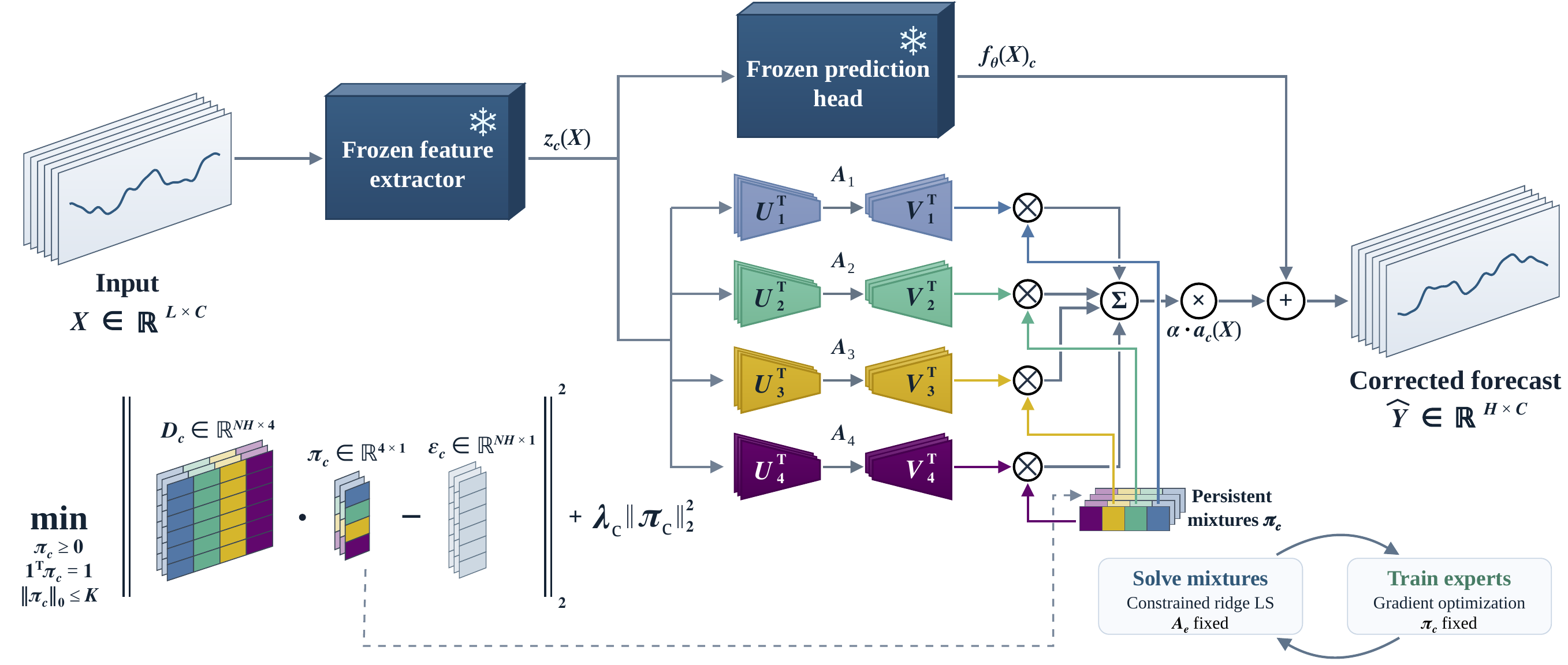}
\caption{\textbf{Persistent Closed-Form Routing (PCR).} Channel-specific mixtures combine low-rank experts to correct a frozen forecast. Training alternates constrained ridge allocation with expert updates, preserving the backbone throughout.}
\label{fig:pcr-architecture}
\end{figure*}

We propose \emph{Persistent Closed-Form Routing} (PCR; Figure~\ref{fig:pcr-architecture}), a plug-and-play residual method for specializing pretrained forecasters. Building on low-rank adaptation and residual correction \citep{hu2022lora,liang2026deltaadapter}, PCR alternates exact closed-form updates of channel mixture weights for fixed experts with gradient-based expert training, keeping the backbone frozen. PCR-K4, which permits all four experts per channel, reduces dataset-balanced backbone MSE by 3.86-14.25\% across the three backbones, averaged over four horizons. It outperforms expert-budget-matched variants of MoLE \citep{ni2024mole}, CCM \citep{chen2024ccm}, and VE \citep{wang2024ve} on average, with fewer trainable parameters. PCR achieves comparable accuracy to jointly learned mixtures with shorter observed transformer fitting times, producing forecasts with lower median single-window latency than the evaluated expert-based competitors.  

\textbf{Contributions.}\footnote{Training and evaluation code is not currently publicly available; a public release is planned.}
\begin{enumerate}
\setlength{\itemsep}{0pt}
\setlength{\parskip}{0pt}
\item \textbf{Mechanistic analysis:} comparisons controlling expert capacity and examining allocation, routing information, and backbone adaptation show that informative sharing improves average forecasting accuracy beyond matched shared capacity.
\item \textbf{GalaxyTS:} a reproducible synthetic benchmark with graded predictive differences, exact forecasting references, and capacity-matched tests of informative channel allocation.
\item \textbf{PCR:} plug-and-play specialization of frozen forecasters via exact conditional mixture optimization, achieving higher average accuracy with fewer trainable parameters and lower median inference latency than expert-budget-matched methods.
\end{enumerate}
\label{intro-end}

\section{Preliminaries}
\label{sec:preliminaries}

\paragraph{Forecasting problem.}
Let $x_t\in\mathbb{R}^{C}$ collect the observations of $C$ channels
at time $t$. An input window $X_t\in\mathbb{R}^{L\times C}$ contains
observations from $t-L+1$ through $t$; multivariate forecasting predicts
the next $H$ observations, $Y_t\in\mathbb{R}^{H\times C}$.
We write $\widehat Y_t=f_\theta(X_t)$, suppressing backbone-specific
calendar covariates. A channel index identifies the same variable or
sensor across windows within a dataset. We omit the time index when
unnecessary and treat channel histories, forecasts, and features as
column vectors.

\paragraph{Datasets and backbones.}
Our real-data framework comprises eleven datasets: ETTh1, ETTh2, ETTm1,
and ETTm2 \citep{zhou2021informer}; Electricity
\citep{trindade2015electricity}; Traffic \citep{caltrans_pems};
Weather \citep{mpibgc_weather}; and PEMS03, PEMS04, PEMS07, and PEMS08
\citep{guo2022astgnn}. Our backbones are DLinear \citep{zeng2023transformers},
PatchTST \citep{nie2023patchtst}, and iTransformer
\citep{liu2024itransformer}, representing linear forecasting,
channel-independent temporal attention, and cross-channel attention,
respectively. This selection allows us to assess specialization
across distinct model classes and treatments of channel dependence.
All three backbones use Time Series Library (TSLib) implementations
\citep{wang2026tslib}.
Dataset characteristics and preprocessing appear in
Appendix~\ref{app:datasets}; backbone configurations appear in
Table~\ref{tab:backbone-recipes}. 

\paragraph{Correction framework.}
We study specialization and routing through low-rank residual experts
added to a common frozen forecaster
\citep{wu2022residualmoe,hu2022lora,liang2026deltaadapter}.
Fixing the backbone's predictions and representations separates
allocation effects from backbone adaptation, while matching expert
parameter budgets controls for added capacity.
For channel $c$, the corrected forecast is
\begin{equation}
 \begin{aligned}
 \widehat y_c(X)&=f_\theta(X)_c+\alpha a_c(X)r_c(X),\\
 r_c(X)&=\sum_{e=1}^{E}\pi_e(c,z_c(X))A_ez_c(X),
 \end{aligned}
 \label{eq:corrected-predictor}
\end{equation}
where $z_c(X)\in\mathbb{R}^{F}$ is the correction input, taken from the
backbone's prediction-head features in the real-data experiments.
Each expert $A_e\in\mathbb{R}^{H\times F}$ has rank at most $r$;
the allocation weights $\pi_e$ are nonnegative and sum to one.
The scalar $\alpha$ controls correction strength, and $a_c(X)$ reverses
window-normalization scaling ($a_c=1$ for DLinear).
Appendix~\ref{app:correction-framework} provides the full implementation details.

\paragraph{Evaluation setting.}
We use chronological splits and training-only channel standardization,
and evaluate mean squared error (MSE) and mean absolute error (MAE)
in standardized observation coordinates.
Primary results report means and sample standard deviations over three random seeds. Appendix~\ref{app:datasets} details preprocessing, splits,
and aggregation.

\begin{table}[t]
\centering
\small
\setlength{\tabcolsep}{3pt}
\caption{\textbf{Meaningful allocation beyond shared capacity.}
MSE reduction over the pretrained backbone (\%), averaged over 22 settings.
Global uses rank 64; routed variants use four rank-16 experts, two active.
Bold/underline indicate best/second-best means.}
\label{tab:allocation-gains}
\resizebox{\columnwidth}{!}{%
\begin{tabular}{@{}lccc@{}}
\toprule
Variant & DLinear & PatchTST & iTransformer \\
\midrule
Global correction
& $0.96\pm0.30$ & $5.79\pm0.05$ & $0.60\pm0.05$ \\
Random routing
& $0.85\pm0.11$ & $6.27\pm0.15$ & $1.18\pm0.11$ \\
Ridge-based
& $2.32\pm0.18$ & $7.95\pm0.06$ & $2.24\pm0.04$ \\
Persistent ID
& $3.57\pm0.20$ & $\mathbf{12.41}\pm0.29$ & $\mathbf{7.62}\pm0.10$ \\
Input-dependent
& $\underline{10.46}\pm0.45$ & $6.46\pm0.10$ & $0.49\pm0.03$ \\
Identity + input
& $\mathbf{11.49}\pm0.22$ & $\underline{10.36}\pm0.13$ & $\underline{7.20}\pm0.19$ \\
\bottomrule
\end{tabular}%
}
\end{table}

\section{What Drives Specialization Gains?}
\label{sec:routing}
\label{sec:success}

This section examines whether meaningful channel allocation improves
on shared capacity, whether routing benefits from the current input
window, and whether further accuracy gains justify adapting the backbone.

\paragraph{Experimental setting.}
\label{sec:setup}
The experiments follow the frozen protocol of
Section~\ref{sec:preliminaries}, using the three backbones with
$L=96$ and $H\in\{96,720\}$.
Expert banks contain four rank-16 experts, with two of them active per channel
in the primary comparisons. Validation selects hyperparameters,
checkpoints, and a global correction scale $\alpha$; training and
aggregation details appear in Appendices~\ref{app:datasets}
and~\ref{app:allocation-training}.

\begin{figure*}[t]
\centering
\includegraphics[width=\textwidth]{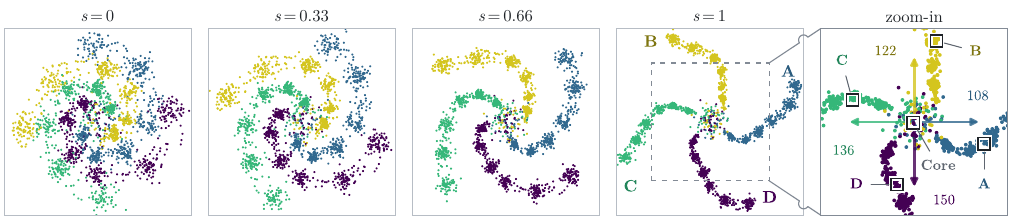}
\caption{\textbf{Channel layout of GalaxyTS.} Dots sample the layout, colored by family; direction sets the echo mix. Larger $s$ unwinds the arms around a shared core. Boxes mark the channels of Figure~\ref{fig:toy-hidden}.}
\label{fig:synthetic-galaxy}
\end{figure*}

\subsection{Does Meaningful Allocation Help beyond Capacity?}
\label{sec:capacity}
\label{sec:method-definition}

We examine four sharing and allocation strategies.
\emph{Global correction} applies one residual map to every channel,
matching the expert bank's parameter budget. \emph{Random routing}
assigns fixed, equally weighted expert pairs.
\emph{Persistent ID routing} learns channel-specific mixture weights,
reusing them across input windows.
We additionally introduce \emph{Ridge-based routing} to test whether
predictive grouping improves on random allocation without learning
routing weights. Channels are grouped by training-fitted ridge maps
that describe how their histories predict their futures. Each channel
equally combines the experts for its own group and its nearest
alternative (Appendix~\ref{app:allocation-grouping}).

\textbf{Informative allocation improves on matched shared capacity.}
Averaged over the eleven datasets and both horizons, Persistent ID
routing reduces MSE relative to Global correction by $2.48\%$,
$7.41\%$, and $7.07\%$ on DLinear, PatchTST, and iTransformer,
respectively. Relative to Random routing, it improves on nine of
eleven datasets for DLinear (mean reduction $2.74\%$) and ten of
eleven for both PatchTST ($6.96\%$) and iTransformer ($6.53\%$).
Thus, channel-specific sharing provides benefits beyond extra capacity
or arbitrary allocation. These comparisons match expert budgets,
while retaining each variant's optimization rule
(Appendix~\ref{app:allocation-arms}).

Predictive grouping improves average accuracy even without learned
routing weights. With matched expert budgets and equally weighted
expert pairs, Ridge-based routing reduces MSE relative to Random
routing by $1.47\%$, $1.91\%$, and $1.08\%$ on DLinear,
PatchTST, and iTransformer, respectively, averaged over 22 settings
per backbone. Persistent ID routing remains stronger on average
in these matched comparisons.

\label{sec:comparison-variants}
The benefit also persists with all four experts active: Dense ID
achieves $3.76\%$, $13.40\%$, and $9.79\%$ mean reductions.
Dense, global-mixture, and fixed-group controls are detailed in
Appendix~\ref{app:allocation-results}; sparsity is not a prerequisite
for effective persistent allocation.

\begin{figure*}[t]
\centering
\includegraphics[width=\textwidth]{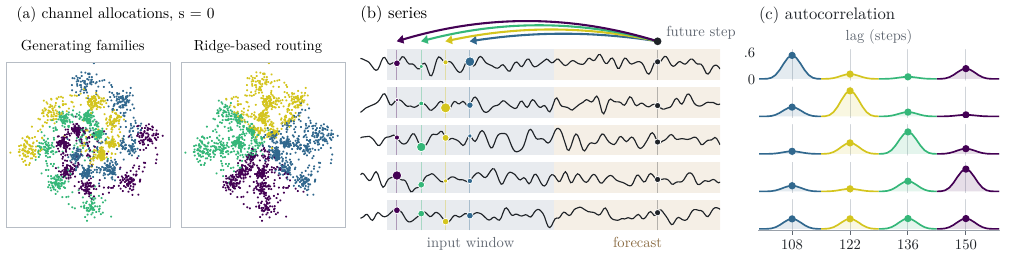}
\caption{\textbf{Predictive grouping and echo dependence.} (a) Generating families and Ridge groups at $s=0$; (b) boxed channels at $s=1$ with echo links; (c) their exact autocorrelation near the echo delays (same scale).}
\label{fig:toy-hidden}
\end{figure*}

\begin{figure}[t]
\raggedright
\includegraphics[width=0.85\columnwidth]{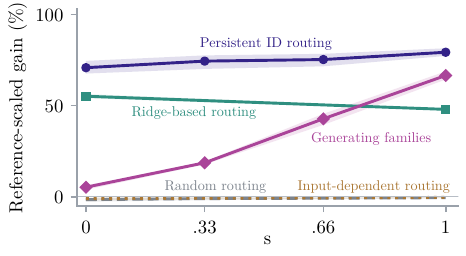}
\caption{\textbf{Reference-scaled gains over Global.} $H=96$. Curves: approximate mean $G$ values; shaded bands: range across backbones. See Equation~\ref{eq:synthetic-gap}.}
\label{fig:synthetic-allocation-gap}
\end{figure}

\subsection{What Does Input Conditioning Add?}
\label{sec:routing-results}

\emph{Input-dependent routing} uses features of the current input window
without explicit channel identity; \emph{Identity + input routing}
combines both. Comparing the latter with Persistent ID routing tests
the value of input information, while comparison with Input-dependent
routing tests the value of explicit channel identity.
For each backbone, all three variants use the same frozen forecaster
and expert bank architecture, with two active experts per channel.
Corrections are fitted separately for each variant.

\textbf{Persistent allocation is effective on PatchTST and iTransformer.}
With their pretrained weights fixed, Persistent ID routing reduces MSE
relative to Identity + input routing by $2.51\%$ and $0.36\%$,
respectively, winning on $9/11$ and $6/11$ dataset averages.
Against Input-dependent routing, its iTransformer MSE reduction
is $7.19\%$, with wins in $21/22$ settings
(Appendix~\ref{app:tslib-routing}).
Table~\ref{tab:allocation-gains} shows a different pattern for DLinear:
Identity + input routing reduces backbone MSE by $11.49\%$,
versus $3.57\%$ for Persistent ID routing.
Input-conditioned mixtures add nonlinearity to its linear correction
framework, so this contrast need not imply a general benefit from
changing assignments across input windows.

\subsection{What Does Backbone Adaptation Add?}
\label{sec:freezing-full}

\textbf{Backbone adaptation introduces an accuracy-resource trade-off.} We compare Persistent ID routing with a frozen backbone against
joint fitting of the pretrained backbone and freshly initialized
experts. Joint fitting reduces MSE relative to freezing by
$0.72\%$, $1.71\%$, and $3.83\%$ on DLinear, PatchTST, and
iTransformer, respectively (Table~\ref{tab:adaptation-main}).

On a separately profiled four-dataset panel, joint fitting requires
$2.53$-$2.75\times$ the fitting time for the transformers and
$2.85$-$7.35\times$ the peak training memory of freezing
(Appendix~\ref{app:adaptation-resources}).
DLinear instead fits faster jointly: for its lightweight linear
backbone, pipeline overhead may outweigh the computation saved
by freezing.
The supporting controls show that jointly adapted Persistent ID
also outperforms joint Global correction and backbone fine-tuning
alone on every backbone average.

These findings motivate a controlled test of whether stable
differences in forecasting rules make informative channel allocation
useful. A synthetic setting with known forecasting rules can quantify
the cost of a common predictor and test which allocation strategies
exploit these differences under matched capacity.

\section{When Should Specialization Help?}
\label{sec:synthetic}

When does learning what to share improve forecasting? Real benchmarks offer no systematic control over differences and overlap in channels' forecasting rules, nor an exact reference for the attainable benefit of specialization. We introduce \emph{GalaxyTS}, a synthetic dataset with graded predictive heterogeneity and overlapping families, for which the optimal shared and channel-specific forecasting errors are known exactly. Their difference characterizes the benefit of specialization from single-channel histories; the matched correction experiments below test which allocation strategies improve on a fitted Global correction.

\subsection{Synthetic Dataset Design}
\label{sec:synthetic-design}

\paragraph{Design principles.}
(i)~\emph{Forecasting rules.} Channels differ in how their past predicts their future despite sharing local statistics, so useful grouping must follow forecasting behavior.
(ii)~\emph{Overlapping families.} Each channel has a generating-family label, A, B, C, or D. These tags record its origin, not its optimal expert assignment, and remain fixed as $s$ varies. Within-family variation and a shared core allow overlap.
(iii)~\emph{Graded separation.} A parameter $s\in[0,1]$ increases family separation and strengthens the temporal dependencies.
(iv)~\emph{Exact references.} For predictors using only one channel's input history, the Gaussian construction and known covariances yield exact optimal shared and channel-specific expected MSEs. Their difference, the \emph{specialization gap}, quantifies the cost of imposing one common forecasting rule in this reference class.

\paragraph{Generation.}
Each channel is a weighted sum of four \emph{echo} components. At each step, a component adds fresh, smoothly varying noise to a damped copy of its own value a fixed number of steps earlier, its \emph{delay}; components differ only in this delay (Appendix~\ref{app:synthetic-generation}). A channel's weights, its \emph{echo mix}, determine which past values dominate its future and hence its forecasting rule. Its position in a two-dimensional layout determines this mix (\hyperref[fig:synthetic-galaxy]{Figure~\ref*{fig:synthetic-galaxy}}): direction selects the dominant echo, four spiral arms form the families, and the shared core holds balanced mixes. Visually, increasing $s$ unwinds the arms and strengthens the echoes. The parameter $s$ changes both family separation and echo strength; the sweep therefore characterizes their joint variation rather than isolating either factor.

\hyperref[fig:toy-hidden]{Figure~\ref*{fig:toy-hidden}b} links an illustrated future step to its echoed inputs. \hyperref[fig:toy-hidden]{Figure~\ref*{fig:toy-hidden}c} shows autocorrelation near the echo delays, with peak heights reflecting the corresponding component weights. Each GalaxyTS realization has 512 channels and 12,000 steps; we generate three independent realizations at each $s\in\{0,0.33,0.66,1\}$. Appendix~\ref{app:synthetic-generation} specifies the generator and shows the full autocorrelation profiles; Appendix~\ref{app:synthetic-references} explains the exact reference errors.

\subsection{Which Allocations Realize the Gain?}
\label{sec:synthetic-results}
\label{sec:synthetic-residuals}

We compare rank-16 expert corrections using features from frozen DLinear, PatchTST, and iTransformer, with parameter-matched Global corrections. Here we report four separation levels and $H\in\{96,720\}$. Random routing and Ridge-based routing are evaluated at $s\in\{0,1\}$ with two active experts per channel, matching Persistent ID routing and Input-dependent routing (Section~\ref{sec:routing}). Each setting uses three training seeds within each of three realizations (Appendix~\ref{app:synthetic-protocol}).

At $s=0$, echo mixes can differ within generating families, while forecasting rules can be similar across families. Ridge-based routing clusters estimated forecasting maps, not family labels, yielding cross-family allocations (\hyperref[fig:toy-hidden]{Figure~\ref*{fig:toy-hidden}a}; raw-window allocation visualization in Appendix~\ref{app:numerical-synthetic}).

\paragraph{Predictive differences create an opportunity for specialization.}
The \emph{specialization gap} is the reduction in expected MSE available by replacing one optimal shared predictor with optimal channel-specific predictors, using the same input window and horizon. As $s$ increases from 0 to 1, the optimal channel-specific error falls more than the optimal shared error. Evaluating these exact predictors on the same test windows, the corresponding MSE gap grows from 2.8\% to 14.2\% of optimal shared MSE at $H=96$, and from 0.7\% to 9.6\% at $H=720$. Thus predictive heterogeneity creates measurable room for specialization even when channels share local statistics (Appendix~\ref{app:synthetic-references}).

\paragraph{Persistent channel information improves matched corrections.}
We report the MSE improvement over the matched Global correction,
expressed as a percentage of the exact-reference gap ($G$;
Appendix~\ref{app:synthetic-protocol}). Persistent ID routing
achieves larger gains than matched two-expert Ridge-based routing
at both separation endpoints and horizons. At $s=1$, its
reference-scaled gains are 79.2\% at $H=96$ and 64.0\% at $H=720$,
versus 47.9\% and 42.5\% for Ridge-based routing.
Figure~\ref{fig:synthetic-allocation-gap} summarizes the $H=96$
comparison. Predictive heterogeneity and useful specialization persist
at $s=0$, despite overlapping families. Input-dependent routing and
adding input conditioning provide no consistent benefit
(Appendix~\ref{app:synthetic-native}).

\paragraph{Lessons for specialization.}
GalaxyTS illustrates how persistent channel mixtures exploit predictive differences that local input windows do not reveal. Persistent ID routing achieves larger MSE reductions over matched Global corrections than matched two-expert Ridge-based routing, while input conditioning provides no consistent benefit in this setting. These gains motivate learning persistent channel-specific sharing around a frozen forecaster.

\section{Persistent Specialization via Closed-Form Routing}
\label{sec:method}

The preceding experiments motivate a compact shared expert bank with
persistent channel-specific mixtures. We propose \emph{Persistent
Closed-Form Routing} (PCR), a plug-and-play residual-specialization
method for pretrained forecasters (Figure~\ref{fig:pcr-architecture}). PCR preserves the backbone and
exploits the conditional quadratic structure of the mixture weights
to estimate channel allocations analytically.

\subsection{Persistent Residual Specialization}
\label{sec:correction-fitting}

We specialize Equation~\eqref{eq:corrected-predictor} to a persistent
channel vector $\pi_c$, obtaining
\begin{equation}
 \widehat y_c(X)=f_\theta(X)_c+
 \alpha a_c(X)\sum_{e=1}^{E}\pi_{ce}A_ez_c(X).
 \label{eq:frozen-forecast}
\end{equation}
Each $A_e=V_e^\top U_e^\top$ is a shared rank-$r$ correction, the mixing weights are nonnegative and sum to one.
The weights are constant across windows; the expert outputs respond
to the current features. We use $E=4$, $r=16$, and either permit all
experts ($K=4$) or impose $\|\pi_c\|_0\leq2$ ($K=2$).
PCR alternates exact conditional gate estimation with expert training.
The expert output factors are initialized to zero, so PCR initially
preserves the backbone's predictions. Section~\ref{sec:method-cost}
compares PCR with learned allocation controls to assess the
accuracy-cost trade-off.

\subsection{Persistent Closed-Form Routing}
\label{sec:pcr-solver}

For a fixed expert bank, each column of $D_c$ stacks one expert's
correction outputs, scaled by $a_c(X)$, across $N$ training windows
and $H$ forecast steps. The vector $\varepsilon_c$ stacks the
corresponding differences between the ground-truth targets and the
frozen backbone's predictions. With the overall correction scale set to one,
PCR estimates the mixture weights by
\begin{equation}
 \min_{\substack{\pi_c\geq0,\;\mathbf1^\top\pi_c=1\\
                   \|\pi_c\|_0\leq K}}
 \|\varepsilon_c-D_c\pi_c\|_2^2+\lambda_c\|\pi_c\|_2^2.
 \label{eq:pcr-objective}
\end{equation}
Defining $G_c=D_c^\top D_c$ and
$b_c=D_c^\top\varepsilon_c$, the objective can be written,
up to the constant $\|\varepsilon_c\|_2^2$, as
\begin{equation}
 \pi_c^\top(G_c+\lambda_c I)\pi_c-2b_c^\top\pi_c.
 \label{eq:pcr-quadratic}
\end{equation}
Thus, gate estimation depends only on an $E\times E$ Gram
matrix and an $E$-dimensional vector, which can be accumulated
without storing the full matrix $D_c$.

Consider a candidate subset $S$ of at most $K$ experts, with all
other expert weights set to zero. Let $G_{c,SS}$ contain the rows
and columns of $G_c$ corresponding to these experts, and let
$b_{c,S}$ contain the corresponding entries of $b_c$.
Defining $Q_S=G_{c,SS}+\lambda_c I$, the solution subject to the equality constraints is
\begin{equation}
 \begin{aligned}
 \pi_{c,S}&=Q_S^{-1}(b_{c,S}-\tau\mathbf1),\\
 \tau&=\frac{\mathbf1^\top Q_S^{-1}b_{c,S}-1}
                  {\mathbf1^\top Q_S^{-1}\mathbf1}.
 \end{aligned}
 \label{eq:pcr-solution}
\end{equation}
PCR selects the feasible candidate with the lowest regularized
training loss across all admissible expert subsets, yielding the
exact mixture solution for a fixed bank. After two initial epochs
with uniform mixtures, training alternates between exact mixture
estimation and two-epoch expert updates with the mixtures held fixed. Validation selects
the ridge multiplier, scale $\alpha$, and checkpoint. Exact feature
and moment caches can avoid repeated backbone evaluation.
Appendix~\ref{app:pcr-details} gives optimization and selection details.

\subsection{Specializing Forecasters on GalaxyTS}
\label{sec:pcr-synthetic}

We evaluate PCR-$K2$ on GalaxyTS using four rank-16 experts
with at most two active per channel. At $s=1$, its reference-scaled
gains over its matched Global correction are 82.4\% and 65.9\%
at $H=96$ and $H=720$, respectively, averaged across backbones
(Table~\ref{tab:synthetic-native-main}).
Table~\ref{tab:synthetic-final-complete} reports absolute MSEs
alongside both exact references.

Shuffling fitted mixture vectors across channels increases test MSE
in every tested permutation while keeping the experts, backbone,
and correction scale fixed
(Appendix~\ref{app:synthetic-perm}). This supports the functional
importance of the fitted channel assignments.

\begin{table}[t]
\centering\small
\setlength{\tabcolsep}{1.5pt}
\caption{\textbf{Reference-scaled gains over Global ($K=2$).}
Mean $G$ (\%) over backbones,
realizations, and training seeds.}
\label{tab:synthetic-native-main}
\begin{tabular}{@{}cccccc@{\hspace{3pt}{\vrule width 0.3pt}\hspace{3pt}}c@{}}
\toprule
& & \multicolumn{4}{c@{\hspace{3pt}{\vrule width 0.3pt}\hspace{3pt}}}{{\scriptsize Allocation study}} &
{\scriptsize PCR evaluation}\\
\cmidrule(lr){3-6}\cmidrule(l){7-7}
\shortstack{\strut $H$\\\strut} &
\shortstack{\strut $s$\\\strut} &
\shortstack{\strut Random\\\strut} &
\shortstack{\strut Ridge\\\strut} &
\shortstack{\strut Input-\\\strut dependent} &
\shortstack{\strut Persistent\\\strut ID} &
\shortstack{\strut PCR-$K2$\\\strut}\\
\midrule
\multirow{2}{*}{$96$}
& $0$ & $-1.0$ & $55.1$ & $-1.1$ & $70.8$ & $75.7$\\
& $1$ & $-0.1$ & $47.9$ & $-0.3$ & $79.2$ & $82.4$\\
\midrule[0.3pt]
\multirow{2}{*}{$720$}
& $0$ & $-1.4$ & $50.5$ & $-0.9$ & $58.0$ & $61.8$\\
& $1$ & $-0.4$ & $42.5$ & $-0.3$ & $64.0$ & $65.9$\\
\bottomrule
\end{tabular}
\par\smallskip
\parbox{\columnwidth}{\footnotesize{Gains use each study's matched Global control; protocols in Appendices~\ref{app:synthetic-protocol} and~\ref{app:pcr-synthetic}.}}
\end{table}

\subsection{Forecasting Accuracy across Backbones}
\label{sec:method-performance}

We evaluate the full eleven-dataset suite at $L=96$ and
$H\in\{96,192,336,720\}$ over three seeds.
Within each setting and seed, correction methods share the same
frozen backbone checkpoint. MSE and MAE are reported in
train-standardized coordinates, with configurations and checkpoints
selected on validation data.

\begin{table*}[t]
\centering\footnotesize
\setlength{\tabcolsep}{3pt}
\caption{\textbf{Forecasting gains across backbones.}
Dataset-balanced MSE/MAE reductions (\%) over the frozen base. Bold/underline mark best/second best values. Detailed MSE and MAE results across datasets and backbones are reported in Appendices~\ref{app:pcr-full-grid-tables} and~\ref{app:pcr-competitors}.}
\label{tab:method-main}
\label{tab:pcr-competitors}
\resizebox{\textwidth}{!}{\begin{tabular}{@{}lrrrrrrrr@{}}
\toprule
Backbone
& \shortstack{PCR-K4\\(ours)}
& \shortstack{PCR-K2\\(ours)}
& \shortstack{CCM\\original}
& \shortstack{CCM\\matched}
& \shortstack{MoLE\\original}
& \shortstack{MoLE\\matched}
& \shortstack{VE\\original}
& \shortstack{VE\\matched} \\
\midrule
DLinear
& $\underline{3.86}\,/\,\underline{1.87}$
& $3.63\,/\,1.76$
& $2.14\,/\,1.59$
& $1.00\,/\,0.41$
& $\mathbf{10.37}\,/\,\mathbf{9.17}$
& $1.68\,/\,0.72$
& $-2.49\, /\, -1.11$
& $-1.65\, /\, -1.46$ \\
PatchTST
& $\mathbf{14.25}\,/\,\mathbf{11.17}$
& $\underline{13.31}\,/\,\underline{10.44}$
& $8.05\,/\,7.78$
& $6.27\,/\,5.67$
& \textemdash
& \textemdash
& $8.97\,/\,6.75$
& $11.29\,/\,7.14$ \\
iTransformer
& $\mathbf{10.22}\,/\,\mathbf{5.87}$
& $\underline{8.56}\,/\,\underline{4.75}$
& $4.70\,/\,3.36$
& $-1.70\,/\,-1.91$
& \textemdash
& \textemdash
& $7.32\,/\,4.10$
& $3.57\,/\,0.93$ \\
\bottomrule
\end{tabular}}
\end{table*}

PCR-K4 reduces dataset-balanced MSE by 3.86\%, 14.25\%, and
10.22\% on DLinear, PatchTST, and iTransformer, respectively
(Table~\ref{tab:method-main}). PCR-K2 retains most of the
accuracy benefit with at most two nonzero channel weights.
The complete grid includes ties and regressions, documented alongside
uncertainty and paired comparisons in Appendix~\ref{app:pcr-coverage}.

\textbf{PCR gains extend beyond matched shared capacity.}
We additionally fit a rank-64 Global correction, matching PCR's expert-factor budget, on the identical frozen checkpoints and under the same final optimization and model-selection protocol. Relative to this control, PCR-K4 reduces MSE by \(9.82\%\) on iTransformer, \(9.19\%\) on PatchTST, and \(1.87\%\) on DLinear. These results show that PCR's gains, particularly on the transformer backbones, cannot be explained by additional shared residual capacity alone. Further results and analyses are reported in Appendix~\ref{app:pcr-global-control}.

\paragraph{Comparison with existing methods.}
We compare PCR with original-capacity and expert-budget-matched variants of
MoLE~\citep{ni2024mole}, CCM~\citep{chen2024ccm}, and
VE~\citep{wang2024ve} (Table~\ref{tab:method-main}). Matching concerns
expert-bank capacity, not total trainable parameters; MoLE matched
also aligns input information. Implementation details appear in
Appendix~\ref{app:pcr-competitors}.

With its backbone frozen, PCR-K4 outperforms both CCM and VE variants
in dataset-balanced MSE and MAE across all three backbones.
Reductions relative to full-size/matched CCM are
$1.37\%/2.69\%$ on DLinear, $7.19\%/9.02\%$ on PatchTST,
and $5.66\%/10.97\%$ on iTransformer.
PCR-K4 also reduces MSE over matched MoLE by $1.96\%$.
Original MoLE achieves higher average DLinear accuracy;
fixed-capacity ablations show that calendar conditioning contributes
substantially to its gains under the evaluated recipe
(Appendix~\ref{app:pcr-calendar}).
Original $\delta$-Adapter achieves higher average MSE accuracy at
much larger parameter budgets
and higher median latency (Appendix~\ref{app:pcr-delta}).

\subsection{Computational Efficiency}
\label{sec:method-cost}

Learned-K4/K2 share PCR's expert bank, persistent channel mixtures,
and frozen backbone, comparing allocation-fitting procedures. They optimize
mixtures through softmax or straight-through top-two gradients,
respectively. PCR gradient-trains $Er(F+H)$ expert parameters and
estimates mixtures analytically; learned controls additionally
optimize $EC$ gate logits.

PCR has lower observed median fitting times than the corresponding
learned controls on both transformers. On DLinear, PCR-K2 is faster
and PCR-K4 slightly slower. Timings include cache construction and
exclude backbone pretraining; Appendix~\ref{app:pcr-cost} details
execution conditions, caching-related memory trade-offs, and
deployment measurements.

\begin{figure}[t]
\centering
\includegraphics[width=\columnwidth]{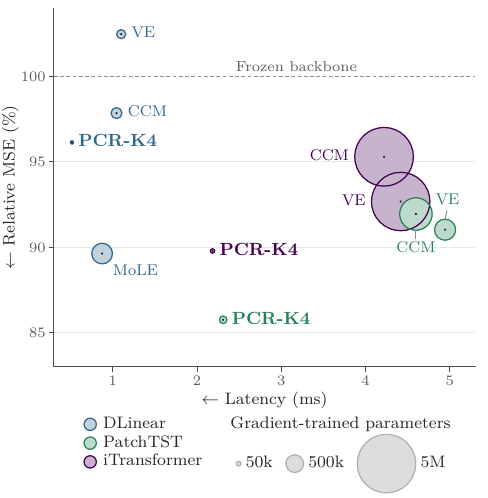}
\caption{\textbf{Accuracy-latency trade-offs.}
PCR-K4 and original-capacity CCM, MoLE, and VE.
Relative MSE to frozen backbone versus median single-window inference
latency; lower-left is better. Colours identify backbones, and circle
area denotes gradient-trained parameters. Matched variants appear in
Figure~\ref{fig:pcr-accuracy-latency-full}.}
\label{fig:pcr-accuracy-latency}
\end{figure}

PCR-K4 combines higher average accuracy, lower median single-window
latency, and fewer trainable parameters than both CCM and VE variants
on all three backbones and matched MoLE on DLinear.
Figure~\ref{fig:pcr-accuracy-latency} shows original-capacity competitors;
Figure~\ref{fig:pcr-accuracy-latency-full} includes the matched variants.
On iTransformer, PCR reduces backbone MSE by $10.22\%$, versus
$7.32\%$ for original VE, with approximately half the median latency
($2.18$ versus $4.42$\,ms) and $111$-fold fewer gradient-trained
parameters.
Original MoLE achieves higher average DLinear accuracy with higher
latency and more trainable parameters.
Parameter counts refer to gradient-based fitting; PCR retains its
frozen backbone and analytically fitted mixtures at deployment.

\section{Related Work}
\label{sec:related}

\paragraph{Parameter sharing and channel identity.}
Global and local forecasting formalize the choice between shared models and series-specific predictors \citep{monteromanso2021global}. STID incorporates spatial and temporal identity into forecasting \citep{shao2022stid}, while \citet{cini2023taming} study how node embeddings capture local effects within shared spatiotemporal models. Our study asks how additional expert capacity should be shared across channels, and evaluates this question across forecasting backbones with different treatments of channel dependence.

\paragraph{Expert allocation in forecasting.}
Time-MoE uses sparse expert activation to scale forecasting foundation models \citep{shi2025timemoe}. MoLE learns calendar-conditioned, channel-specific mixtures of linear forecasters \citep{ni2024mole}. CCM computes soft channel memberships from input embeddings and learned cluster prototypes, combining cluster-specific prediction heads \citep{chen2024ccm}. VE learns persistent variate embeddings that weight low-rank projection experts \citep{wang2024ve}, making it a direct comparator for persistent expert specialization (Section~\ref{sec:method-performance}). Building on these approaches, we isolate allocation quality and routing information under matched expert budgets and combine exact synthetic forecasting references with matched correction experiments to evaluate the benefits of informative channel allocation.

\paragraph{Efficient adaptation and mixture fitting.}
LoRA learns low-rank updates to frozen weights \citep{hu2022lora}.
C-LoRA introduces channel-specific low-rank feature adaptations
\citep{nie2024clora}, while ST-LoRA constructs node-specific low-rank
predictors \citep{ruan2026stlora}. $\delta$-Adapter improves frozen
forecasters through input nudging and output residual corrections
\citep{liang2026deltaadapter}; Appendix~\ref{app:pcr-delta} compares its accuracy and resource requirements with PCR. PCR focuses on persistent channel
allocation, estimating mixture weights analytically rather than
through a learned input-dependent router.

\section{Discussion and Conclusion}
\label{sec:discussion}

Effective specialization depends on learning what to share.
Informative channel allocation improves forecasting beyond matched shared capacity and random assignment. Persistent mixtures are particularly effective on frozen transformers, whereas DLinear benefits from input conditioning.

GalaxyTS makes the opportunity for specialization measurable through exact forecasting references. It shows how persistent channel information captures predictive differences concealed by short windows. Permuting fitted allocations degrades accuracy across all three backbones, supporting the functional importance of channel-specific sharing.

PCR translates these insights into practical specialization.
PCR-K4 reduces dataset-balanced MSE by 3.86-14.25\% across three frozen backbones. It outperforms both CCM and VE variants on all three backbones and MoLE under matched capacity and input information, while original MoLE remains stronger on DLinear. PCR achieves accuracy comparable to learned mixtures with shorter observed transformer fitting times, and has lower median single-window latency than the evaluated external competitors.

\paragraph{Limitations.}
Our study does not establish how expert rank and bank size should scale with predictive heterogeneity. Longer lookbacks may also reveal channel differences concealed by short windows, changing the benefit of persistent allocation. Understanding these interactions, and scaling exact mixture estimation to larger expert banks, remains open.

\label{main-text-end}

\nobalance
\subsection*{AI Use Statement}
Generative AI tools were used to assist with drafting and polishing the manuscript, including improving grammar, clarity, concision, and organization, and to assist with the refinement and organization of existing code. The authors reviewed and revised the AI-assisted text and code to ensure that the manuscript accurately reflected the methods, experimental findings, and conclusions. The authors take full responsibility for the final manuscript, including the accuracy of its claims, results, and references.

\bibliography{references}

\appendix
\onecolumn
\section*{Supplementary Materials}

\section{Shared Experimental Setup}
\label{app:implementation}
\label{app:numerical}

This appendix gives the datasets, preprocessing, and backbone configurations
used throughout the paper. Appendix~\ref{app:routing} expands the allocation
comparisons in Section~\ref{sec:routing}, and Appendix~\ref{app:observed}
examines backbone adaptation and lookback. Appendix~\ref{app:synthetic}
provides the GalaxyTS construction and allocation results for
Section~\ref{sec:synthetic}; Appendix~\ref{app:pcr-benchmark} gives PCR's
fitting details and evaluation for Section~\ref{sec:method}.
Additional deployment results and the supporting ridge identity appear in
Appendix~\ref{app:residuals}.

\subsection{Datasets, Splits, and Aggregation}
\label{app:datasets}

\paragraph{Datasets.}
Table~\ref{tab:datasets} summarizes the eleven datasets. ETT uses all
seven variables from ETDataset \citep{zhou2021informer}. Electricity and
Traffic use the hourly benchmark versions \citep{zeng2023transformers}
of the UCI electricity archive \citep{trindade2015electricity} and
California PeMS road occupancy \citep{caltrans_pems}, respectively.
Weather uses the Jena station measurements \citep{mpibgc_weather}
distributed with Autoformer \citep{wu2021autoformer}.
PEMS03/04/07/08 use the five-minute traffic-flow variable from the
ASTGNN releases \citep{caltrans_pems,guo2022astgnn}. Supporting panels specify
their dataset subsets.

\begin{table}[htbp]
\centering\small
\caption{Dataset dimensions before splitting. Each dataset in an ETT
pair has the listed channel count, length, and sampling interval.}
\label{tab:datasets}
\begin{tabular}{lrrl}
\toprule
Dataset & Channels & Rows & Sampling interval \\
\midrule
PEMS03 & 358 & 26,208 & 5 minutes \\
PEMS04 & 307 & 16,992 & 5 minutes \\
PEMS07 & 883 & 28,224 & 5 minutes \\
PEMS08 & 170 & 17,856 & 5 minutes \\
ETTh1 / ETTh2 & 7 & 17,420 & 1 hour \\
ETTm1 / ETTm2 & 7 & 69,680 & 15 minutes \\
Electricity & 321 & 26,304 & 1 hour \\
Traffic & 862 & 17,544 & 1 hour \\
Weather & 21 & 52,696 & 10 minutes \\
\bottomrule
\end{tabular}
\end{table}

\paragraph{Splits and preprocessing.}
For non-ETT datasets with $T$ observations, training and test contain
$\lfloor0.7T\rfloor$ and $\lfloor0.2T\rfloor$ observations; the remainder
between them forms validation. ETT uses 12/4/4 months for
training/validation/test, with 30 days per month; later observations are unused.
Each channel is standardized by its training mean and population standard
deviation, replacing deviations below $10^{-8}$ by one. Validation and test
use the same transformation. The evaluated files contain no NaNs; zeros
are retained without masking or imputation.

Windows use stride one. Validation and test histories may precede the split
boundary, but targets stay within their own partition. The loader produces
$n_{\mathrm{tr}}-L-H$, $n_{\mathrm{val}}-H$, and $n_{\mathrm{te}}-H$
training, validation, and test windows, respectively, omitting the last
otherwise admissible window. Ridge-based grouping uses an additional split
within training (Appendix~\ref{app:allocation-grouping}); GalaxyTS windowing
is specified in Appendix~\ref{app:synthetic-generation}.

iTransformer also receives hour, weekday, day-of-month, and day-of-year
features from the input window. These inputs are shared across its allocation
variants; no future calendar features are supplied.

\paragraph{Evaluation coordinates and metrics.}
MSE and MAE retain training-set channel standardization but reverse any
internal window normalization. For $N$ evaluated windows,
\begin{align}
 \mathrm{MSE}
 &= \frac{1}{NHC}
    \sum_{i=1}^{N}
    \|\widehat Y_i-Y_i\|_F^2, \\
 \mathrm{MAE}
 &= \frac{1}{NHC}
    \sum_{i=1}^{N}\sum_{h=1}^{H}\sum_{c=1}^{C}
    |(\widehat Y_i-Y_i)_{h,c}|.
\end{align}
All scalar targets receive equal weight, including those in incomplete batches.

\paragraph{Aggregation and uncertainty.}
Errors are reported as mean and sample standard deviation across seeds.
For methods $A$ and $B$, the MSE reduction of $A$ relative to $B$ is
$100(\mathrm{MSE}_B-\mathrm{MSE}_A)/\mathrm{MSE}_B$; positive values
favor $A$. Unless stated otherwise, reductions are computed within each seed
before averaging. Aggregate results average horizons within each dataset
and weight datasets equally; their SDs describe seed-level aggregates,
not variation across datasets. Calculations use unrounded errors.

Each study states its lookbacks, horizons, seeds, and reference.
Results with different selection rules are kept separate.
GalaxyTS distinguishes training-seed and realization variability
(Appendix~\ref{app:synthetic-protocol}); PCR's dataset-bootstrap intervals
are defined in Appendix~\ref{app:pcr-contrasts}. SDs are not confidence intervals.

\subsubsection{Asset Sources, Licenses, and Content}
\label{app:asset-provenance}

\paragraph{Existing assets.}
The following table identifies the data and implementation
sources used in the reported experiments. A code repository's license
is not treated as the license of datasets obtained through that repository.
Where no explicit license was identified, we report that status rather
than infer permission from public availability.

\begingroup\small
\edef\assettablecounter{\number\value{table}}
\renewcommand{\theHtable}{assets-\arabic{table}}
\setlength{\tabcolsep}{4pt}
\begin{longtable}{@{}p{0.18\textwidth}p{0.38\textwidth}p{0.38\textwidth}@{}}
\caption*{Existing asset sources and verified license information.
Unresolved entries describe the inspected release; they do not establish
redistribution permission.}\\
\toprule
Asset & Creator and source & License or reuse status \\
\midrule
\endfirsthead
\toprule
Asset & Creator and source & License or reuse status \\
\midrule
\endhead
\bottomrule
\endfoot
\endlastfoot
ETTh1/2, ETTm1/2 &
Zhou et al.; \href{https://github.com/zhouhaoyi/ETDataset}{ETDataset}
\citep{zhou2021informer}. &
\href{https://github.com/zhouhaoyi/ETDataset/blob/main/LICENSE}{CC BY-ND 4.0}. \\[3pt]
Electricity &
Trindade; \href{https://doi.org/10.24432/C58C86}{UCI archive}
\citep{trindade2015electricity}; hourly benchmark distributed through
Autoformer/LTSF \citep{wu2021autoformer,zeng2023transformers}. &
Original archive: \href{https://creativecommons.org/licenses/by/4.0/}{CC BY 4.0}.
Benchmark preprocessing is described separately from this source license. \\[3pt]
Weather &
\href{https://www.bgc-jena.mpg.de/wetter/weather_data.html}{MPI-BGC Jena station}
\citep{mpibgc_weather}; Autoformer benchmark \citep{wu2021autoformer}. &
\href{https://creativecommons.org/licenses/by/4.0/}{CC BY 4.0},
as stated by the station archive. \\[3pt]
Traffic &
Caltrans PeMS \citep{caltrans_pems}; hourly benchmark distributed through
Autoformer/LTSF \citep{wu2021autoformer,zeng2023transformers}. &
\href{https://dot.ca.gov/conditions-of-use}{Caltrans Conditions of Use};
no separate benchmark-specific data license identified. \\[3pt]
PEMS03/04/07/08 &
Caltrans PeMS \citep{caltrans_pems}; five-minute benchmark releases from
\href{https://github.com/guoshnBJTU/ASTGNN}{ASTGNN}
\citep{guo2022astgnn}. &
Caltrans source terms; no explicit separate benchmark-data license
identified in the inspected ASTGNN release. \\[3pt]
TSLib &
\href{https://github.com/thuml/Time-Series-Library}{THUML, Tsinghua University}
\citep{wang2026tslib}. &
\href{https://github.com/thuml/Time-Series-Library/blob/4e938a1767106324dd753b2a44832bf870a0252e/LICENSE}{MIT}. \\[3pt]
MoLE &
Ni et al.; \href{https://github.com/RogerNi/MoLE}{official implementation}
\citep{ni2024mole}. &
No explicit license identified in the inspected upstream release. \\[3pt]
CCM &
Chen et al.; \href{https://github.com/Graph-and-Geometric-Learning/TimeSeriesCCM}{official implementation}
\citep{chen2024ccm}. &
No explicit license identified in the inspected upstream release. \\[3pt]
VE &
\href{https://github.com/swang-song/VE}{Official implementation}
\citep{wang2024ve}. &
No explicit license identified in the inspected upstream release. \\[3pt]
$\delta$-Adapter &
\href{https://github.com/Anoise/Adapter}{Official implementation}
\citep{liang2026deltaadapter}. &
No explicit license identified in the inspected upstream release. \\[3pt]
Supporting implementation utilities &
\href{https://github.com/Muyiiiii/LatentTSF}{LatentTSF} (Jie Yang);
\href{https://github.com/ts-kim/RevIN}{RevIN} (ETRI);
\href{https://github.com/awslabs/gluonts}{GluonTS} (Amazon). &
LatentTSF retained subset: MIT, with upstream third-party notices;
RevIN core: MIT; retained GluonTS time features: Apache-2.0.
These licenses do not resolve permissions for CCM-specific additions. \\
\end{longtable}
\setcounter{table}{\assettablecounter}
\endgroup

\paragraph{Source identifiers.}
The recorded implementation revisions are TSLib
\texttt{4e938a1}, MoLE \texttt{a4f9fa8}, CCM \texttt{e4769ba},
the retained LatentTSF subset \texttt{7c8ae94}, VE \texttt{81e7ede},
and $\delta$-Adapter \texttt{0add06e}.
Full implementation commit identifiers and benchmark-file hashes will
accompany the public code release.
RevIN and GluonTS are credited for their retained utility components;
a separate exact upstream revision was not recorded for those components.

\paragraph{Release scope.}
The arXiv source package does not include experiment code or GalaxyTS dataset files.
Training and evaluation code is not currently publicly available; a public release is planned. The author-owned
GalaxyTS generator will be released under MIT, and its generated
synthetic datasets and metadata under CC BY 4.0.
These grants exclude third-party code and real-world benchmark data.
The public training-code release will retain applicable upstream notices
and require resolving redistribution permission for copied or adapted
components whose releases lack an explicit license.
For ETT, the NoDerivatives terms distinguish local adaptation from
sharing adapted material; reproducing the experiments does not require
redistributing processed ETT arrays.

\paragraph{Data provenance and sensitive content.}
The experiments use existing public benchmark releases of transformer
measurements, weather observations, road occupancy and traffic flow,
and electricity consumption indexed by pseudonymous client codes.
Inputs comprise numeric measurements and timestamps; GalaxyTS is
entirely synthetic. We collect no new participant data and perform no
identity linkage. Public availability is not treated as evidence of
original participant consent. Original collection-consent procedures
and benchmark-specific PeMS reuse requirements have not been independently
established here.

\subsection{Backbone Configurations and Pretraining}
\label{app:backbones}

We use the Time Series Library (TSLib) implementation of iTransformer \citep{wang2026tslib}.

Tables~\ref{tab:backbone-recipes} and~\ref{tab:base-rates} give the
architectures and initial learning rates. Standard pretraining minimizes
forecasting MSE with Adam, no weight decay, epoch-wise learning-rate
halving, a ten-epoch cap, and patience three. Validation selects the checkpoint.
DLinear and PatchTST use batch size 32 and select the initial rate on
seed-zero validation, keeping it fixed for other seeds. iTransformer uses
the listed dataset-specific rates and batch size 16, except on ETT and Weather,
where it uses 32. Exceptions are stated with the relevant study;
Appendix~\ref{app:interfaces} describes checkpoint matching, and
Appendix~\ref{app:lookback-protocol} gives the longer-lookback recipe.

\begingroup\small
\setlength{\tabcolsep}{4pt}
\begin{longtable}{llrrrrrl}
\caption{Backbone architecture recipes for the frozen benchmark. $d$ is model width, $F$ correction-feature width, $N$ encoder layers, $A$ attention heads, and $d_{\rm ff}$ feed-forward width. All iTransformer recipes use four time-mark tokens. PatchTST uses twelve length-16 patches at stride eight. DLinear uses decomposition width 25.}\label{tab:backbone-recipes}\\
\toprule
Backbone & Datasets / horizon & $d$ & $F$ & $N$ & $A$ & $d_{\rm ff}$ & Dropout \\
\midrule
\endfirsthead
\toprule
Backbone & Datasets / horizon & $d$ & $F$ & $N$ & $A$ & $d_{\rm ff}$ & Dropout \\
\midrule
\endhead
\bottomrule
\endfoot
\endlastfoot

DLinear & All & - & 192 & - & - & - & - \\
PatchTST & ETTh1, ETTh2 & 16 & 192 & 3 & 4 & 128 & 0.3 \\
 & All others & 128 & 1536 & 3 & 16 & 256 & 0.2 \\
iTransformer & ETTh1 / 96 & 256 & 256 & 2 & 8 & 256 & 0.1 \\
 & ETTh1 / 720 & 512 & 512 & 2 & 8 & 512 & 0.1 \\
 & Other ETT & 128 & 128 & 2 & 8 & 128 & 0.1 \\
 & Non-ETT & 512 & 512 & 3 & 8 & 512 & 0.1 \\
\bottomrule
\end{longtable}
\endgroup
\begingroup\small
\setlength{\tabcolsep}{4pt}
\begin{longtable}{lrrrr}
\caption{Base learning rates. A single value applies to both horizons; $a/b$ gives $H=96/720$. DLinear and PatchTST select rates on seed-zero validation; iTransformer rates are fixed by dataset.}\label{tab:base-rates}\\
\toprule
Dataset & $C$ & DLinear & PatchTST & iTransformer \\
\midrule
\endfirsthead
\multicolumn{5}{l}{\tablename~\thetable\ (continued)}\\
\toprule
Dataset & $C$ & DLinear & PatchTST & iTransformer \\
\midrule
\endhead
\midrule
\multicolumn{5}{r}{Continued on next page}\\
\endfoot
\bottomrule
\endlastfoot
ETTh1 & 7 & 0.01 & 0.001 & 0.0001 \\
ETTh2 & 7 & 0.01 & 0.001 & 0.0001 \\
ETTm1 & 7 & 0.01/0.001 & 0.0001/0.001 & 0.0001 \\
ETTm2 & 7 & 0.01 & 0.0001 & 0.0001 \\
Electricity & 321 & 0.01 & 0.001 & 0.0005 \\
PEMS03 & 358 & 0.001/0.01 & 0.001 & 0.0005 \\
PEMS04 & 307 & 0.01 & 0.001 & 0.0005 \\
PEMS07 & 883 & 0.01 & 0.001 & 0.0005 \\
PEMS08 & 170 & 0.01/0.001 & 0.001 & 0.0005 \\
Traffic & 862 & 0.01 & 0.001 & 0.0005 \\
Weather & 21 & 0.01/0.001 & 0.001 & 0.0001 \\
\end{longtable}
\endgroup

\Needspace{24\baselineskip}
\subsection{Experimental Coverage}

\begin{table}[H]
\centering\small
\setlength{\tabcolsep}{5pt}
\caption{Experimental coverage. A setting is a backbone-dataset-horizon
combination, extended by lookback or separation when these vary.
Real-data accuracy uses three reporting seeds.}
\label{tab:backbone-summary}
\begin{tabular}{p{0.25\textwidth}p{0.67\textwidth}}
\toprule
Study & Evaluated settings and comparisons \\
\midrule
Frozen allocation & Three backbones, eleven datasets, $H\in\{96,720\}$:
66 settings. Input-dependent and Ridge-based routing cover this grid. \\
Backbone adaptation & The same 66 settings: three jointly adapted
correction variants, backbone-only fine-tuning, and Input-dependent adaptation. \\
Lookback & iTransformer on Electricity, PEMS04, and Traffic,
$L,H\in\{96,720\}$: twelve settings, including
Input-dependent routing. \\
GalaxyTS & Three backbones, four separation levels, four horizons:
48 settings. Raw-window checks use three realizations; backbone-feature
allocation and PCR use three realizations and three training seeds. \\
PCR benchmark & Three backbones, eleven datasets, four horizons:
132 settings, four PCR/learned-allocation arms and the frozen base.
MoLE is evaluated on DLinear; CCM on DLinear, PatchTST, and our iTransformer adaptation. \\
Resource measurements & Allocation inference: 64 of 66 settings;
frozen/joint fitting: 24 settings; iTransformer lookback inference: twelve settings.
PCR and external deployment: 660 and 396 configurations, respectively, including 44 MoLE expert-budget controls. \\
\bottomrule
\end{tabular}
\end{table}

\subsection{Checkpoint Matching}
\label{app:interfaces}

Allocation, adaptation, and same-lookback accuracy comparisons start
from the same pretrained backbone checkpoint within each setting and seed.
Appendix~\ref{app:allocation-training} gives the selection rules.

Rank 16 was chosen by exploratory validation on four development datasets
and then held fixed across allocation methods. Results are conditional on
that choice, not evidence that it is globally optimal. The experiments do
not test a matched PEMS normalization intervention or a prospective rule
for predicting specialization gains or choosing the number of groups.

\clearpage
\section{What Drives Specialization Gains?}
\label{app:routing}
\label{app:allocation-controlled}

This appendix expands Section~\ref{sec:routing}: the correction inputs,
allocation rules, fitting procedure, and full comparisons behind the
capacity and routing conclusions. Results use seeds 0/1/2.
Appendix~\ref{app:interfaces} describes checkpoint matching across these studies.

\subsection{Correction Inputs and Routing Rules}
\label{app:correction-framework}

The residual framework of Section~\ref{sec:preliminaries} holds the
backbone fixed while changing how channels share low-rank experts.
For correction input $z_c(X)\in\mathbb{R}^{F}$,
\begin{equation}
 r_c(X)=\sum_{e=1}^{E}\pi_e(c,z_c(X))V_e^\top U_e^\top z_c(X),
 \label{eq:allocation-residual}
\end{equation}
where $U_e\in\mathbb{R}^{F\times r}$ and $V_e\in\mathbb{R}^{r\times H}$;
mixture weights are nonnegative and sum to one. Real-data corrections
use prediction-head features $z_c=h_c(X)$: concatenated seasonal/trend
histories for DLinear, flattened patch features for PatchTST, and channel
tokens for iTransformer. DLinear features retain training-set
standardization. GalaxyTS uses the same backbone-feature inputs at rank 16;
its supplementary raw-window checks instead use $z_c=x_c$, $F=L$.

If $b_c$ is the uncorrected head output and $a_c,m_c$ reverse window
normalization, the corrected forecast is
$\widehat y_c=a_c[b_c+\alpha r_c]+m_c$.
DLinear and raw-window synthetic corrections use $a_c=1,m_c=0$;
the latter also fixes $\alpha=1$. Transformer corrections are applied
before reversing per-window, per-channel normalization; the offset
already belongs to the backbone forecast.

Primary comparisons use $E=4$, rank $r=16$, and $K=2$ active experts.
Top-$K$ routing keeps the largest logits and softmax-normalizes them.
Persistent ID, Input-dependent, and Identity + input routing use
$\ell_c$, $g(z_c)$, and $\ell_c+g(z_c)$, respectively.
Dense ID uses every expert. Persistent mixtures are fixed across windows,
but their forecasts still respond to the current features.

\subsection{Allocation Rules and Parameter Budgets}
\label{app:allocation-arms}

Table~\ref{tab:comparison-glossary} defines each allocation rule.
Primary routed methods share a four-expert architecture; the rank-64
Global correction matches its expert parameter budget. Counts exclude
the validation-fitted scale and fixed group labels. Two active experts
do not imply lower computation: every expert input projection is evaluated.

\begingroup\small
\setlength{\tabcolsep}{3pt}
\begin{longtable}{p{0.22\textwidth}p{0.40\textwidth}ll}
\caption{Allocation rules and trainable parameters. $C$, $F$, and $H$ are channel count, feature width, and horizon. Dense ID uses $E\in\{2,4\}$ and $r=16$; per-channel correction is a larger-capacity reference.}\label{tab:comparison-glossary}\label{tab:allocation-parameters}\label{tab:comparison-variants}\\
\toprule
Alternative & Allocation & Expert weights & Router weights \\
\midrule
\endfirsthead
\toprule
Alternative & Allocation & Expert weights & Router weights \\
\midrule
\endhead
\bottomrule
\endlastfoot
Base & Frozen forecaster, without correction & $0$ & $0$ \\
Global correction & One shared rank-64 map & $64(F+H)$ & $0$ \\
Random routing & Balanced, equally weighted expert pairs & $64(F+H)$ & $0$ \\
Ridge-based routing & Predictive cluster and nearest alternative, equally weighted & $64(F+H)$ & $0$ \\
Persistent ID routing & Channel-specific top-2 logits & $64(F+H)$ & $4C$ \\
Input-dependent routing & Affine feature projection, top-2 & $64(F+H)$ & $4(F+1)$ \\
Identity + input routing & Channel logits plus feature projection, top-2 & $64(F+H)$ & $4C+4F$ \\
Dense ID (2/4) & Channel-specific mixture of all experts & $Er(F+H)$ & $EC$ \\
Global mixture & One top-2 mixture shared by channels and windows & $64(F+H)$ & $4$ \\
Donor-window routing & Identity + input router receives another window of the same channel & $64(F+H)$ & $4C+4F$ \\
Scalar modulation & One shared map scaled by $2\operatorname{sigmoid}(a_c+w^\top h_c)$ & $64(F+H)$ & $C+F$ \\
Per-channel correction & Separate rank-16 map per channel & $16C(F+H)$ & $0$ \\

\end{longtable}
\endgroup

Expert input factors have no bias and are initialized uniformly on
$[-r^{-1/2},r^{-1/2}]$; output factors start at zero, preserving the
initial backbone forecast. Global correction uses width 64 and bound
$64^{-1/2}$. Random pairs use seed $2000+s$ for training seed $s$.
Channel logits start from $\mathcal{N}(0,0.01^2)$.
Identity + input adds a zero-initialized, bias-free feature projection;
Input-dependent routing uses an affine projection without channel logits.
Router inputs receive no additional normalization.
For $q=\operatorname{softmax}(\ell)$ and top-2 mask $z$,
$p=(q\odot z)/\sum_e q_ez_e$; adding $q-\operatorname{stopgrad}(q)$
keeps the sparse forward mixture while supplying the dense-softmax gradient.

Donor-window routing is fitted separately and draws evaluation donors
from training data. The mean-gate check instead replaces an already fitted
input-conditioned mixture by its per-channel training-window mean and
recalibrates $\alpha$ without refitting experts. It therefore differs
from training Persistent ID routing.

\subsection{Optimization and Calibration}
\label{app:allocation-training}
\label{app:method-details}

Frozen corrections train at $\alpha=1$ with Adam, no weight decay,
batch size 16, and epoch-wise learning-rate halving. Training stops at
ten epochs or after three epochs without improvement in either calibrated
or uncalibrated validation MSE (threshold $10^{-12}$). The initial
zero-correction checkpoint remains eligible. Router rates are ten times
expert rates. Sparse Persistent ID, Input-dependent, Identity + input,
and Donor-window routing use the straight-through estimator and
load-balancing coefficient $5\times10^{-3}$; Global mixture has no
load-balancing term.

Validation jointly selects the checkpoint and correction scale.
Expert learning rates minimize mean validation MSE separately for each
backbone, dataset, horizon, and variant. The ten capacity/control variants
select over five seeds and report seeds 0/1/2 without retuning.
Ties favor the rate nearest $5\times10^{-4}$, then the smaller rate.
All variants select their expert learning rate from
$\{1.5\times10^{-4},5\times10^{-4},1.5\times10^{-3},5\times10^{-3}\}$.
Input-dependent routing selects over seeds 0/1/2.
Ridge-based routing also uses three selection seeds, except for the
six iTransformer settings (Electricity, Traffic, PEMS04 at both
horizons), which use five.

For validation residual $e=y-\widehat y^{\,0}$ and correction
$D=ar$ in evaluation coordinates, float64 accumulated statistics give
\begin{equation}
 \operatorname{MSE}_{\mathrm{val}}(\alpha)
 =\frac{S_{ee}-2\alpha S_{eD}+\alpha^2S_{DD}}{n},\qquad
 S_{uv}=\sum_i u_iv_i,
 \label{eq:global-calibration}
\end{equation}
where $i$ spans all validation targets and $n$ is their count.
Validation selects one global $\alpha\in[0,1]$ from the evaluated values,
including zero; ties favor the smaller scale. The uncalibrated result
selects its own checkpoint at $\alpha=1$.
Raw-window GalaxyTS corrections keep $\alpha=1$; backbone-feature
GalaxyTS and PCR use the selection rules stated with those studies.

\subsection{How Ridge-Based Routing Forms Groups}
\label{app:allocation-grouping}

Ridge-based routing groups channels by estimated forecasting maps,
not by their observed similarity. For each dataset and horizon, a
bias-free ridge map predicts each channel's $H$-step future from its
96-step history, using all contiguous training windows. The first 80\%
of these windows fit the maps; the remaining 20\% select a shared ridge
penalty by aggregate squared error. These adjacent inner windows are
not purged and can have overlapping targets, but both sets lie within
the outer training partition.

Let $n_{\mathrm{inner}}$ be the number of inner fitting windows.
Table~\ref{tab:selected-ridge} reports
$\lambda_{\mathrm{rel}}=\lambda_{\mathrm{selected}}/n_{\mathrm{inner}}$.
The full-training refit uses
$\lambda_{\mathrm{selected}}/0.8$.
Flattened maps are normalized by their Euclidean norm with floor
$10^{-12}$, reduced by PCA to $\min(50,C-1)$ components, and clustered
by $k$-means into four groups. PCA and clustering use random state zero;
$k$-means uses ten initializations. The same partition is used across
backbones for each dataset-horizon pair, including four groups on
seven-channel ETT, where singleton groups are possible.

Each channel equally combines the expert for its own group and the expert
at the nearest other centroid. Maps and groups are fixed before correction
training: they determine allocation, not the deployed forecasting maps,
and introduce no trainable router parameters.

\begin{table}[htbp]
\centering\small
\caption{Validation-selected relative ridge penalties. Each value
multiplies the number of inner fitting windows; the full-training
refit divides that penalty by $0.8$. These settings are shared
across backbones.}
\label{tab:selected-ridge}
\begin{tabular}{lrr}
\toprule
Dataset & $H=96$ & $H=720$ \\
\midrule
ETTh1 & 0.1 & 0.1 \\
ETTh2 & 0.001 & 10 \\
ETTm1 & 0.01 & 0.01 \\
ETTm2 & 0.001 & 1 \\
Electricity & 0.001 & 0.01 \\
PEMS03 & 0.001 & 0.001 \\
PEMS04 & 0.001 & 0.01 \\
PEMS07 & 0.001 & 0.001 \\
PEMS08 & 0.001 & 0.001 \\
Traffic & 0.01 & 0.01 \\
Weather & 0.1 & 0.001 \\
\bottomrule
\end{tabular}
\end{table}

\subsection{Allocation Results and Paired Comparisons}
\label{app:allocation-results}

For method $A$ and reference $B$, the dataset-level contrast is
\begin{equation}
 R_{b,d,s}^{A\mid B}=\frac{1}{2}\sum_{H\in\{96,720\}}
 100\frac{\mathrm{MSE}^{B}_{b,d,H,s}-\mathrm{MSE}^{A}_{b,d,H,s}}
 {\mathrm{MSE}^{B}_{b,d,H,s}}.
 \label{eq:allocation-dataset-contrast}
\end{equation}
Means and sample SDs summarize seeds 0/1/2; dataset averages receive
equal weight. Dataset wins count positive mean paired reductions.
Table~\ref{tab:allocation-all-arms} shows gains over the
frozen backbone, Table~\ref{tab:allocation-by-dataset} shows where
informative allocation and input conditioning help, and
Table~\ref{tab:allocation-contrasts} compares each method with its named
reference. Backbone-relative gains and pairwise reductions have different
denominators and should not be subtracted to obtain a comparison.

\begingroup\scriptsize
\setlength{\tabcolsep}{3pt}
\begin{longtable}{p{0.25\textwidth}rrr}
\caption{Frozen-backbone MSE reductions (\%). Results average all eleven datasets. Bold/underline mark best/second means, including the larger per-channel reference. All backbone comparisons use matched checkpoints within each setting and seed.}\label{tab:allocation-all-arms}\\
\toprule
Alternative & DLinear & PatchTST & iTransformer \\
\midrule
\endfirsthead
\toprule
Alternative & DLinear & PatchTST & iTransformer \\
\midrule
\endhead
\midrule
\multicolumn{4}{r}{Continued on next page}\\
\endfoot
\bottomrule
\endlastfoot
Global correction & $0.96$ & $5.79$ & $0.60$ \\
Random routing & $0.85$ & $6.27$ & $1.18$ \\
Ridge-based routing & $2.32$ & $7.95$ & $2.24$ \\
Persistent ID routing & $3.57$ & $12.41$ & $7.62$ \\
Input-dependent routing & $\underline{10.46}$ & $6.46$ & $0.49$ \\
Identity + input routing & $\mathbf{11.49}$ & $10.36$ & $7.20$ \\
Dense ID routing (2) & $2.54$ & $10.17$ & $5.37$ \\
Dense ID routing (4) & $3.76$ & $\underline{13.40}$ & $\underline{9.79}$ \\
Global mixture & $0.86$ & $5.64$ & $0.46$ \\
Scalar modulation & $4.20$ & $8.80$ & $3.13$ \\
Donor-window routing & $3.46$ & $9.83$ & $6.63$ \\
Per-channel correction & $4.23$ & $\mathbf{17.60}$ & $\mathbf{13.83}$ \\
\end{longtable}
\endgroup

\begingroup\small
\setlength{\tabcolsep}{3pt}
\begin{longtable}{lrrr}
\caption{Dataset-level relative MSE reductions (\%), mean $\pm$ sample SD over three seed-level horizon averages. Positive favors the first method. All backbone comparisons use matched checkpoints.}\label{tab:allocation-by-dataset}\\
\toprule
Dataset & \shortstack{Persistent ID routing\\vs. Global correction} & \shortstack{Persistent ID routing\\vs. Random routing} & \shortstack{Identity + input routing\\vs. Persistent ID routing} \\
\midrule
\endfirsthead
\toprule
Dataset & \shortstack{Persistent ID routing\\vs. Global correction} & \shortstack{Persistent ID routing\\vs. Random routing} & \shortstack{Identity + input routing\\vs. Persistent ID routing} \\
\midrule
\endhead
\midrule
\multicolumn{4}{r}{Continued on next page}\\
\endfoot
\bottomrule
\endlastfoot
\multicolumn{4}{l}{\textit{DLinear}} \\*
ETTh1 & $2.89\pm1.00$ & $0.47\pm0.67$ & $-2.38\pm0.70$ \\
ETTh2 & $-6.45\pm6.02$ & $-1.50\pm0.72$ & $-0.39\pm2.57$ \\
ETTm1 & $3.34\pm0.11$ & $0.32\pm0.31$ & $-2.05\pm0.13$ \\
ETTm2 & $-5.58\pm5.27$ & $-0.56\pm0.71$ & $2.84\pm1.45$ \\
Weather & $10.89\pm0.63$ & $9.41\pm0.33$ & $-1.80\pm1.62$ \\
Electricity & $2.59\pm0.06$ & $2.45\pm0.14$ & $3.68\pm0.20$ \\
Traffic & $1.46\pm0.02$ & $1.46\pm0.05$ & $6.33\pm0.30$ \\
PEMS03 & $1.91\pm0.03$ & $1.94\pm0.02$ & $22.61\pm0.69$ \\
PEMS04 & $3.31\pm0.07$ & $3.25\pm0.10$ & $19.99\pm0.44$ \\
PEMS07 & $5.06\pm0.19$ & $5.04\pm0.23$ & $22.25\pm0.54$ \\
PEMS08 & $7.90\pm0.74$ & $7.85\pm0.80$ & $20.12\pm1.54$ \\
\multicolumn{4}{l}{\textit{PatchTST}} \\*
ETTh1 & $0.08\pm0.50$ & $0.23\pm0.16$ & $0.70\pm0.22$ \\
ETTh2 & $-0.55\pm1.97$ & $-1.13\pm2.02$ & $0.94\pm1.59$ \\
ETTm1 & $1.57\pm0.11$ & $0.41\pm0.43$ & $-1.80\pm0.47$ \\
ETTm2 & $1.12\pm0.26$ & $0.29\pm0.28$ & $-1.14\pm0.30$ \\
Weather & $2.54\pm0.24$ & $1.24\pm0.21$ & $-2.07\pm0.39$ \\
Electricity & $4.25\pm0.44$ & $4.53\pm0.31$ & $-1.47\pm0.21$ \\
Traffic & $0.26\pm0.41$ & $0.72\pm0.48$ & $-0.19\pm0.61$ \\
PEMS03 & $13.81\pm0.33$ & $13.51\pm0.56$ & $-3.59\pm1.20$ \\
PEMS04 & $21.96\pm0.62$ & $21.19\pm0.79$ & $-8.30\pm3.54$ \\
PEMS07 & $19.26\pm0.16$ & $19.01\pm0.24$ & $-2.13\pm1.00$ \\
PEMS08 & $17.24\pm0.62$ & $16.52\pm0.67$ & $-10.98\pm3.23$ \\
\multicolumn{4}{l}{\textit{iTransformer}} \\*
ETTh1 & $-0.05\pm0.91$ & $-0.46\pm1.26$ & $-0.97\pm1.16$ \\
ETTh2 & $0.52\pm0.67$ & $0.08\pm0.27$ & $-0.99\pm0.59$ \\
ETTm1 & $1.70\pm0.33$ & $0.49\pm0.30$ & $-0.97\pm0.36$ \\
ETTm2 & $1.83\pm0.06$ & $0.78\pm0.27$ & $-1.62\pm0.64$ \\
Weather & $4.00\pm0.06$ & $2.65\pm0.41$ & $-2.03\pm0.64$ \\
Electricity & $2.29\pm0.22$ & $2.18\pm0.23$ & $0.04\pm0.19$ \\
Traffic & $0.05\pm0.15$ & $0.18\pm0.20$ & $0.07\pm0.18$ \\
PEMS03 & $13.43\pm0.32$ & $13.18\pm0.29$ & $0.30\pm0.98$ \\
PEMS04 & $20.33\pm0.41$ & $19.77\pm0.42$ & $0.07\pm1.56$ \\
PEMS07 & $18.56\pm0.47$ & $18.38\pm0.54$ & $2.35\pm0.61$ \\
PEMS08 & $15.14\pm0.41$ & $14.56\pm0.39$ & $-0.52\pm0.14$ \\
\end{longtable}
\endgroup

\begingroup\scriptsize
\setlength{\tabcolsep}{2.5pt}
\begin{longtable}{p{0.23\textwidth}p{0.21\textwidth}rrr}
\caption{Pairwise MSE reductions (\%), mean $\pm$ sample SD over three dataset-balanced seed averages. Brackets count favorable dataset averages. Every comparison covers 22 settings per backbone and uses matched checkpoints.}\label{tab:allocation-contrasts}\label{tab:allocation-diagnostics}\\
\toprule
Method & Reference & DLinear & PatchTST & iTransformer \\
\midrule
\endfirsthead
\toprule
Method & Reference & DLinear & PatchTST & iTransformer \\
\midrule
\endhead
\midrule
\multicolumn{5}{r}{Continued on next page}\\
\endfoot
\bottomrule
\endlastfoot
Persistent ID routing & Global correction & $2.48\pm0.57$ & $7.41\pm0.30$ & $7.07\pm0.07$ \\
Persistent ID routing & Random routing & $2.74\pm0.13$ [9/11] & $6.96\pm0.34$ [10/11] & $6.53\pm0.15$ [10/11] \\
Ridge-based routing & Random routing & $1.47\pm0.28$ & $1.91\pm0.12$ & $1.08\pm0.15$ \\
Identity + input routing & Persistent ID routing & $8.29\pm0.30$ & $-2.73\pm0.51$ & $-0.39\pm0.20$ \\
Input-dependent routing & Persistent ID routing & $7.10\pm0.63$ & $-8.12\pm0.42$ & $-8.53\pm0.08$ \\
Identity + input routing & Input-dependent routing & $1.20\pm0.55$ & $4.46\pm0.11$ & $6.77\pm0.20$ \\
Dense ID routing (4) & Persistent ID routing & $0.21\pm0.21$ & $1.38\pm0.26$ & $2.65\pm0.06$ \\
Persistent ID routing & Global mixture & $2.65\pm0.68$ [9/11] & $7.56\pm0.30$ [10/11] & $7.21\pm0.10$ [11/11] \\
Identity + input routing & Donor-window routing & $8.40\pm0.26$ [7/11] & $0.68\pm0.14$ [8/11] & $0.72\pm0.20$ [6/11] \\
Identity + input routing & Scalar modulation & $7.81\pm0.95$ [10/11] & $1.91\pm0.14$ [7/11] & $4.39\pm0.33$ [10/11] \\
\end{longtable}
\endgroup

Ridge-based routing is evaluated on all 66 settings. Comparisons with
Random and Persistent ID use matched checkpoints; it improves on Random
in 49/66 settings and on Persistent ID in 15/66.
Tables~\ref{tab:ridge-frozen-new} and~\ref{tab:input-frozen-new} give
the full Ridge-based and DLinear/PatchTST Input-dependent results;
Appendix~\ref{app:tslib-routing} gives the detailed iTransformer
routing results.

\begingroup\scriptsize
\setlength{\tabcolsep}{4pt}
\begin{longtable}{llrrrr}
\caption{Ridge-based routing, $L=96$: mean $\pm$ sample SD over three seeds. MSE/MAE use train-standardized coordinates; Gain is own-base MSE reduction (\%). $R_{\mathrm{Ridge}\mid\mathrm{Random}}$ uses matched checkpoints in all 66 settings.}\label{tab:ridge-frozen-new}\\
\toprule
Dataset & $H$ & MSE & MAE & Gain (\%) & $R_{\mathrm{Ridge}\mid\mathrm{Random}}$ \\
\midrule
\endfirsthead
\multicolumn{6}{l}{\tablename~\thetable\ (continued)}\\
\toprule
Dataset & $H$ & MSE & MAE & Gain (\%) & $R_{\mathrm{Ridge}\mid\mathrm{Random}}$ \\
\midrule
\endhead
\midrule
\multicolumn{6}{r}{Continued on next page}\\
\endfoot
\bottomrule
\endlastfoot
\multicolumn{6}{l}{\textit{DLinear}} \\*
ETTh1 & 96 & $0.3825\pm0.0029$ & $0.3956\pm0.0039$ & $0.91\pm0.08$ & $-0.31\pm0.56$ \\
ETTh1 & 720 & $0.5061\pm0.0157$ & $0.4998\pm0.0138$ & $5.56\pm2.61$ & $1.57\pm0.50$ \\
ETTh2 & 96 & $0.3184\pm0.0393$ & $0.3701\pm0.0319$ & $0.75\pm0.78$ & $0.16\pm1.61$ \\
ETTh2 & 720 & $0.7485\pm0.0564$ & $0.6225\pm0.0237$ & $0.00\pm0.00$ & $2.04\pm2.57$ \\
ETTm1 & 96 & $0.3311\pm0.0015$ & $0.3638\pm0.0018$ & $3.68\pm0.67$ & $-0.67\pm0.71$ \\
ETTm1 & 720 & $0.4726\pm0.0045$ & $0.4525\pm0.0035$ & $0.53\pm0.54$ & $-1.48\pm1.03$ \\
ETTm2 & 96 & $0.1844\pm0.0029$ & $0.2778\pm0.0052$ & $0.59\pm0.21$ & $-0.11\pm0.40$ \\
ETTm2 & 720 & $0.4695\pm0.0485$ & $0.4622\pm0.0320$ & $4.36\pm6.49$ & $0.76\pm1.45$ \\
Weather & 96 & $0.1837\pm0.0006$ & $0.2538\pm0.0017$ & $6.35\pm0.76$ & $4.10\pm1.53$ \\
Weather & 720 & $0.3433\pm0.0015$ & $0.3894\pm0.0022$ & $1.31\pm0.74$ & $0.70\pm0.94$ \\
Electricity & 96 & $0.1890\pm0.0000$ & $0.2733\pm0.0001$ & $2.85\pm0.02$ & $2.67\pm0.08$ \\
Electricity & 720 & $0.2368\pm0.0002$ & $0.3243\pm0.0003$ & $2.41\pm0.10$ & $2.12\pm0.05$ \\
Traffic & 96 & $0.6435\pm0.0005$ & $0.3955\pm0.0004$ & $0.79\pm0.06$ & $0.73\pm0.03$ \\
Traffic & 720 & $0.6420\pm0.0001$ & $0.3940\pm0.0001$ & $0.52\pm0.02$ & $0.53\pm0.02$ \\
PEMS03 & 96 & $0.4339\pm0.0018$ & $0.4979\pm0.0017$ & $1.67\pm0.33$ & $1.54\pm0.12$ \\
PEMS03 & 720 & $0.4286\pm0.0000$ & $0.4836\pm0.0000$ & $0.47\pm0.03$ & $0.43\pm0.05$ \\
PEMS04 & 96 & $0.4105\pm0.0005$ & $0.4747\pm0.0001$ & $1.04\pm0.09$ & $0.93\pm0.10$ \\
PEMS04 & 720 & $0.4311\pm0.0001$ & $0.4804\pm0.0003$ & $0.82\pm0.14$ & $0.77\pm0.08$ \\
PEMS07 & 96 & $0.5467\pm0.0013$ & $0.5274\pm0.0003$ & $2.93\pm0.25$ & $2.86\pm0.26$ \\
PEMS07 & 720 & $0.5391\pm0.0002$ & $0.5194\pm0.0000$ & $0.88\pm0.11$ & $0.88\pm0.11$ \\
PEMS08 & 96 & $0.4670\pm0.0012$ & $0.5108\pm0.0015$ & $8.18\pm0.65$ & $7.73\pm0.43$ \\
PEMS08 & 720 & $0.5175\pm0.0001$ & $0.5278\pm0.0014$ & $4.40\pm0.18$ & $4.40\pm0.05$ \\
\addlinespace[4pt]
\multicolumn{6}{l}{\textit{PatchTST}} \\*
ETTh1 & 96 & $0.3898\pm0.0018$ & $0.4016\pm0.0014$ & $3.77\pm0.53$ & $0.69\pm0.19$ \\
ETTh1 & 720 & $0.4847\pm0.0094$ & $0.4767\pm0.0066$ & $0.75\pm0.53$ & $0.89\pm0.64$ \\
ETTh2 & 96 & $0.2884\pm0.0023$ & $0.3385\pm0.0023$ & $1.62\pm0.35$ & $0.76\pm0.44$ \\
ETTh2 & 720 & $0.4211\pm0.0028$ & $0.4409\pm0.0018$ & $-0.10\pm0.40$ & $0.06\pm0.28$ \\
ETTm1 & 96 & $0.3207\pm0.0007$ & $0.3584\pm0.0003$ & $6.08\pm0.08$ & $0.60\pm0.86$ \\
ETTm1 & 720 & $0.4530\pm0.0008$ & $0.4446\pm0.0007$ & $1.12\pm0.32$ & $0.44\pm0.09$ \\
ETTm2 & 96 & $0.1754\pm0.0001$ & $0.2600\pm0.0004$ & $1.36\pm0.11$ & $-0.15\pm0.24$ \\
ETTm2 & 720 & $0.3996\pm0.0017$ & $0.3993\pm0.0013$ & $0.36\pm0.34$ & $-0.22\pm0.28$ \\
Weather & 96 & $0.1714\pm0.0006$ & $0.2131\pm0.0006$ & $0.22\pm0.31$ & $-1.69\pm0.67$ \\
Weather & 720 & $0.3510\pm0.0005$ & $0.3458\pm0.0001$ & $0.09\pm0.55$ & $-0.53\pm0.36$ \\
Electricity & 96 & $0.1691\pm0.0004$ & $0.2579\pm0.0003$ & $14.23\pm0.22$ & $3.87\pm0.27$ \\
Electricity & 720 & $0.2360\pm0.0012$ & $0.3171\pm0.0010$ & $11.06\pm0.38$ & $2.32\pm0.18$ \\
Traffic & 96 & $0.4808\pm0.0010$ & $0.3057\pm0.0008$ & $2.86\pm0.07$ & $-0.32\pm0.30$ \\
Traffic & 720 & $0.5324\pm0.0009$ & $0.3314\pm0.0004$ & $3.37\pm0.21$ & $-0.21\pm0.23$ \\
PEMS03 & 96 & $0.3871\pm0.0227$ & $0.4361\pm0.0155$ & $20.33\pm3.17$ & $5.30\pm0.44$ \\
PEMS03 & 720 & $0.4185\pm0.0026$ & $0.4469\pm0.0015$ & $18.49\pm0.68$ & $5.20\pm0.30$ \\
PEMS04 & 96 & $0.5211\pm0.0041$ & $0.5147\pm0.0028$ & $17.00\pm0.75$ & $5.77\pm0.17$ \\
PEMS04 & 720 & $0.5269\pm0.0025$ & $0.5151\pm0.0013$ & $14.08\pm0.08$ & $3.87\pm0.05$ \\
PEMS07 & 96 & $0.4350\pm0.0023$ & $0.4529\pm0.0008$ & $13.77\pm0.67$ & $3.61\pm0.17$ \\
PEMS07 & 720 & $0.4610\pm0.0012$ & $0.4693\pm0.0007$ & $10.11\pm0.70$ & $2.94\pm0.24$ \\
PEMS08 & 96 & $0.4495\pm0.0006$ & $0.4660\pm0.0006$ & $19.69\pm1.17$ & $4.62\pm0.27$ \\
PEMS08 & 720 & $0.5271\pm0.0072$ & $0.5000\pm0.0036$ & $14.69\pm0.61$ & $4.25\pm0.49$ \\
\addlinespace[4pt]
\multicolumn{6}{l}{\textit{iTransformer}} \\*
ETTh1 & 96 & $0.3853\pm0.0006$ & $0.4039\pm0.0010$ & $1.13\pm0.22$ & $0.68\pm0.50$ \\
ETTh1 & 720 & $0.5000\pm0.0040$ & $0.4885\pm0.0021$ & $-0.09\pm0.10$ & $-0.97\pm1.23$ \\
ETTh2 & 96 & $0.2953\pm0.0043$ & $0.3465\pm0.0026$ & $1.86\pm0.59$ & $0.32\pm0.30$ \\
ETTh2 & 720 & $0.4256\pm0.0015$ & $0.4446\pm0.0014$ & $0.57\pm0.34$ & $0.40\pm0.27$ \\
ETTm1 & 96 & $0.3352\pm0.0027$ & $0.3702\pm0.0019$ & $2.84\pm0.55$ & $0.67\pm0.53$ \\
ETTm1 & 720 & $0.4892\pm0.0021$ & $0.4547\pm0.0004$ & $0.66\pm0.07$ & $-0.17\pm0.38$ \\
ETTm2 & 96 & $0.1808\pm0.0010$ & $0.2646\pm0.0001$ & $2.38\pm0.38$ & $-0.39\pm0.20$ \\
ETTm2 & 720 & $0.4107\pm0.0019$ & $0.4027\pm0.0009$ & $0.84\pm0.18$ & $-0.17\pm0.46$ \\
Weather & 96 & $0.1722\pm0.0010$ & $0.2120\pm0.0007$ & $2.88\pm0.47$ & $-0.53\pm0.26$ \\
Weather & 720 & $0.3528\pm0.0014$ & $0.3467\pm0.0008$ & $2.02\pm0.33$ & $1.05\pm0.53$ \\
Electricity & 96 & $0.1471\pm0.0001$ & $0.2404\pm0.0000$ & $1.89\pm0.02$ & $1.24\pm0.08$ \\
Electricity & 720 & $0.2093\pm0.0002$ & $0.2994\pm0.0005$ & $1.32\pm0.15$ & $0.86\pm0.05$ \\
Traffic & 96 & $0.4071\pm0.0009$ & $0.2773\pm0.0005$ & $0.27\pm0.02$ & $-0.03\pm0.12$ \\
Traffic & 720 & $0.4770\pm0.0003$ & $0.3117\pm0.0000$ & $-0.06\pm0.02$ & $-0.20\pm0.06$ \\
PEMS03 & 96 & $0.2506\pm0.0039$ & $0.3500\pm0.0032$ & $3.86\pm0.32$ & $3.10\pm0.27$ \\
PEMS03 & 720 & $0.3188\pm0.0009$ & $0.3852\pm0.0013$ & $4.20\pm0.37$ & $3.35\pm0.29$ \\
PEMS04 & 96 & $0.2887\pm0.0056$ & $0.3766\pm0.0038$ & $5.05\pm0.12$ & $3.26\pm0.22$ \\
PEMS04 & 720 & $0.3232\pm0.0057$ & $0.3997\pm0.0042$ & $4.72\pm0.16$ & $3.58\pm0.16$ \\
PEMS07 & 96 & $0.2253\pm0.0015$ & $0.3235\pm0.0014$ & $3.39\pm0.19$ & $1.90\pm0.20$ \\
PEMS07 & 720 & $0.2762\pm0.0042$ & $0.3593\pm0.0026$ & $3.91\pm0.32$ & $2.61\pm0.33$ \\
PEMS08 & 96 & $0.3153\pm0.0060$ & $0.3847\pm0.0033$ & $2.78\pm0.12$ & $1.62\pm0.26$ \\
PEMS08 & 720 & $0.4010\pm0.0059$ & $0.4251\pm0.0043$ & $2.93\pm0.36$ & $1.50\pm0.20$ \\
\addlinespace[4pt]
\end{longtable}
\endgroup

\begingroup\scriptsize
\setlength{\tabcolsep}{3pt}
\begin{longtable}{llrrrrrr}
\caption{Input-dependent routing on DLinear and PatchTST, $L=96$: mean $\pm$ sample SD over three seeds. Gain is same-backbone MSE reduction (\%); $R_{I\mid ID}$ and $R_{ID+I\mid I}$ compare the indicated routers on matched checkpoints before horizon averaging. Bold/underline mark best/second-best Base versus Input-dependent MSE; ties share rank. iTransformer appears in Tables~\ref{tab:tslib-routing-errors} and~\ref{tab:tslib-routing-pairs}.}\label{tab:input-frozen-new}\\
\toprule
Dataset & $H$ & Base MSE & \shortstack{Input-dependent\\MSE} & \shortstack{Input-dependent\\MAE} & Gain (\%) & $R_{I\mid ID}$ & $R_{ID+I\mid I}$ \\
\midrule
\endfirsthead
\toprule
Dataset & $H$ & Base MSE & \shortstack{Input-dependent\\MSE} & \shortstack{Input-dependent\\MAE} & Gain (\%) & $R_{I\mid ID}$ & $R_{ID+I\mid I}$ \\
\midrule
\endhead
\midrule
\multicolumn{8}{r}{Continued on next page}\\
\endfoot
\bottomrule
\endlastfoot
\multicolumn{8}{l}{\textit{DLinear}} \\*
ETTh1 & 96 & $\underline{0.3860}\pm0.0031$ & $\mathbf{0.3859}\pm0.0017$ & $0.3981\pm0.0025$ & $0.01\pm0.40$ & $-1.60\pm0.25$ & $-0.06\pm0.07$ \\
ETTh1 & 720 & $\underline{0.5363}\pm0.0294$ & $\mathbf{0.5346}\pm0.0049$ & $0.5205\pm0.0043$ & $0.15\pm4.67$ & $-4.61\pm1.00$ & $1.44\pm1.58$ \\
ETTh2 & 96 & $\underline{0.3208}\pm0.0397$ & $\mathbf{0.3101}\pm0.0230$ & $0.3650\pm0.0216$ & $2.99\pm4.57$ & $2.10\pm5.41$ & $-0.91\pm1.02$ \\
ETTh2 & 720 & $\mathbf{0.7485}\pm0.0564$ & $\underline{0.7599}\pm0.0827$ & $0.6213\pm0.0271$ & $-1.48\pm6.78$ & $3.63\pm9.01$ & $-6.61\pm11.71$ \\
ETTm1 & 96 & $\underline{0.3438}\pm0.0014$ & $\mathbf{0.3360}\pm0.0021$ & $0.3661\pm0.0017$ & $2.27\pm1.02$ & $-2.58\pm0.65$ & $-0.54\pm1.06$ \\
ETTm1 & 720 & $\mathbf{0.4751}\pm0.0029$ & $\underline{0.4757}\pm0.0022$ & $0.4544\pm0.0026$ & $-0.13\pm0.37$ & $-2.37\pm0.37$ & $1.37\pm0.89$ \\
ETTm2 & 96 & $\underline{0.1855}\pm0.0032$ & $\mathbf{0.1830}\pm0.0040$ & $0.2742\pm0.0056$ & $1.38\pm1.10$ & $1.32\pm1.28$ & $-0.70\pm1.47$ \\
ETTm2 & 720 & $\underline{0.4912}\pm0.0428$ & $\mathbf{0.4507}\pm0.0341$ & $0.4532\pm0.0228$ & $8.09\pm5.09$ & $5.02\pm2.04$ & $0.02\pm1.11$ \\
Electricity & 96 & $\underline{0.1945}\pm0.0000$ & $\mathbf{0.1857}\pm0.0004$ & $0.2694\pm0.0005$ & $4.52\pm0.19$ & $1.85\pm0.07$ & $4.71\pm0.16$ \\
Electricity & 720 & $\underline{0.2426}\pm0.0004$ & $\mathbf{0.2369}\pm0.0001$ & $0.3239\pm0.0002$ & $2.34\pm0.13$ & $-0.31\pm0.06$ & $1.19\pm0.18$ \\
PEMS03 & 96 & $\underline{0.4413}\pm0.0030$ & $\mathbf{0.3184}\pm0.0021$ & $0.4122\pm0.0017$ & $27.84\pm0.42$ & $25.99\pm0.13$ & $2.34\pm1.53$ \\
PEMS03 & 720 & $\underline{0.4306}\pm0.0002$ & $\mathbf{0.3534}\pm0.0019$ & $0.4288\pm0.0009$ & $17.94\pm0.47$ & $16.65\pm0.43$ & $1.00\pm0.99$ \\
PEMS04 & 96 & $\underline{0.4148}\pm0.0004$ & $\mathbf{0.3093}\pm0.0024$ & $0.4051\pm0.0010$ & $25.44\pm0.65$ & $22.57\pm0.68$ & $2.32\pm1.06$ \\
PEMS04 & 720 & $\underline{0.4347}\pm0.0006$ & $\mathbf{0.3603}\pm0.0015$ & $0.4313\pm0.0012$ & $17.12\pm0.33$ & $14.61\pm0.34$ & $1.19\pm0.36$ \\
PEMS07 & 96 & $\underline{0.5632}\pm0.0008$ & $\mathbf{0.3860}\pm0.0057$ & $0.4325\pm0.0008$ & $31.47\pm1.11$ & $26.85\pm1.01$ & $2.28\pm2.13$ \\
PEMS07 & 720 & $\underline{0.5439}\pm0.0004$ & $\mathbf{0.4491}\pm0.0085$ & $0.4604\pm0.0020$ & $17.43\pm1.52$ & $14.13\pm1.68$ & $2.12\pm0.62$ \\
PEMS08 & 96 & $\underline{0.5086}\pm0.0042$ & $\mathbf{0.3519}\pm0.0128$ & $0.4243\pm0.0065$ & $30.79\pm3.02$ & $22.33\pm2.20$ & $2.26\pm4.71$ \\
PEMS08 & 720 & $\underline{0.5413}\pm0.0010$ & $\mathbf{0.4362}\pm0.0050$ & $0.4665\pm0.0010$ & $19.42\pm1.03$ & $15.00\pm0.50$ & $1.28\pm2.31$ \\
Traffic & 96 & $\underline{0.6487}\pm0.0002$ & $\mathbf{0.5859}\pm0.0036$ & $0.3632\pm0.0007$ & $9.67\pm0.54$ & $8.39\pm0.50$ & $-0.47\pm0.53$ \\
Traffic & 720 & $\underline{0.6454}\pm0.0001$ & $\mathbf{0.6049}\pm0.0036$ & $0.3716\pm0.0016$ & $6.27\pm0.57$ & $4.76\pm0.74$ & $-0.06\pm0.32$ \\
Weather & 96 & $\underline{0.1961}\pm0.0010$ & $\mathbf{0.1877}\pm0.0024$ & $0.2424\pm0.0038$ & $4.28\pm1.72$ & $-15.61\pm1.86$ & $11.28\pm3.86$ \\
Weather & 720 & $\underline{0.3478}\pm0.0028$ & $\mathbf{0.3396}\pm0.0030$ & $0.3689\pm0.0098$ & $2.37\pm0.07$ & $-1.91\pm1.12$ & $0.81\pm0.62$ \\
\multicolumn{8}{l}{\textit{PatchTST}} \\*
ETTh1 & 96 & $\underline{0.4050}\pm0.0004$ & $\mathbf{0.3888}\pm0.0017$ & $0.4024\pm0.0008$ & $4.00\pm0.45$ & $1.04\pm0.40$ & $-0.10\pm0.58$ \\
ETTh1 & 720 & $\underline{0.4883}\pm0.0092$ & $\mathbf{0.4828}\pm0.0130$ & $0.4776\pm0.0069$ & $1.14\pm0.83$ & $0.71\pm1.22$ & $-0.25\pm0.64$ \\
ETTh2 & 96 & $\underline{0.2931}\pm0.0015$ & $\mathbf{0.2911}\pm0.0021$ & $0.3415\pm0.0012$ & $0.68\pm0.21$ & $-0.01\pm0.02$ & $0.09\pm0.26$ \\
ETTh2 & 720 & $\mathbf{0.4207}\pm0.0015$ & $\underline{0.4257}\pm0.0041$ & $0.4430\pm0.0022$ & $-1.19\pm0.85$ & $0.95\pm2.95$ & $0.88\pm0.65$ \\
ETTm1 & 96 & $\underline{0.3414}\pm0.0010$ & $\mathbf{0.3279}\pm0.0017$ & $0.3664\pm0.0011$ & $3.95\pm0.44$ & $-2.23\pm0.47$ & $-0.83\pm0.76$ \\
ETTm1 & 720 & $\underline{0.4582}\pm0.0012$ & $\mathbf{0.4561}\pm0.0009$ & $0.4459\pm0.0002$ & $0.44\pm0.13$ & $-0.49\pm0.33$ & $-0.03\pm0.14$ \\
ETTm2 & 96 & $\underline{0.1778}\pm0.0002$ & $\mathbf{0.1771}\pm0.0005$ & $0.2623\pm0.0007$ & $0.38\pm0.16$ & $-1.49\pm0.16$ & $0.01\pm0.16$ \\
ETTm2 & 720 & $\underline{0.4010}\pm0.0008$ & $\mathbf{0.4007}\pm0.0007$ & $0.4002\pm0.0007$ & $0.08\pm0.03$ & $-0.75\pm0.52$ & $-0.04\pm0.09$ \\
Electricity & 96 & $\underline{0.1972}\pm0.0005$ & $\mathbf{0.1760}\pm0.0002$ & $0.2622\pm0.0002$ & $10.76\pm0.14$ & $-5.98\pm0.56$ & $3.79\pm0.43$ \\
Electricity & 720 & $\underline{0.2653}\pm0.0021$ & $\mathbf{0.2409}\pm0.0011$ & $0.3203\pm0.0010$ & $9.18\pm0.40$ & $-3.29\pm0.28$ & $2.23\pm0.15$ \\
PEMS03 & 96 & $\underline{0.4856}\pm0.0090$ & $\mathbf{0.4039}\pm0.0214$ & $0.4455\pm0.0137$ & $16.86\pm2.84$ & $-17.38\pm1.03$ & $11.85\pm1.18$ \\
PEMS03 & 720 & $\underline{0.5135}\pm0.0040$ & $\mathbf{0.4398}\pm0.0026$ & $0.4602\pm0.0015$ & $14.35\pm1.02$ & $-12.21\pm0.57$ & $7.57\pm1.34$ \\
PEMS04 & 96 & $\underline{0.6279}\pm0.0022$ & $\mathbf{0.5507}\pm0.0051$ & $0.5301\pm0.0023$ & $12.29\pm1.10$ & $-30.91\pm1.07$ & $15.12\pm5.31$ \\
PEMS04 & 720 & $\underline{0.6132}\pm0.0028$ & $\mathbf{0.5476}\pm0.0025$ & $0.5240\pm0.0015$ & $10.70\pm0.14$ & $-22.54\pm0.51$ & $13.90\pm0.97$ \\
PEMS07 & 96 & $\underline{0.5044}\pm0.0033$ & $\mathbf{0.4415}\pm0.0039$ & $0.4601\pm0.0030$ & $12.48\pm0.75$ & $-26.14\pm0.62$ & $19.57\pm1.85$ \\
PEMS07 & 720 & $\underline{0.5129}\pm0.0053$ & $\mathbf{0.4674}\pm0.0018$ & $0.4749\pm0.0012$ & $8.88\pm0.98$ & $-16.54\pm0.34$ & $11.79\pm0.29$ \\
PEMS08 & 96 & $\underline{0.5598}\pm0.0075$ & $\mathbf{0.4649}\pm0.0038$ & $0.4756\pm0.0026$ & $16.94\pm0.44$ & $-21.38\pm2.33$ & $10.12\pm3.59$ \\
PEMS08 & 720 & $\underline{0.6180}\pm0.0119$ & $\mathbf{0.5379}\pm0.0171$ & $0.5043\pm0.0095$ & $12.96\pm1.70$ & $-14.05\pm2.15$ & $1.05\pm3.08$ \\
Traffic & 96 & $\underline{0.4949}\pm0.0007$ & $\mathbf{0.4777}\pm0.0007$ & $0.3052\pm0.0005$ & $3.48\pm0.01$ & $-0.47\pm0.89$ & $0.30\pm0.16$ \\
Traffic & 720 & $\underline{0.5510}\pm0.0010$ & $\mathbf{0.5293}\pm0.0011$ & $0.3301\pm0.0005$ & $3.94\pm0.03$ & $-0.30\pm1.07$ & $0.09\pm0.16$ \\
Weather & 96 & $\mathbf{0.1718}\pm0.0002$ & $\mathbf{0.1718}\pm0.0004$ & $0.2135\pm0.0003$ & $-0.03\pm0.13$ & $-4.38\pm0.20$ & $0.78\pm0.97$ \\
Weather & 720 & $\mathbf{0.3513}\pm0.0018$ & $\underline{0.3514}\pm0.0015$ & $0.3450\pm0.0013$ & $-0.03\pm0.11$ & $-0.81\pm0.47$ & $0.23\pm0.20$ \\
\end{longtable}
\endgroup

\subsection{Input Conditioning}
\label{app:tslib-routing}

For iTransformer, all three routers use four rank-16 experts, two active,
and the fitting and selection rules in Appendix~\ref{app:allocation-training}.
Validation selects the checkpoint and $\alpha\in\{0,0.05,\ldots,1\}$.

Persistent ID improves on Input-dependent routing in 21/22 settings
and all eleven dataset averages; all three seeds agree in 20 settings.
Identity + input improves on Input-dependent in 19 settings and ten
datasets. Persistent ID improves on Identity + input in 13 settings
and six datasets, with all three seeds favoring Persistent ID in
nine settings and Identity + input in four. Input conditioning helps
most clearly on PEMS07 at $H=96$
($R_{ID\mid ID+I}=-3.58\pm1.61\%$).
Tables~\ref{tab:tslib-routing-summary}-\ref{tab:tslib-routing-pairs}
give the aggregate and setting-level results.

\begingroup\small
\setlength{\tabcolsep}{3pt}
\begin{longtable}{lrrr}
\caption{iTransformer routing: backbone-relative reductions (\%), mean $\pm$ sample SD over three seed-level averages. All results average the 22 settings across eleven datasets. The uncalibrated readout uses each router's selected learning rate, with the checkpoint selected at $\alpha=1$.}\label{tab:tslib-routing-summary}\\
\toprule
Router & \shortstack{MSE\\All} & \shortstack{MAE\\All} & \shortstack{MSE\\$\alpha=1$} \\
\midrule
\endfirsthead
\toprule
Router & \shortstack{MSE\\All} & \shortstack{MAE\\All} & \shortstack{MSE\\$\alpha=1$} \\
\midrule
\endhead
\midrule
\multicolumn{4}{r}{Continued on next page}\\
\endfoot
\bottomrule
\endlastfoot
Input-dependent & $0.49\pm0.03$ & $0.28\pm0.02$ & $0.29\pm0.06$ \\
Persistent ID & $7.62\pm0.10$ & $4.14\pm0.05$ & $7.26\pm0.09$ \\
Identity + input & $7.20\pm0.19$ & $3.79\pm0.12$ & $7.02\pm0.23$ \\
\end{longtable}
\endgroup

\begingroup\scriptsize
\setlength{\tabcolsep}{3pt}
\begin{longtable}{llrrr}
\caption{iTransformer routing errors. Each entry gives MSE (top) and MAE (bottom), mean $\pm$ sample SD over three seeds in train-standardized coordinates. Input, ID and ID+input denote the three routing rules. All 22 settings are retained.}\label{tab:tslib-routing-errors}\\
\toprule
Dataset & $H$ & Input & ID & ID+input \\
\midrule
\endfirsthead
\toprule
Dataset & $H$ & Input & ID & ID+input \\
\midrule
\endhead
\bottomrule
\endlastfoot
ETTh1 & 96 & \shortstack{$0.3928\pm0.0016$\\$0.4087\pm0.0013$} & \shortstack{$0.3886\pm0.0019$\\$0.4062\pm0.0015$} & \shortstack{$0.3939\pm0.0002$\\$0.4096\pm0.0008$} \\
ETTh1 & 720 & \shortstack{$0.5020\pm0.0069$\\$0.4887\pm0.0038$} & \shortstack{$0.4990\pm0.0061$\\$0.4861\pm0.0039$} & \shortstack{$0.5018\pm0.0076$\\$0.4887\pm0.0042$} \\
ETTh2 & 96 & \shortstack{$0.3006\pm0.0023$\\$0.3504\pm0.0008$} & \shortstack{$0.2958\pm0.0040$\\$0.3465\pm0.0021$} & \shortstack{$0.3007\pm0.0021$\\$0.3506\pm0.0007$} \\
ETTh2 & 720 & \shortstack{$0.4296\pm0.0032$\\$0.4482\pm0.0023$} & \shortstack{$0.4272\pm0.0025$\\$0.4455\pm0.0017$} & \shortstack{$0.4286\pm0.0026$\\$0.4473\pm0.0021$} \\
ETTm1 & 96 & \shortstack{$0.3440\pm0.0020$\\$0.3775\pm0.0014$} & \shortstack{$0.3353\pm0.0017$\\$0.3704\pm0.0011$} & \shortstack{$0.3380\pm0.0033$\\$0.3733\pm0.0022$} \\
ETTm1 & 720 & \shortstack{$0.4929\pm0.0020$\\$0.4582\pm0.0007$} & \shortstack{$0.4866\pm0.0019$\\$0.4538\pm0.0006$} & \shortstack{$0.4923\pm0.0014$\\$0.4559\pm0.0014$} \\
ETTm2 & 96 & \shortstack{$0.1837\pm0.0010$\\$0.2682\pm0.0004$} & \shortstack{$0.1781\pm0.0014$\\$0.2605\pm0.0013$} & \shortstack{$0.1817\pm0.0012$\\$0.2653\pm0.0023$} \\
ETTm2 & 720 & \shortstack{$0.4134\pm0.0026$\\$0.4067\pm0.0016$} & \shortstack{$0.4083\pm0.0021$\\$0.4005\pm0.0010$} & \shortstack{$0.4133\pm0.0016$\\$0.4067\pm0.0010$} \\
Electricity & 96 & \shortstack{$0.1492\pm0.0002$\\$0.2413\pm0.0001$} & \shortstack{$0.1450\pm0.0003$\\$0.2392\pm0.0003$} & \shortstack{$0.1449\pm0.0006$\\$0.2394\pm0.0004$} \\
Electricity & 720 & \shortstack{$0.2115\pm0.0002$\\$0.3006\pm0.0005$} & \shortstack{$0.2075\pm0.0003$\\$0.2977\pm0.0004$} & \shortstack{$0.2073\pm0.0003$\\$0.2979\pm0.0001$} \\
Traffic & 96 & \shortstack{$0.4065\pm0.0008$\\$0.2769\pm0.0006$} & \shortstack{$0.4058\pm0.0010$\\$0.2764\pm0.0004$} & \shortstack{$0.4055\pm0.0013$\\$0.2769\pm0.0002$} \\
Traffic & 720 & \shortstack{$0.4753\pm0.0004$\\$0.3111\pm0.0001$} & \shortstack{$0.4757\pm0.0015$\\$0.3112\pm0.0003$} & \shortstack{$0.4754\pm0.0003$\\$0.3117\pm0.0003$} \\
Weather & 96 & \shortstack{$0.1748\pm0.0010$\\$0.2155\pm0.0010$} & \shortstack{$0.1642\pm0.0006$\\$0.2073\pm0.0004$} & \shortstack{$0.1690\pm0.0014$\\$0.2103\pm0.0010$} \\
Weather & 720 & \shortstack{$0.3593\pm0.0020$\\$0.3504\pm0.0008$} & \shortstack{$0.3525\pm0.0021$\\$0.3481\pm0.0017$} & \shortstack{$0.3566\pm0.0014$\\$0.3496\pm0.0008$} \\
PEMS03 & 96 & \shortstack{$0.2594\pm0.0047$\\$0.3555\pm0.0035$} & \shortstack{$0.2188\pm0.0022$\\$0.3252\pm0.0021$} & \shortstack{$0.2158\pm0.0047$\\$0.3222\pm0.0039$} \\
PEMS03 & 720 & \shortstack{$0.3305\pm0.0011$\\$0.3933\pm0.0014$} & \shortstack{$0.2937\pm0.0007$\\$0.3691\pm0.0008$} & \shortstack{$0.2960\pm0.0028$\\$0.3705\pm0.0008$} \\
PEMS04 & 96 & \shortstack{$0.3013\pm0.0060$\\$0.3840\pm0.0040$} & \shortstack{$0.2334\pm0.0042$\\$0.3403\pm0.0033$} & \shortstack{$0.2308\pm0.0071$\\$0.3386\pm0.0048$} \\
PEMS04 & 720 & \shortstack{$0.3359\pm0.0062$\\$0.4056\pm0.0046$} & \shortstack{$0.2757\pm0.0063$\\$0.3689\pm0.0046$} & \shortstack{$0.2785\pm0.0096$\\$0.3712\pm0.0068$} \\
PEMS07 & 96 & \shortstack{$0.2302\pm0.0017$\\$0.3267\pm0.0017$} & \shortstack{$0.1820\pm0.0003$\\$0.2955\pm0.0004$} & \shortstack{$0.1758\pm0.0025$\\$0.2895\pm0.0026$} \\
PEMS07 & 720 & \shortstack{$0.2833\pm0.0048$\\$0.3644\pm0.0033$} & \shortstack{$0.2382\pm0.0045$\\$0.3363\pm0.0033$} & \shortstack{$0.2351\pm0.0031$\\$0.3340\pm0.0022$} \\
PEMS08 & 96 & \shortstack{$0.3219\pm0.0050$\\$0.3876\pm0.0026$} & \shortstack{$0.2674\pm0.0064$\\$0.3522\pm0.0036$} & \shortstack{$0.2655\pm0.0087$\\$0.3514\pm0.0048$} \\
PEMS08 & 720 & \shortstack{$0.4069\pm0.0084$\\$0.4276\pm0.0059$} & \shortstack{$0.3560\pm0.0060$\\$0.3960\pm0.0039$} & \shortstack{$0.3623\pm0.0085$\\$0.4006\pm0.0073$} \\
\end{longtable}
\endgroup

\begingroup\scriptsize
\setlength{\tabcolsep}{3pt}
\begin{longtable}{llrrr}
\caption{iTransformer relative MSE reductions (\%), mean $\pm$ sample SD over three seeds. Positive favors the first router; the reference router supplies the denominator. All averages the 22 settings within each seed.}\label{tab:tslib-routing-pairs}\\
\toprule
Dataset & $H$ & $R_{ID\mid I}$ & $R_{ID+I\mid I}$ & $R_{ID\mid ID+I}$ \\
\midrule
\endfirsthead
\toprule
Dataset & $H$ & $R_{ID\mid I}$ & $R_{ID+I\mid I}$ & $R_{ID\mid ID+I}$ \\
\midrule
\endhead
\bottomrule
\endlastfoot
ETTh1 & 96 & $1.07\pm0.87$ & $-0.28\pm0.35$ & $1.35\pm0.54$ \\
ETTh1 & 720 & $0.61\pm1.35$ & $0.04\pm0.45$ & $0.56\pm1.76$ \\
ETTh2 & 96 & $1.57\pm0.60$ & $-0.05\pm0.17$ & $1.62\pm0.71$ \\
ETTh2 & 720 & $0.56\pm0.18$ & $0.23\pm0.65$ & $0.33\pm0.48$ \\
ETTm1 & 96 & $2.52\pm0.44$ & $1.76\pm0.62$ & $0.77\pm0.47$ \\
ETTm1 & 720 & $1.27\pm0.17$ & $0.12\pm0.16$ & $1.15\pm0.26$ \\
ETTm2 & 96 & $3.07\pm0.19$ & $1.11\pm1.17$ & $1.97\pm1.36$ \\
ETTm2 & 720 & $1.23\pm0.30$ & $0.02\pm0.25$ & $1.21\pm0.31$ \\
Electricity & 96 & $2.86\pm0.34$ & $2.87\pm0.32$ & $-0.01\pm0.52$ \\
Electricity & 720 & $1.89\pm0.23$ & $1.95\pm0.21$ & $-0.07\pm0.15$ \\
Traffic & 96 & $0.16\pm0.05$ & $0.24\pm0.15$ & $-0.08\pm0.11$ \\
Traffic & 720 & $-0.09\pm0.25$ & $-0.02\pm0.13$ & $-0.07\pm0.37$ \\
Weather & 96 & $6.08\pm0.63$ & $3.34\pm1.33$ & $2.82\pm0.80$ \\
Weather & 720 & $1.88\pm0.14$ & $0.75\pm0.42$ & $1.13\pm0.53$ \\
PEMS03 & 96 & $15.66\pm0.71$ & $16.83\pm0.41$ & $-1.41\pm1.21$ \\
PEMS03 & 720 & $11.14\pm0.44$ & $10.44\pm1.13$ & $0.78\pm0.94$ \\
PEMS04 & 96 & $22.53\pm0.24$ & $23.42\pm0.89$ & $-1.17\pm1.47$ \\
PEMS04 & 720 & $17.92\pm0.80$ & $17.11\pm1.32$ & $0.97\pm1.65$ \\
PEMS07 & 96 & $20.93\pm0.65$ & $23.66\pm0.63$ & $-3.58\pm1.61$ \\
PEMS07 & 720 & $15.94\pm0.77$ & $16.99\pm0.72$ & $-1.27\pm0.60$ \\
PEMS08 & 96 & $16.92\pm0.72$ & $17.54\pm1.44$ & $-0.76\pm0.92$ \\
PEMS08 & 720 & $12.51\pm1.01$ & $10.95\pm0.50$ & $1.75\pm0.99$ \\
\midrule
All & - & $7.19\pm0.09$ & $6.77\pm0.20$ & $0.36\pm0.21$ \\
\end{longtable}
\endgroup

\subsection{What the Supporting Controls Establish}
\label{app:allocation-diagnostics}

Persistent ID outperforms Random routing and Global mixture on every
backbone average (Table~\ref{tab:allocation-diagnostics}).
These controls are informative but not fully isolated interventions:
Random changes both expert choices and weights, and Global mixture omits
the sparse-router load-balancing penalty. Dense ID also retains gains
under its own optimization rule. Donor-window and Scalar modulation
results support aligned routing inputs and multiple correction directions,
most clearly for DLinear.

The comparisons concern the stated fitting budgets, not
converged optima: 45/110 iTransformer backbone fits in the five-seed selection pool selected the
epoch cap. Different backbone recipes and normalization mean that
backbone-relative gains do not rank absolute backbone accuracy.
Validation calibration includes the uncorrected forecast, but 58/660
capacity/control combinations still have negative three-seed mean test
gains (worst $-7.66\%$). These regressions remain in all summaries.

\clearpage
\section{Backbone Adaptation and Lookback}
\label{app:functional-specialization}\label{app:observed}
This appendix expands the accuracy-resource trade-off in
Section~\ref{sec:freezing-full}, checks preservation of frozen forecasts,
and examines lookback dependence. Results use three seeds, four rank-16
experts with $K=2$, and a parameter-matched rank-64 Global correction.

\subsection{What Changes When the Backbone Is Adapted?}
\label{app:adaptation-protocol}
Frozen fitting updates only experts and routing weights, with the
backbone in evaluation mode. Joint fitting also updates the backbone,
including dropout and PatchTST BatchNorm statistics. Both use evaluation
mode for validation and test, recomputing window-normalization statistics
from each input. Corrections start from fresh input factors and zero
output factors; trained frozen corrections are not reused.

Across all three backbones, frozen and joint variants in each adaptation
comparison start from the same pretrained checkpoint for that setting
and seed.

The optimizer, calibration, ten-epoch cap, patience three, and router
regularization follow Appendix~\ref{app:allocation-training}.
Windows and seeded ordering are shared. Frozen fitting uses batch 16;
joint fitting and backbone-only fine-tuning use batch 32 on DLinear/PatchTST
and batch 16 on iTransformer for PEMS, Electricity, and Traffic, or 32
for ETT and Weather. PatchTST uses two microbatches of 16 when there are
at least 800 channels. The comparison therefore includes batch-size
and stopping effects, not only whether backbone gradients are enabled.

Joint fitting inherits the frozen expert learning rate, with router rate
ten times larger. Global, Persistent ID, and Identity + input rates are
selected over five validation seeds; Input-dependent rates use three.
Each joint method selects a backbone-rate multiplier on seed-zero
calibrated validation MSE and fixes it across reporting seeds.
Backbone-only fine-tuning selects independently. Selected multipliers
are $0.1$ or $1.0$ times the rates in Table~\ref{tab:base-rates}.

Validation selects the complete predictor's checkpoint and scale $\alpha$.
The \emph{residual-off} check sets $\alpha=0$ at that checkpoint without
refitting or reselection. Backbone-only fine-tuning is an independently
selected control for additional backbone training.

\subsection{Relative Accuracy}
\label{app:adaptation-coverage}
The four adaptation procedures cover 66 settings and three seeds.
For a predictor $A$ and reference $B$, relative improvement is
\begin{equation}
 I(A;B)=100\,\frac{\operatorname{MSE}(B)-\operatorname{MSE}(A)}{\operatorname{MSE}(B)}.
 \label{eq:adapt-relative-mse}
\end{equation}
Improvements are computed per run, averaged over 22 settings within each
seed, then summarized by mean and sample SD across seeds.
Table~\ref{tab:adaptation-all} reports aggregate accuracy and residual-off
checks; Table~\ref{tab:adaptation-cells} gives dataset-horizon comparisons.
Joint Persistent ID improves on its frozen counterpart by
$0.72/1.71/3.83\%$ for DLinear/PatchTST/iTransformer.

\begingroup\scriptsize
\setlength{\tabcolsep}{4pt}
\begin{longtable}{p{.23\textwidth}rrrrr}
\caption{Adaptation summary over three seeds. Frozen, joint, and residual-off columns give MSE above and improvement over the pretrained backbone (\%) below; parentheses denote SD in final decimal places. The final columns compare joint fitting with the corresponding frozen correction and independently fine-tuned backbone. Bold/underline mark the best/second frozen or joint result within each backbone.}\label{tab:adaptation-all}\label{tab:adaptation-main}\label{tab:adaptation-absolute}\label{tab:adaptation-coverage}\\
\toprule
Correction & Frozen & Joint & Residual off & \shortstack{Joint vs.\\frozen (\%)} & \shortstack{Joint vs.\\fine-tuned (\%)} \\
\midrule
\endfirsthead
\toprule
Correction & Frozen & Joint & Residual off & \shortstack{Joint vs.\\frozen (\%)} & \shortstack{Joint vs.\\fine-tuned (\%)} \\
\midrule
\endhead
\midrule
\multicolumn{6}{r}{Continued on next page}\\
\endfoot
\bottomrule
\endlastfoot
\multicolumn{6}{l}{\textit{DLinear}} \\*
Global correction & \shortstack{$0.4335(26)$\\$0.96(30)$} & \shortstack{$0.4368(32)$\\$0.23(11)$} & \shortstack{$0.4375(25)$\\$0.10(9)$} & $-0.85(46)$ & $0.23(11)$ \\
Persistent ID routing & \shortstack{$0.4245(38)$\\$3.57(20)$} & \shortstack{$0.4193(68)$\\$4.31(98)$} & \shortstack{$0.4421(65)$\\$-1.22(40)$} & $0.72(92)$ & $4.31(98)$ \\
Identity + input routing & \shortstack{$\mathbf{0.3860}(35)$\\$\mathbf{11.49}(22)$} & \shortstack{$\mathbf{0.3822}(32)$\\$\mathbf{11.94}(17)$} & \shortstack{$0.4839(30)$\\$-10.58(67)$} & $0.66(39)$ & $11.94(18)$ \\
Input-dependent routing & \shortstack{$\underline{0.3882}(10)$\\$\underline{10.46}(45)$} & \shortstack{$\underline{0.3884}(48)$\\$\underline{10.33}(48)$} & \shortstack{$0.4940(34)$\\$-12.37(53)$} & $0.09(102)$ & $10.33(49)$ \\
\multicolumn{6}{l}{\textit{PatchTST}} \\*
Global correction & \shortstack{$0.4014(5)$\\$5.79(5)$} & \shortstack{$0.3932(10)$\\$7.39(20)$} & \shortstack{$1.0789(82)$\\$-137.76(217)$} & $1.79(28)$ & $2.85(31)$ \\
Persistent ID routing & \shortstack{$\mathbf{0.3669}(11)$\\$\mathbf{12.41}(29)$} & \shortstack{$\mathbf{0.3605}(4)$\\$\mathbf{13.68}(9)$} & \shortstack{$1.0734(17)$\\$-137.07(69)$} & $1.71(23)$ & $9.83(19)$ \\
Identity + input routing & \shortstack{$\underline{0.3773}(11)$\\$\underline{10.36}(13)$} & \shortstack{$\underline{0.3710}(5)$\\$\underline{11.66}(3)$} & \shortstack{$1.0445(48)$\\$-130.32(129)$} & $1.52(16)$ & $7.58(13)$ \\
Input-dependent routing & \shortstack{$0.3978(9)$\\$6.46(10)$} & \shortstack{$0.3937(9)$\\$7.25(12)$} & \shortstack{$1.0597(24)$\\$-133.25(70)$} & $0.89(8)$ & $2.70(16)$ \\
\multicolumn{6}{l}{\textit{iTransformer}} \\*
Global correction & \shortstack{$0.3316(10)$\\$0.60(5)$} & \shortstack{$0.3200(5)$\\$4.11(19)$} & \shortstack{$0.3216(8)$\\$3.62(30)$} & $3.53(25)$ & $0.34(66)$ \\
Persistent ID routing & \shortstack{$\mathbf{0.3107}(10)$\\$\mathbf{7.62}(10)$} & \shortstack{$\underline{0.3004}(13)$\\$\underline{10.80}(64)$} & \shortstack{$0.3498(27)$\\$-5.59(63)$} & $3.83(69)$ & $7.64(58)$ \\
Identity + input routing & \shortstack{$\underline{0.3122}(12)$\\$\underline{7.20}(19)$} & \shortstack{$\mathbf{0.2985}(21)$\\$\mathbf{11.33}(100)$} & \shortstack{$0.3391(4)$\\$-1.97(14)$} & $4.88(141)$ & $8.21(99)$ \\
Input-dependent routing & \shortstack{$0.3320(7)$\\$0.49(3)$} & \shortstack{$0.3204(7)$\\$4.03(9)$} & \shortstack{$0.3229(4)$\\$3.24(17)$} & $3.57(10)$ & $0.26(55)$ \\
\end{longtable}
\endgroup

\begingroup\scriptsize
\setlength{\tabcolsep}{4pt}
\begin{longtable}{lrrrrr}
\caption{Joint fitting: backbone-relative MSE reduction (\%), mean$\pm$SD over three seeds. Bold/underline mark the best/second result per row. Input-dependent results appear in Table~\ref{tab:input-adaptation-results}.}\label{tab:adaptation-cells}\\
\toprule
Dataset & $H$ & \shortstack{Backbone fine-tuning\\without correction} & Global correction & \shortstack{Persistent ID\\routing} & \shortstack{Identity + input\\routing} \\
\midrule
\endfirsthead
\toprule
Dataset & $H$ & \shortstack{Backbone fine-tuning\\without correction} & Global correction & \shortstack{Persistent ID\\routing} & \shortstack{Identity + input\\routing} \\
\midrule
\endhead
\midrule
\multicolumn{6}{r}{Continued on next page}\\
\endfoot
\bottomrule
\endlastfoot
\multicolumn{6}{l}{\textit{DLinear}} \\*
ETTh1 & 96 & $-0.09\pm0.15$ & $\underline{0.00}\pm0.00$ & $\mathbf{1.54}\pm0.50$ & $-0.22\pm0.31$ \\
ETTh1 & 720 & $\underline{0.00}\pm0.00$ & $-0.06\pm0.10$ & $\mathbf{3.88}\pm4.03$ & $-0.87\pm4.74$ \\
ETTh2 & 96 & $0.18\pm0.32$ & $\underline{2.18}\pm3.25$ & $\mathbf{4.34}\pm7.51$ & $0.25\pm0.23$ \\
ETTh2 & 720 & $0.00\pm0.00$ & $\underline{3.18}\pm5.52$ & $\mathbf{8.47}\pm14.67$ & $0.00\pm0.00$ \\
ETTm1 & 96 & $0.00\pm0.00$ & $0.00\pm0.00$ & $\mathbf{5.12}\pm0.74$ & $\underline{3.28}\pm0.39$ \\
ETTm1 & 720 & $\underline{0.00}\pm0.00$ & $-0.33\pm0.58$ & $-0.33\pm0.68$ & $\mathbf{0.14}\pm0.45$ \\
ETTm2 & 96 & $\underline{0.00}\pm0.00$ & $\mathbf{0.47}\pm0.80$ & $\underline{0.00}\pm0.00$ & $\underline{0.00}\pm0.00$ \\
ETTm2 & 720 & $0.00\pm0.00$ & $0.54\pm0.93$ & $\mathbf{7.90}\pm8.66$ & $\underline{1.17}\pm2.02$ \\
Weather & 96 & $0.00\pm0.00$ & $-0.97\pm0.84$ & $\mathbf{17.66}\pm0.80$ & $\underline{11.79}\pm3.71$ \\
Weather & 720 & $0.00\pm0.00$ & $-0.02\pm0.03$ & $\mathbf{2.97}\pm2.51$ & $\underline{0.78}\pm2.87$ \\
Electricity & 96 & $0.00\pm0.00$ & $0.00\pm0.00$ & $\underline{3.01}\pm0.13$ & $\mathbf{8.71}\pm0.33$ \\
Electricity & 720 & $0.00\pm0.00$ & $0.00\pm0.00$ & $\underline{2.13}\pm0.71$ & $\mathbf{3.37}\pm0.53$ \\
Traffic & 96 & $0.00\pm0.00$ & $0.00\pm0.00$ & $\underline{1.38}\pm0.07$ & $\mathbf{9.57}\pm0.51$ \\
Traffic & 720 & $0.00\pm0.00$ & $0.00\pm0.00$ & $\underline{1.51}\pm0.27$ & $\mathbf{6.01}\pm0.42$ \\
PEMS03 & 96 & $-0.04\pm0.20$ & $0.01\pm0.11$ & $\underline{2.38}\pm0.48$ & $\mathbf{35.65}\pm0.33$ \\
PEMS03 & 720 & $0.00\pm0.00$ & $0.00\pm0.00$ & $\underline{0.99}\pm0.87$ & $\mathbf{21.89}\pm1.17$ \\
PEMS04 & 96 & $0.00\pm0.00$ & $0.00\pm0.00$ & $\underline{3.76}\pm0.18$ & $\mathbf{28.05}\pm0.74$ \\
PEMS04 & 720 & $0.00\pm0.00$ & $0.00\pm0.00$ & $\underline{2.90}\pm0.11$ & $\mathbf{18.24}\pm0.45$ \\
PEMS07 & 96 & $0.00\pm0.00$ & $0.00\pm0.00$ & $\underline{6.24}\pm0.38$ & $\mathbf{36.20}\pm0.42$ \\
PEMS07 & 720 & $0.00\pm0.00$ & $0.00\pm0.00$ & $\underline{3.70}\pm0.14$ & $\mathbf{21.25}\pm1.26$ \\
PEMS08 & 96 & $-0.02\pm0.04$ & $-0.03\pm0.03$ & $\underline{10.62}\pm0.72$ & $\mathbf{34.92}\pm1.08$ \\
PEMS08 & 720 & $-0.01\pm0.02$ & $0.00\pm0.00$ & $\underline{4.58}\pm0.11$ & $\mathbf{22.55}\pm0.19$ \\
\multicolumn{6}{l}{\textit{PatchTST}} \\*
ETTh1 & 96 & $1.81\pm0.34$ & $\mathbf{4.06}\pm0.42$ & $\underline{2.89}\pm0.81$ & $2.66\pm0.25$ \\
ETTh1 & 720 & $0.00\pm0.00$ & $0.78\pm0.69$ & $\underline{1.22}\pm0.28$ & $\mathbf{1.27}\pm0.22$ \\
ETTh2 & 96 & $0.22\pm0.39$ & $\underline{0.45}\pm0.17$ & $0.40\pm0.60$ & $\mathbf{0.51}\pm0.20$ \\
ETTh2 & 720 & $\mathbf{0.34}\pm0.38$ & $-1.57\pm1.13$ & $\underline{0.00}\pm0.00$ & $-0.34\pm0.05$ \\
ETTm1 & 96 & $0.90\pm1.56$ & $4.21\pm0.31$ & $\mathbf{5.57}\pm0.11$ & $\underline{4.50}\pm0.31$ \\
ETTm1 & 720 & $\underline{-0.07}\pm0.11$ & $-0.33\pm0.58$ & $-0.27\pm1.09$ & $\mathbf{0.04}\pm0.12$ \\
ETTm2 & 96 & $\underline{0.10}\pm0.13$ & $0.00\pm0.00$ & $\mathbf{2.16}\pm0.29$ & $0.00\pm0.00$ \\
ETTm2 & 720 & $\underline{0.00}\pm0.00$ & $\underline{0.00}\pm0.00$ & $\mathbf{0.31}\pm0.63$ & $\underline{0.00}\pm0.00$ \\
Weather & 96 & $0.16\pm0.28$ & $-0.24\pm0.35$ & $\mathbf{5.06}\pm0.64$ & $\underline{3.62}\pm0.40$ \\
Weather & 720 & $0.07\pm0.52$ & $-0.06\pm0.28$ & $\underline{1.28}\pm0.89$ & $\mathbf{1.51}\pm0.37$ \\
Electricity & 96 & $3.78\pm0.24$ & $13.06\pm0.26$ & $\mathbf{17.31}\pm0.26$ & $\underline{15.35}\pm0.60$ \\
Electricity & 720 & $2.34\pm0.17$ & $10.96\pm0.62$ & $\mathbf{13.10}\pm0.70$ & $\underline{12.22}\pm0.17$ \\
Traffic & 96 & $4.93\pm0.09$ & $\mathbf{6.91}\pm0.10$ & $3.92\pm0.66$ & $\underline{6.42}\pm0.41$ \\
Traffic & 720 & $3.68\pm0.18$ & $\mathbf{6.13}\pm0.11$ & $3.19\pm0.35$ & $\underline{5.28}\pm1.20$ \\
PEMS03 & 96 & $11.68\pm1.48$ & $20.03\pm0.16$ & $\mathbf{32.98}\pm0.25$ & $\underline{30.11}\pm1.28$ \\
PEMS03 & 720 & $9.18\pm0.78$ & $15.66\pm0.85$ & $\mathbf{25.80}\pm0.75$ & $\underline{22.47}\pm0.70$ \\
PEMS04 & 96 & $8.96\pm0.40$ & $13.63\pm0.63$ & $\mathbf{36.46}\pm0.83$ & $\underline{29.24}\pm1.28$ \\
PEMS04 & 720 & $6.30\pm1.02$ & $10.53\pm1.16$ & $\mathbf{28.58}\pm0.87$ & $\underline{22.12}\pm2.53$ \\
PEMS07 & 96 & $12.99\pm0.82$ & $13.99\pm0.70$ & $\mathbf{34.07}\pm0.59$ & $\underline{30.83}\pm1.22$ \\
PEMS07 & 720 & $9.10\pm0.47$ & $9.76\pm0.60$ & $\mathbf{24.55}\pm0.46$ & $\underline{21.12}\pm1.08$ \\
PEMS08 & 96 & $15.46\pm1.58$ & $19.82\pm1.41$ & $\mathbf{35.04}\pm0.82$ & $\underline{26.64}\pm2.47$ \\
PEMS08 & 720 & $12.62\pm0.77$ & $14.87\pm1.31$ & $\mathbf{27.35}\pm0.98$ & $\underline{20.95}\pm1.60$ \\
\multicolumn{6}{l}{\textit{iTransformer}} \\*
ETTh1 & 96 & $-0.19\pm0.48$ & $\underline{-0.02}\pm0.21$ & $\mathbf{0.34}\pm0.26$ & $-0.21\pm0.27$ \\
ETTh1 & 720 & $\underline{0.00}\pm0.00$ & $\mathbf{0.07}\pm0.16$ & $\underline{0.00}\pm0.00$ & $\underline{0.00}\pm0.00$ \\
ETTh2 & 96 & $-0.04\pm0.10$ & $\underline{0.75}\pm0.47$ & $\mathbf{1.94}\pm0.31$ & $0.69\pm0.23$ \\
ETTh2 & 720 & $0.13\pm0.01$ & $\mathbf{0.21}\pm0.26$ & $\underline{0.20}\pm0.12$ & $-0.16\pm0.74$ \\
ETTm1 & 96 & $1.25\pm0.11$ & $1.80\pm0.33$ & $\mathbf{5.05}\pm0.35$ & $\underline{4.38}\pm0.43$ \\
ETTm1 & 720 & $0.26\pm0.10$ & $0.49\pm0.08$ & $\mathbf{1.62}\pm0.59$ & $\underline{1.00}\pm0.78$ \\
ETTm2 & 96 & $0.00\pm0.00$ & $2.15\pm0.22$ & $\mathbf{4.00}\pm0.28$ & $\underline{2.71}\pm0.28$ \\
ETTm2 & 720 & $0.14\pm0.18$ & $0.31\pm0.28$ & $\mathbf{1.97}\pm0.39$ & $\underline{0.55}\pm0.12$ \\
Weather & 96 & $0.31\pm0.54$ & $0.28\pm0.48$ & $\mathbf{5.94}\pm1.54$ & $\underline{3.84}\pm0.79$ \\
Weather & 720 & $0.07\pm0.12$ & $0.20\pm0.34$ & $\mathbf{1.07}\pm0.68$ & $\underline{1.02}\pm0.45$ \\
Electricity & 96 & $1.06\pm0.93$ & $0.99\pm0.86$ & $\mathbf{5.63}\pm0.42$ & $\underline{5.41}\pm0.16$ \\
Electricity & 720 & $0.00\pm0.00$ & $0.00\pm0.00$ & $\mathbf{2.68}\pm0.52$ & $\underline{1.85}\pm0.26$ \\
Traffic & 96 & $2.60\pm0.25$ & $2.50\pm0.13$ & $\underline{2.79}\pm0.28$ & $\mathbf{3.28}\pm0.15$ \\
Traffic & 720 & $\mathbf{2.60}\pm0.18$ & $\underline{2.59}\pm0.17$ & $-0.06\pm0.95$ & $1.99\pm0.15$ \\
PEMS03 & 96 & $6.09\pm5.28$ & $9.24\pm0.99$ & $\underline{16.69}\pm14.46$ & $\mathbf{28.31}\pm0.88$ \\
PEMS03 & 720 & $6.60\pm0.87$ & $6.87\pm0.77$ & $\underline{17.28}\pm1.53$ & $\mathbf{18.19}\pm1.18$ \\
PEMS04 & 96 & $8.28\pm0.92$ & $8.83\pm1.58$ & $\underline{32.45}\pm0.79$ & $\mathbf{36.44}\pm1.65$ \\
PEMS04 & 720 & $7.66\pm0.70$ & $8.19\pm1.22$ & $\underline{25.52}\pm0.45$ & $\mathbf{26.37}\pm1.89$ \\
PEMS07 & 96 & $7.10\pm6.17$ & $3.94\pm0.35$ & $\mathbf{31.97}\pm0.93$ & $\underline{24.58}\pm21.29$ \\
PEMS07 & 720 & $11.02\pm2.15$ & $12.11\pm0.95$ & $\underline{25.37}\pm0.58$ & $\mathbf{27.60}\pm1.47$ \\
PEMS08 & 96 & $13.55\pm2.86$ & $13.97\pm1.78$ & $\underline{28.39}\pm3.91$ & $\mathbf{34.18}\pm2.19$ \\
PEMS08 & 720 & $14.15\pm1.11$ & $14.93\pm1.16$ & $\underline{26.72}\pm0.58$ & $\mathbf{27.13}\pm1.52$ \\
\end{longtable}
\endgroup

Input-dependent joint fitting improves its own pretrained backbone by
$10.33\pm0.48\%$, $7.25\pm0.12\%$, and $4.03\pm0.09\%$ for
DLinear, PatchTST, and iTransformer. Relative to the corresponding
frozen correction, the reductions are $0.09\pm1.02\%$,
$0.89\pm0.08\%$, and $3.57\pm0.10\%$. On iTransformer, the joint fit is within
seed variation of backbone-only fine-tuning ($0.26\pm0.55\%$).
Table~\ref{tab:input-adaptation-results} reports joint and residual-off
MSE; Table~\ref{tab:adaptation-all} summarizes the
joint-versus-fine-tuned contrasts.

\begin{table}[htbp]
\centering\scriptsize
\setlength{\tabcolsep}{4pt}
\caption{Input-dependent adaptation: joint and residual-off test MSE, mean(SD in final decimal places) over three seeds. Residual-off disables the correction at the selected joint checkpoint without refitting.}
\label{tab:input-adaptation-results}
\begin{tabular}{llrrrrrr}
\toprule
& & \multicolumn{2}{c}{DLinear} & \multicolumn{2}{c}{PatchTST} & \multicolumn{2}{c}{iTransformer} \\
\cmidrule(lr){3-4}\cmidrule(lr){5-6}\cmidrule(lr){7-8}
Dataset & $H$ & Joint & Residual off & Joint & Residual off & Joint & Residual off \\
\midrule
ETTh1 & 96 & $0.3868(29)$ & $0.3849(22)$ & $0.3923(32)$ & $0.6531(35)$ & $0.3922(7)$ & $0.3882(8)$ \\
ETTh1 & 720 & $0.5586(358)$ & $0.5403(295)$ & $0.4768(87)$ & $0.6538(101)$ & $0.5004(43)$ & $0.4984(44)$ \\
ETTh2 & 96 & $0.3193(407)$ & $0.3224(389)$ & $0.2929(32)$ & $0.3084(34)$ & $0.3001(16)$ & $0.2976(14)$ \\
ETTh2 & 720 & $0.7678(784)$ & $0.8111(1123)$ & $0.4294(69)$ & $0.4308(43)$ & $0.4280(13)$ & $0.4282(14)$ \\
ETTm1 & 96 & $0.3438(14)$ & $0.3438(14)$ & $0.3267(13)$ & $0.6019(59)$ & $0.3359(32)$ & $0.3377(40)$ \\
ETTm1 & 720 & $0.4733(44)$ & $0.4728(47)$ & $0.4582(12)$ & $0.4582(12)$ & $0.4894(21)$ & $0.4858(18)$ \\
ETTm2 & 96 & $0.1855(32)$ & $0.1855(32)$ & $0.1778(2)$ & $0.1778(2)$ & $0.1833(10)$ & $0.1842(5)$ \\
ETTm2 & 720 & $0.4816(439)$ & $0.4908(357)$ & $0.4010(8)$ & $0.4010(8)$ & $0.4135(15)$ & $0.4135(15)$ \\
Weather & 96 & $0.1878(26)$ & $0.2442(174)$ & $0.1729(9)$ & $0.2127(7)$ & $0.1770(18)$ & $0.1770(18)$ \\
Weather & 720 & $0.3649(116)$ & $0.4876(275)$ & $0.3513(18)$ & $0.3513(18)$ & $0.3597(16)$ & $0.3596(16)$ \\
Electricity & 96 & $0.1853(4)$ & $0.1975(3)$ & $0.1723(2)$ & $0.8194(8)$ & $0.1475(1)$ & $0.1475(1)$ \\
Electricity & 720 & $0.2375(9)$ & $0.2457(1)$ & $0.2380(15)$ & $0.8693(5)$ & $0.2121(3)$ & $0.2121(3)$ \\
Traffic & 96 & $0.5812(37)$ & $0.6651(39)$ & $0.4619(13)$ & $1.3367(10)$ & $0.3974(2)$ & $0.3978(1)$ \\
Traffic & 720 & $0.6084(28)$ & $0.6598(7)$ & $0.5187(1)$ & $1.3519(16)$ & $0.4642(11)$ & $0.4652(8)$ \\
PEMS03 & 96 & $0.2902(31)$ & $0.5035(319)$ & $0.3926(95)$ & $1.9696(40)$ & $0.2530(38)$ & $0.2532(39)$ \\
PEMS03 & 720 & $0.3324(29)$ & $0.5317(78)$ & $0.4344(29)$ & $1.6263(28)$ & $0.3086(6)$ & $0.3100(17)$ \\
PEMS04 & 96 & $0.3048(11)$ & $0.5271(150)$ & $0.5389(50)$ & $2.0616(206)$ & $0.2772(20)$ & $0.2775(22)$ \\
PEMS04 & 720 & $0.3599(17)$ & $0.4915(102)$ & $0.5495(41)$ & $1.7188(41)$ & $0.3106(40)$ & $0.3116(39)$ \\
PEMS07 & 96 & $0.3797(23)$ & $0.7212(487)$ & $0.4306(41)$ & $1.8907(148)$ & $0.2071(11)$ & $0.2172(42)$ \\
PEMS07 & 720 & $0.4351(29)$ & $0.6752(301)$ & $0.4634(40)$ & $1.5714(17)$ & $0.2522(40)$ & $0.2625(57)$ \\
PEMS08 & 96 & $0.3340(53)$ & $0.7454(380)$ & $0.4484(22)$ & $2.0661(115)$ & $0.2802(26)$ & $0.2935(69)$ \\
PEMS08 & 720 & $0.4266(65)$ & $0.6212(125)$ & $0.5342(61)$ & $1.7829(24)$ & $0.3586(35)$ & $0.3846(112)$ \\
\bottomrule
\end{tabular}
\end{table}

\subsection{Fitting Costs of Backbone Adaptation}
\label{app:adaptation-resources}
For Persistent ID routing, joint fitting increases the median
gradient-trained parameter count from $40.18$ to $119.33$ thousand
on DLinear, $106.89$ to $654.18$ thousand on PatchTST, and
$48.37$ to $4873.03$ thousand on iTransformer
(22 settings per backbone; calibration and fixed labels excluded).

Table~\ref{tab:adaptation-fitcost} measures time and memory on Electricity,
PEMS04, ETTm1, and Weather at $L=96$, $H\in\{96,720\}$:
24 settings, three seeds, and 144 independent fits. Frozen/joint pairs
start from the same checkpoint and fresh correction and run back-to-back
on an exclusive RTX 2080 Ti, without retuning. Frozen fitting uses batch 16;
joint fitting uses 32 for DLinear/PatchTST and 16 for iTransformer.
Times include setup, training, validation, calibration, and checkpoint
restoration, but exclude pretraining, search, test evaluation, verification,
and serialization. Peak memory covers training, not validation or test.

These separately timed fits select different epochs from the batched
learning-rate search in 48/144 cases, with maximum relative test-MSE
difference $2.15\%$. The table therefore reports their own accuracy;
Table~\ref{tab:adaptation-all} remains the full accuracy comparison.

Taking the geometric mean of the three paired joint/frozen ratios per
setting, then the median over eight settings per backbone, gives time
ratios $0.70$, $2.75$, and $2.53$, and training-memory ratios
$1.35$, $7.35$, and $2.85$ for DLinear, PatchTST, and iTransformer.
Joint fitting is faster on DLinear; freezing lowers peak training memory
in every measured setting. These costs include batch-size and stopping effects.

\begin{table}[htbp]
\centering\scriptsize
\setlength{\tabcolsep}{3pt}
\caption{Frozen/joint Persistent ID fitting on 24 settings. Times (seconds), joint/frozen time ratios, and joint-over-frozen MSE reduction (\%) are mean$\pm$SD over three seeds. Memory gives frozen/joint mean peaks in GiB (SD rounds to zero); ratios are joint/frozen. Arithmetic mean ratios here differ from the geometric aggregation in the text.}
\label{tab:adaptation-fitcost}
\begin{tabular}{llrrrrrr}
\toprule
Backbone & Dataset & $H$ & Fit frozen / joint & Time ratio & Peak frozen / joint & Memory ratio & $\Delta$MSE (\%) \\
\midrule
DLinear & ETTm1 & 96 & $107\pm30$ / $84\pm31$ & $0.86\pm0.44 $ & $0.020$ / $0.021$ & $1.02$ & $0.41\pm0.20 $ \\
DLinear & ETTm1 & 720 & $116\pm33$ / $53\pm7$ & $0.47\pm0.08 $ & $0.024$ / $0.030$ & $1.24$ & $-2.58\pm0.45 $ \\
DLinear & Electricity & 96 & $69\pm38$ / $59\pm3$ & $1.08\pm0.68 $ & $0.081$ / $0.111$ & $1.37$ & $0.30\pm0.15 $ \\
DLinear & Electricity & 720 & $93\pm5$ / $44\pm10$ & $0.47\pm0.12 $ & $0.199$ / $0.324$ & $1.62$ & $-0.04\pm0.91 $ \\
DLinear & PEMS04 & 96 & $50\pm14$ / $35\pm6$ & $0.74\pm0.33 $ & $0.067$ / $0.094$ & $1.40$ & $0.12\pm0.28 $ \\
DLinear & PEMS04 & 720 & $55\pm5$ / $34\pm4$ & $0.63\pm0.12 $ & $0.181$ / $0.299$ & $1.65$ & $-0.10\pm0.20 $ \\
DLinear & Weather & 96 & $165\pm29$ / $114\pm7$ & $0.71\pm0.15 $ & $0.023$ / $0.025$ & $1.06$ & $1.12\pm0.86 $ \\
DLinear & Weather & 720 & $116\pm16$ / $83\pm20$ & $0.73\pm0.24 $ & $0.032$ / $0.042$ & $1.33$ & $-2.51\pm2.96 $ \\
\addlinespace
PatchTST & ETTm1 & 96 & $153\pm56$ / $367\pm47$ & $2.58\pm0.70 $ & $0.032$ / $0.103$ & $3.20$ & $-0.50\pm0.17 $ \\
PatchTST & ETTm1 & 720 & $94\pm9$ / $194\pm3$ & $2.09\pm0.23 $ & $0.038$ / $0.123$ & $3.27$ & $-0.79\pm0.79 $ \\
PatchTST & Electricity & 96 & $371\pm22$ / $1178\pm6$ & $3.18\pm0.20 $ & $0.410$ / $4.526$ & $11.04$ & $1.70\pm0.03 $ \\
PatchTST & Electricity & 720 & $318\pm41$ / $1163\pm10$ & $3.71\pm0.54 $ & $0.472$ / $4.635$ & $9.82$ & $1.50\pm0.47 $ \\
PatchTST & PEMS04 & 96 & $198\pm50$ / $678\pm39$ & $3.64\pm1.27 $ & $0.382$ / $4.330$ & $11.34$ & $6.03\pm1.70 $ \\
PatchTST & PEMS04 & 720 & $210\pm13$ / $626\pm5$ & $2.98\pm0.17 $ & $0.444$ / $4.438$ & $10.00$ & $2.44\pm1.44 $ \\
PatchTST & Weather & 96 & $192\pm32$ / $361\pm62$ & $1.93\pm0.54 $ & $0.051$ / $0.249$ & $4.88$ & $0.49\pm0.42 $ \\
PatchTST & Weather & 720 & $177\pm43$ / $290\pm43$ & $1.72\pm0.54 $ & $0.059$ / $0.272$ & $4.62$ & $0.35\pm0.48 $ \\
\addlinespace
iTransformer & ETTm1 & 96 & $144\pm45$ / $148\pm30$ & $1.07\pm0.27 $ & $0.023$ / $0.032$ & $1.36$ & $2.32\pm0.30 $ \\
iTransformer & ETTm1 & 720 & $139\pm28$ / $250\pm46$ & $1.89\pm0.73 $ & $0.026$ / $0.038$ & $1.46$ & $0.45\pm0.72 $ \\
iTransformer & Electricity & 96 & $118\pm19$ / $385\pm44$ & $3.34\pm0.82 $ & $0.160$ / $0.522$ & $3.26$ & $2.40\pm0.47 $ \\
iTransformer & Electricity & 720 & $97\pm33$ / $418\pm4$ & $4.66\pm1.54 $ & $0.239$ / $0.627$ & $2.62$ & $0.60\pm0.47 $ \\
iTransformer & PEMS04 & 96 & $70\pm4$ / $269\pm21$ & $3.85\pm0.52 $ & $0.144$ / $0.491$ & $3.40$ & $12.63\pm0.80 $ \\
iTransformer & PEMS04 & 720 & $63\pm4$ / $246\pm1$ & $3.91\pm0.28 $ & $0.218$ / $0.593$ & $2.72$ & $8.81\pm0.98 $ \\
iTransformer & Weather & 96 & $182\pm27$ / $185\pm37$ & $1.05\pm0.33 $ & $0.048$ / $0.143$ & $2.98$ & $-1.50\pm1.26 $ \\
iTransformer & Weather & 720 & $205\pm7$ / $148\pm18$ & $0.72\pm0.11 $ & $0.053$ / $0.164$ & $3.06$ & $-0.99\pm0.36 $ \\
\addlinespace
\bottomrule
\end{tabular}
\end{table}

\subsection{Does Freezing Preserve the Original Forecast?}
\label{app:preservation}\label{sec:shared-preservation}
Joint adaptation can improve the full predictor while making its backbone
depend on the correction: disabling the correction can give worse MSE
than the pretrained model (Table~\ref{tab:adaptation-all}).
Freezing instead preserves parameters and buffers, so $\alpha=0$
recovers the original forecast. This is confirmed in all 330 tested frozen
settings and all 72 resource-study fits. Relative pretrained-MSE
discrepancies are at most $1.90\times10^{-9}$ for recomputed validation
errors and $3.55\times10^{-9}$ in the adaptation comparisons.

On the first 256 ETTh1 test windows ($H=96$, seed zero, three backbones,
five correction variants), both fresh corrections at $\alpha=1$ and
trained corrections at $\alpha=0$ give zero maximum absolute output
difference. Across 33 saved-checkpoint checks, the largest relative
discrepancy is $1.88\times10^{-6}$, below tolerance $10^{-5}$.
These checks do not establish bitwise equality on every window.

\subsection{Lookback Dependence}
\label{app:lookback-protocol}\label{sec:lookback}
We compare $L=96$ and $L=720$ on PEMS04, Electricity, and Traffic at
$H\in\{96,720\}$, giving twelve settings and three reporting seeds.
All iTransformer routing comparisons share the same pretrained checkpoint
within each lookback setting and seed.
Expert banks have four rank-16 experts, two active; Ridge-based routing
uses the same equally weighted own-group/nearest-group rule.

The $L=96$ results are those of the allocation study. At both lookbacks,
iTransformer uses model/feed-forward widths 512, three
encoder layers, eight attention heads, dropout 0.1, four time-mark tokens,
and non-affine RevIN. Pretraining uses Adam at $5\times10^{-4}$,
batch 16, epoch-wise rate halving, a ten-epoch cap, and patience three.
Correction fitting follows Appendix~\ref{app:allocation-training},
including five-seed learning-rate selection for the allocation-study methods;
Input-dependent routing selects over three seeds at each lookback.

Validation and test targets are identical across lookbacks.
However, $L=720$ has 624 fewer training origins and adds
$(720-96)\times512=319{,}488$ backbone parameters; correction-feature
width stays $F=512$. This tests lookback dependence under a common recipe,
not history length at fixed sample count and backbone capacity.

Table~\ref{tab:lookback-top2} reports same-lookback comparisons.
On PEMS04, Persistent ID's improvement falls from $23.24\%$ to $4.82\%$
at $H=96$ and from $18.74\%$ to $5.26\%$ at $H=720$. Electricity gains
are smaller, and Traffic gains remain below $1\%$.

\begingroup\scriptsize
\setlength{\tabcolsep}{4pt}
\begin{longtable}{lrrrrrrr}
\caption{Frozen iTransformer at two lookbacks: MSE above and same-lookback backbone-relative reduction (\%) below, mean(SD in final decimal places) over three seeds. All bank-based routers use $E=4,K=2$; Global correction matches factor count. Bold/underline mark the best/second result per row and metric.}\label{tab:lookback-top2}\\
\toprule
Dataset & $H$ & $L$ & \shortstack{Global\\correction} & \shortstack{Random\\routing} & \shortstack{Ridge-based\\routing} & \shortstack{Persistent ID\\routing} & \shortstack{Identity + input\\routing} \\
\midrule
\endfirsthead
\toprule
Dataset & $H$ & $L$ & \shortstack{Global\\correction} & \shortstack{Random\\routing} & \shortstack{Ridge-based\\routing} & \shortstack{Persistent ID\\routing} & \shortstack{Identity + input\\routing} \\
\midrule
\endhead
\midrule
\multicolumn{8}{r}{Continued on next page}\\
\endfoot
\bottomrule
\endlastfoot
PEMS04 & 96 & 96 & \shortstack{$0.3014(61)$\\$0.90(1)$} & \shortstack{$0.2985(58)$\\$1.85(16)$} & \shortstack{$0.2887(56)$\\$5.05(12)$} & \shortstack{$\underline{0.2334}(42)$\\$\underline{23.24}(24)$} & \shortstack{$\mathbf{0.2308}(71)$\\$\mathbf{24.12}(87)$} \\
PEMS04 & 96 & 720 & \shortstack{$0.1083(13)$\\$0.78(22)$} & \shortstack{$0.1082(13)$\\$0.82(20)$} & \shortstack{$0.1075(14)$\\$1.51(17)$} & \shortstack{$\mathbf{0.1038}(15)$\\$\mathbf{4.82}(8)$} & \shortstack{$\underline{0.1047}(15)$\\$\underline{4.06}(22)$} \\
PEMS04 & 720 & 96 & \shortstack{$0.3367(60)$\\$0.75(27)$} & \shortstack{$0.3352(58)$\\$1.18(29)$} & \shortstack{$0.3232(57)$\\$4.72(16)$} & \shortstack{$\mathbf{0.2757}(63)$\\$\mathbf{18.74}(54)$} & \shortstack{$\underline{0.2785}(96)$\\$\underline{17.93}(129)$} \\
PEMS04 & 720 & 720 & \shortstack{$0.1488(33)$\\$0.39(30)$} & \shortstack{$0.1491(33)$\\$0.24(33)$} & \shortstack{$0.1473(32)$\\$1.43(29)$} & \shortstack{$\mathbf{0.1416}(32)$\\$\mathbf{5.26}(37)$} & \shortstack{$\underline{0.1428}(31)$\\$\underline{4.46}(45)$} \\
Electricity & 96 & 96 & \shortstack{$0.1491(2)$\\$0.50(8)$} & \shortstack{$0.1489(2)$\\$0.65(9)$} & \shortstack{$0.1471(1)$\\$1.89(2)$} & \shortstack{$\underline{0.1450}(3)$\\$\underline{3.29}(28)$} & \shortstack{$\mathbf{0.1449}(6)$\\$\mathbf{3.30}(36)$} \\
Electricity & 96 & 720 & \shortstack{$\underline{0.1340}(17)$\\$\underline{1.56}(18)$} & \shortstack{$0.1341(17)$\\$1.43(19)$} & \shortstack{$\mathbf{0.1339}(17)$\\$\mathbf{1.57}(20)$} & \shortstack{$0.1341(17)$\\$1.43(20)$} & \shortstack{$\underline{0.1340}(18)$\\$1.54(10)$} \\
Electricity & 720 & 96 & \shortstack{$0.2113(2)$\\$0.39(19)$} & \shortstack{$0.2111(3)$\\$0.46(19)$} & \shortstack{$0.2093(2)$\\$1.32(15)$} & \shortstack{$\underline{0.2075}(3)$\\$\underline{2.17}(15)$} & \shortstack{$\mathbf{0.2073}(3)$\\$\mathbf{2.24}(3)$} \\
Electricity & 720 & 720 & \shortstack{$0.1929(17)$\\$0.86(15)$} & \shortstack{$0.1929(16)$\\$0.85(18)$} & \shortstack{$0.1929(17)$\\$0.85(15)$} & \shortstack{$\underline{0.1922}(19)$\\$\underline{1.19}(35)$} & \shortstack{$\mathbf{0.1908}(20)$\\$\mathbf{1.93}(23)$} \\
Traffic & 96 & 96 & \shortstack{$0.4066(8)$\\$0.39(4)$} & \shortstack{$0.4070(6)$\\$0.29(13)$} & \shortstack{$0.4071(9)$\\$0.27(2)$} & \shortstack{$\underline{0.4058}(10)$\\$\underline{0.58}(7)$} & \shortstack{$\mathbf{0.4055}(13)$\\$\mathbf{0.66}(18)$} \\
Traffic & 96 & 720 & \shortstack{$\mathbf{0.3585}(18)$\\$\mathbf{0.44}(4)$} & \shortstack{$\underline{0.3588}(18)$\\$\underline{0.35}(4)$} & \shortstack{$\underline{0.3588}(17)$\\$\underline{0.35}(5)$} & \shortstack{$0.3593(18)$\\$0.22(3)$} & \shortstack{$0.3589(20)$\\$0.33(3)$} \\
Traffic & 720 & 96 & \shortstack{$\mathbf{0.4754}(3)$\\$\mathbf{0.29}(1)$} & \shortstack{$0.4761(2)$\\$0.14(4)$} & \shortstack{$0.4770(3)$\\$-0.06(2)$} & \shortstack{$\underline{0.4757}(15)$\\$0.21(27)$} & \shortstack{$\mathbf{0.4754}(3)$\\$\underline{0.27}(10)$} \\
Traffic & 720 & 720 & \shortstack{$\mathbf{0.4293}(11)$\\$\mathbf{0.33}(14)$} & \shortstack{$0.4300(11)$\\$0.16(10)$} & \shortstack{$\underline{0.4299}(12)$\\$\underline{0.20}(13)$} & \shortstack{$0.4318(16)$\\$-0.25(21)$} & \shortstack{$0.4301(8)$\\$0.14(1)$} \\
\end{longtable}
\endgroup

Input-dependent gains on iTransformer are modest at both lookbacks:
$0.29$-$0.99\%$ at $L=96$ and $0.27$-$1.45\%$ at $L=720$
(Table~\ref{tab:input-lookback-results}). Longer history reduces backbone
error, but the correction's additional benefit increases on Electricity
and decreases on PEMS04 and Traffic. Inference measurements cover all
twelve settings (Appendix~\ref{app:lookback-cost}).

\begingroup\small
\setlength{\tabcolsep}{4pt}
\begin{longtable}{lrrlll}
\caption{Input-dependent lookback results for iTransformer, mean$\pm$SD over three seeds. Improvements use each row's pretrained backbone; the checkpoint is matched to the other routing methods in Table~\ref{tab:lookback-top2} at the same setting and seed. Bold marks the lower MSE in each row.}\label{tab:input-lookback-results}\\
\toprule
Dataset & $H$ & $L$ & \shortstack{Pretrained\\backbone MSE} & \shortstack{Input-dependent\\routing MSE} & Improvement (\%) \\
\midrule
\endfirsthead
\multicolumn{6}{l}{\tablename~\thetable\ (continued)}\\
\toprule
Dataset & $H$ & $L$ & \shortstack{Pretrained\\backbone MSE} & \shortstack{Input-dependent\\routing MSE} & Improvement (\%) \\
\midrule
\endhead
\midrule
\multicolumn{6}{r}{Continued on next page}\\
\endfoot
\bottomrule
\endlastfoot
PEMS04 & 96 & 96 & ${0.3041}\pm0.0062$ & $\mathbf{0.3013}\pm0.0060$ & $0.92\pm0.05$ \\
PEMS04 & 96 & 720 & ${0.1091}\pm0.0015$ & $\mathbf{0.1084}\pm0.0013$ & $0.68\pm0.18$ \\
PEMS04 & 720 & 96 & ${0.3392}\pm0.0065$ & $\mathbf{0.3359}\pm0.0062$ & $0.99\pm0.30$ \\
PEMS04 & 720 & 720 & ${0.1494}\pm0.0029$ & $\mathbf{0.1489}\pm0.0031$ & $0.35\pm0.12$ \\
Electricity & 96 & 96 & ${0.1499}\pm0.0001$ & $\mathbf{0.1492}\pm0.0002$ & $0.45\pm0.07$ \\
Electricity & 96 & 720 & ${0.1361}\pm0.0020$ & $\mathbf{0.1341}\pm0.0017$ & $1.45\pm0.20$ \\
Electricity & 720 & 96 & ${0.2121}\pm0.0003$ & $\mathbf{0.2115}\pm0.0002$ & $0.29\pm0.19$ \\
Electricity & 720 & 720 & ${0.1945}\pm0.0017$ & $\mathbf{0.1934}\pm0.0016$ & $0.58\pm0.03$ \\
Traffic & 96 & 96 & ${0.4082}\pm0.0009$ & $\mathbf{0.4065}\pm0.0008$ & $0.42\pm0.04$ \\
Traffic & 96 & 720 & ${0.3601}\pm0.0019$ & $\mathbf{0.3589}\pm0.0020$ & $0.32\pm0.05$ \\
Traffic & 720 & 96 & ${0.4767}\pm0.0002$ & $\mathbf{0.4753}\pm0.0004$ & $0.29\pm0.03$ \\
Traffic & 720 & 720 & ${0.4308}\pm0.0009$ & $\mathbf{0.4296}\pm0.0010$ & $0.27\pm0.07$ \\
\end{longtable}
\endgroup
\clearpage
\section{When Should Specialization Help?}
\label{app:synthetic}

This appendix supports Section~\ref{sec:synthetic}: how GalaxyTS is
generated, how its exact forecasting references quantify the opportunity
for specialization, and which allocations improve on matched Global corrections.
The backbone-feature results come first; supplementary raw-window checks
follow. Generating-family labels A-D record origin and stay fixed as
separation changes; they need not be optimal expert assignments.

\subsection{Generating GalaxyTS}
\label{app:synthetic-generation}

\begin{figure}[htbp]
\centering
\includegraphics[width=0.8\textwidth]{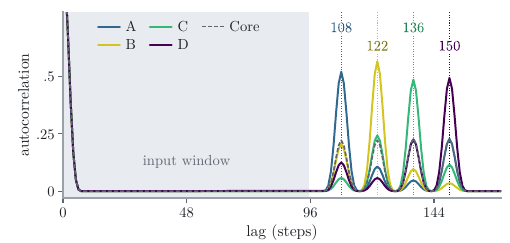}
\caption{\textbf{Autocorrelation of the illustrated channels.} Exact profiles for the five channels highlighted in Figure~\ref{fig:toy-hidden} at $s=1$. The shaded region marks lags within the input window, where the channels share the same autocorrelation; channel-specific echo peaks occur at longer lags.}
\label{fig:synthetic-autocorrelation}
\end{figure}

\paragraph{Echo processes.}
Each channel $c$ combines four independent Gaussian echo components,
\begin{align}
 x_c(t)&=\sum_{k=1}^{4}\sqrt{w_{ck}(s)}\,y_{ck}(t),\qquad
 y_{ck}(t)=\varphi(s)\,y_{ck}(t-P_k)+\sqrt{1-\varphi(s)^2}\,u_{ck}(t),
 \label{eq:hidden-echo-process}\\
 u_{ck}(t)&=\sum_{j=0}^{6}h_j\varepsilon_{ck}(t-j),\qquad
 \varepsilon_{ck}(t)\overset{\mathrm{iid}}{\sim}\mathcal N(0,1),\nonumber
\end{align}
with delays $(P_1,\dots,P_4)=(108,122,136,150)$ and echo strength
$\varphi(s)=0.5+0.4s$. The weights satisfy $\sum_kw_{ck}(s)=1$.
Fresh noise is smoothed by the Hann kernel
$h_j\propto1-\cos(2\pi(j+1)/8)$, normalized so $\sum_jh_j^2=1$.
Every channel is stationary with unit variance and autocovariance
\begin{equation}
 \gamma_c(\tau;s)=\sum_{k=1}^{4}w_{ck}(s)\sum_{m\in\mathbb Z}\varphi(s)^{|m|}\rho(\tau-mP_k),
 \qquad \rho(\tau)=\sum_jh_jh_{j+|\tau|},
 \label{eq:hidden-echo-covariance}
\end{equation}
where $\rho$ vanishes beyond lag 6. Each echo links a future value to
the channel's own past at a fixed delay; the echo mix determines the
forecasting rule. Gaussianity makes the optimal linear forecasts exact.
All delays exceed the 96-step input window: for $|\tau|\le95$,
$\gamma_c(\tau;s)=\rho(\tau)$. Thus channels share local window statistics
while differing in how their past predicts their future.
The autocorrelation peak at $P_k$ has height $\varphi(s)w_{ck}(s)$,
revealing the corresponding echo weight
(Figure~\ref{fig:synthetic-autocorrelation}).

\paragraph{Layout and family overlap.}
A channel's planar position $z_c(s)$ sets its echo mix,
\begin{equation}
 w_{ck}(s)=\frac{\exp\{2.6\,a_k^\top z_c(s)\}}{\sum_{\ell=1}^{4}\exp\{2.6\,a_\ell^\top z_c(s)\}},
 \qquad a_k=(\cos\theta_{k-1},\sin\theta_{k-1})^\top,\quad \theta_g=g\pi/2,
 \label{eq:hidden-echo-weights}
\end{equation}
so direction selects the dominant echo and distance from the center
sets how strongly it dominates. Families A-D have 128 channels each,
home delays 108, 122, 136, and 150, and spiral-arm directions $\theta_g$.
With layout seed 2027, each arm has six beads spaced uniformly on
$[0.14,0.95]$, with jitter SD 0.012. Channels join the shared core with
probability 0.07 (40 channels). Otherwise, arm position $v_c$ is a random
bead plus noise of SD 0.017 with probability 0.72, or a uniform draw on
$[0.05,1]$, clipped to $[0,1]$. For fixed standard normal
$\xi_c,\zeta_c$ and arm phase $\psi_g\sim\mathrm U(0,2\pi)$,
\begin{align}
 r_c(s)&=(0.82+0.18s)(0.14+0.86v_c+0.010\xi_c),\qquad
 \theta_c(s)=\theta_g+(1.6-1.2s)\pi\,(v_c-0.62),\nonumber\\
 b_c(s)&=0.034\sin(2\pi\,1.7v_c+\psi_g)+[1+1.4(1-s)](0.010+0.024v_c)\,\zeta_c,
 \label{eq:hidden-echo-layout}
\end{align}
and $z_c=r_c(\cos\theta_c,\sin\theta_c)^\top+b_cn_c$, where $n_c$ is the
arm's unit normal at $v_c$. Core points are fixed
$\mathcal N(0,0.075^2I)$ draws scaled by $1+0.6(1-s)$.
Figure~\ref{fig:synthetic-galaxy} uses additional draws to illustrate the
arm geometry; experiments use only the 512 channels.

Variation along and across each arm, together with balanced mixes in the
core, makes families an imperfect grouping of forecasting rules.
Increasing $s$ unwinds the arms and strengthens the echoes; mean
home-echo weight rises from about 0.32 to 0.59. Families interleave at
$s=0$ and still overlap at $s=1$. Because separation and predictability
change together, allocations are compared with exact references at the
same $s$. GalaxyTS models graded predictive differences, not any particular
real dataset.

\paragraph{Realizations and splits.}
For each $s\in\{0,0.33,0.66,1\}$ and data seeds 1-3, we generate
12,000 steps after a burn-in of
$150\lceil\log(10^{-6})/\log\varphi(s)\rceil+6$ steps, rounding to five decimals.
Training, validation, and test use rows $[0,8400)$,
$[8400-L,9600)$, and $[9600-L,12000)$, respectively.
Channels are standardized using training statistics; targets never cross
split boundaries. Shared predictors receive neither timestamps nor channel identity.

\subsection{Exact Forecasting References}
\label{app:synthetic-references}
At $L=96$, every channel has the same input covariance $\Sigma_X$.
Channel $c$'s optimal map is $M_c=C_c\Sigma_X^{-1}$, with
$C_c=\operatorname{Cov}(Y_c,X_c)$. An optimal shared or group-specific
map uses the corresponding mean $C_c$. Family grouping is therefore
a reference, not an upper bound over other partitions.

Using map $B$ instead of $M_c$ adds
$\|(B-M_c)\Sigma_X^{1/2}\|_F^2/H$ to channel $c$'s MSE.
Consequently, squared distances between the whitened forecasting maps
$M_c\Sigma_X^{1/2}$ measure excess forecasting error. Their mean squared
spread per forecast step equals the specialization gap; the between-group
share of this spread is the fraction closed by one exact map per group
(Table~\ref{tab:hidden-echo-exact}).

Population errors follow from the known covariances. We also evaluate
these exact maps on the same test windows as trained models.
The best-found four-group reference takes the lowest within-group sum
of squares over 200 $k$-means starts on the whitened maps; it is not a
certified global optimum.

\begin{table}[htbp]
\centering\small
\setlength{\tabcolsep}{6pt}
\caption{Exact population references at $L=H=96$. The lower block gives
the specialization gap closed by one exact map per group (\%; Ridge
averages realizations). Bold/underline mark the best/second distinct
grouping values at each separation; ties share rank. Population errors
in the upper block are unranked.}
\label{tab:hidden-echo-exact}
\begin{tabular}{@{}lcccc@{}}
\toprule
 & $s=0$ & $0.33$ & $0.66$ & $1$ \\
\midrule
Optimal shared MSE & 0.9257 & 0.9000 & 0.8683 & 0.8297 \\
Optimal per-channel MSE & 0.8997 & 0.8561 & 0.8030 & 0.7127 \\
Specialization gap (\%) & 2.81 & 4.87 & 7.52 & 14.10 \\
\midrule
\multicolumn{5}{@{}l}{\emph{Gap closed by one exact map per group (\%)}} \\
Family labels & 9.4 & 23.3 & 52.6 & 77.4 \\
Random groups & 0.5 & 0.4 & 0.4 & 0.5 \\
Dominant echo & 69.7 & \underline{74.7} & \underline{71.2} & \underline{80.0} \\
Ridge groups & \underline{70.4} & 74.1 & 70.8 & 79.4 \\
Best four-group split found & \textbf{71.5} & \textbf{74.8} & \textbf{71.8} & \textbf{80.4} \\
\bottomrule
\end{tabular}
\end{table}

\subsection{Allocation Protocol and Reference-Scaled Gains}
\label{app:synthetic-protocol}

The allocation studies use $L=96$, $H\in\{96,192,336,720\}$,
four separation levels, three backbones, and three realizations.
Section~\ref{sec:synthetic} uses the frozen backbone's prediction-head
features; raw-window corrections provide a supplementary check.
Table~\ref{tab:synthetic-design} distinguishes their ranks and training seeds.
Each study fits its own parameter-matched Global correction.
Matched two-expert Random and Ridge-based controls use
$s\in\{0,1\}$ and $H\in\{96,720\}$, giving twelve settings.

\begin{table}[htbp]
\centering\small
\setlength{\tabcolsep}{5pt}
\caption{GalaxyTS allocation protocols. Rank is per expert; Global
gives the shared correction's factor width. Seeds are per realization.}
\label{tab:synthetic-design}
\begin{tabular}{@{}p{0.25\textwidth}p{0.23\textwidth}lll@{}}
\toprule
Evaluation & Correction input & Rank / Global & Seeds & Fitting rule\\
\midrule
Backbone-feature allocation & Prediction-head features & 16 / 64 & 0,1,2 & Appendix~\ref{app:allocation-training}\\
Raw-window check & Channel history & 32 / 128 & 0 & Fixed-rate Adam\\
\bottomrule
\end{tabular}
\end{table}

\paragraph{Fitting backbone-feature corrections.}
Backbones are trained independently for each seed within each realization. Correction fitting follows
Appendix~\ref{app:allocation-training}, including sparse-router load balancing.
Expert rates are selected from
$\{1.5\times10^{-4},5\times10^{-4},1.5\times10^{-3},5\times10^{-3}\}$
by mean calibrated validation MSE across the three training seeds within
each realization.
\paragraph{Reference-scaled gain over Global.}
We normalize each variant's MSE improvement over its matched
Global correction by the exact shared-to-channel-specific
reference gap:
\begin{equation}
 G=100\,\frac{\mathrm{MSE}_{\mathrm{global\ corr.}}-\mathrm{MSE}_{\mathrm{variant}}}
 {\mathrm{MSE}^{\star}_{\mathrm{shared}}-\mathrm{MSE}^{\star}_{\mathrm{channel}}},
 \label{eq:synthetic-gap}
\end{equation}
The starred errors are obtained by evaluating the exact reference
maps on the same test windows. Because the numerator starts from
fitted Global rather than the exact shared reference, $G$ does not
measure the fraction of the distance to either optimum.
In particular, $G=100\%$ need not imply attaining the
channel-specific reference error.
Relative MSE gain against a reference is $100(1-m/m_{\rm ref})$
for variant error $m$ and reference error $m_{\rm ref}$.
Gains and $G$ average seeds within realizations, then weight
realizations equally. At $s=0$, the small positive reference gap
(0.007-0.026 MSE) makes $G$ sensitive to small absolute-error
changes. The exact reference uses one channel's input history;
backbone features may belong to a different function class.

\paragraph{Reading uncertainty and allocation agreement.}
Backbone-feature tables distinguish training and data variation.
The notation $\mu(a)[b]$ gives the mean of three realization means,
mean within-realization seed SD, and SD of realization means.
Both SDs $a,b$ are in units of $10^{-4}$:
$0.9113(8)[38]$ means $0.9113\pm0.0008\,[0.0038]$.
Raw-window tables report mean and sample SD across realizations.
Training seeds are not independent datasets.
Adjusted Rand index (ARI) compares generating families with each
channel's highest-weight expert, using the most frequent test-window
choice for Input-dependent routing.

\subsection{Allocation with Backbone Features}
\label{app:synthetic-native}

Table~\ref{tab:synthetic-native-complete} extends the allocation
comparisons across all 48 settings. Input-dependent routing does not
improve on Global correction in any setting. Persistent ID reduces
Global-correction MSE by 0.40-11.36\% and outperforms one-expert
Ridge-based routing throughout. Identity + input has lower mean MSE
than Persistent ID in 26/48 settings, but its mean relative effect is
approximately $-0.004\%$: input conditioning brings no consistent
additional benefit. Dense ID remains effective with all four experts.
Useful gains coexist with low family-agreement ARI at $s=0$, so
specialization does not require recovering generating-family labels.

\begingroup
\scriptsize
\setlength{\tabcolsep}{1.6pt}
\begin{longtable}{@{}rrrrrrrrrrr@{}}
\caption{Backbone-feature test MSE for all 48 settings, with uncertainty
notation defined in Appendix~\ref{app:synthetic-protocol}.
Random, Family, and Ridge use one active expert; ID, Input, ID+input,
and Dense denote Persistent ID, Input-dependent, Identity + input, and
Dense ID routing. Bold/underline mark the best/second distinct displayed
means per row.}\label{tab:synthetic-native-complete}\\
\toprule
$s$ & $H$ & Base & Global & Random & Family & Ridge & ID & Input & ID+input & Dense\\
\midrule
\endfirsthead
\multicolumn{11}{c}{\tablename\ \thetable\ continued}\\
\toprule
$s$ & $H$ & Base & Global & Random & Family & Ridge & ID & Input & ID+input & Dense\\
\midrule
\endhead
\midrule
\multicolumn{11}{r}{\footnotesize Continued on the next page}\\
\endfoot
\bottomrule
\endlastfoot
\multicolumn{11}{@{}l}{\emph{DLinear}}\\*
0 & 96 & $0.9342(0)[34]$ & $0.9288(0)[34]$ & $0.9295(1)[34]$ & $0.9278(0)[36]$ & $0.9153(4)[35]$ & $0.9113(8)[38]$ & $0.9293(0)[34]$ & $\underline{0.9101}(2)[36]$ & $\mathbf{0.9068}(3)[39]$\\
0 & 192 & $0.9537(0)[33]$ & $0.9509(0)[33]$ & $0.9513(0)[33]$ & $0.9500(0)[36]$ & $0.9406(1)[34]$ & $\underline{0.9381}(3)[35]$ & $0.9511(0)[33]$ & $0.9382(9)[36]$ & $\mathbf{0.9343}(3)[36]$\\
0 & 336 & $0.9709(0)[29]$ & $0.9693(0)[29]$ & $0.9695(0)[29]$ & $0.9687(0)[30]$ & $0.9624(0)[29]$ & $\underline{0.9608}(2)[30]$ & $0.9694(0)[29]$ & $0.9610(6)[35]$ & $\mathbf{0.9585}(2)[29]$\\
0 & 720 & $0.9863(0)[24]$ & $0.9855(0)[24]$ & $0.9857(0)[24]$ & $0.9853(0)[24]$ & $0.9822(0)[25]$ & $0.9815(2)[26]$ & $0.9856(0)[24]$ & $\underline{0.9814}(1)[26]$ & $\mathbf{0.9805}(1)[23]$\\
0.33 & 96 & $0.9147(0)[27]$ & $0.9091(0)[27]$ & $0.9099(1)[27]$ & $0.9012(1)[24]$ & $0.8842(2)[29]$ & $0.8779(3)[32]$ & $0.9097(0)[27]$ & $\underline{0.8770}(8)[32]$ & $\mathbf{0.8700}(2)[33]$\\
0.33 & 192 & $0.9356(0)[24]$ & $0.9326(0)[24]$ & $0.9330(1)[24]$ & $0.9264(1)[22]$ & $0.9125(0)[28]$ & $0.9084(3)[34]$ & $0.9329(0)[24]$ & $\underline{0.9082}(1)[35]$ & $\mathbf{0.9017}(3)[27]$\\
0.33 & 336 & $0.9565(0)[23]$ & $0.9547(0)[22]$ & $0.9550(1)[23]$ & $0.9502(0)[22]$ & $0.9397(0)[26]$ & $0.9369(1)[28]$ & $0.9549(0)[22]$ & $\underline{0.9368}(1)[27]$ & $\mathbf{0.9325}(1)[25]$\\
0.33 & 720 & $0.9823(0)[23]$ & $0.9814(0)[23]$ & $0.9816(1)[23]$ & $0.9791(0)[24]$ & $0.9737(0)[25]$ & $0.9723(1)[26]$ & $0.9815(0)[23]$ & $\underline{0.9722}(0)[26]$ & $\mathbf{0.9703}(1)[26]$\\
0.66 & 96 & $0.8803(0)[49]$ & $0.8745(0)[49]$ & $0.8754(1)[49]$ & $0.8486(2)[46]$ & $0.8380(1)[45]$ & $0.8275(5)[37]$ & $0.8752(0)[49]$ & $\underline{0.8266}(3)[37]$ & $\mathbf{0.8156}(3)[35]$\\
0.66 & 192 & $0.8998(0)[46]$ & $0.8966(0)[46]$ & $0.8970(1)[46]$ & $0.8748(1)[45]$ & $0.8656(0)[40]$ & $0.8576(2)[41]$ & $0.8969(0)[46]$ & $\underline{0.8574}(1)[38]$ & $\mathbf{0.8471}(2)[40]$\\
0.66 & 336 & $0.9207(0)[47]$ & $0.9187(0)[47]$ & $0.9191(1)[47]$ & $0.9007(0)[45]$ & $0.8934(0)[42]$ & $\underline{0.8866}(2)[42]$ & $0.9189(0)[47]$ & $\underline{0.8866}(1)[41]$ & $\mathbf{0.8799}(2)[41]$\\
0.66 & 720 & $0.9551(0)[51]$ & $0.9541(0)[51]$ & $0.9543(1)[51]$ & $0.9435(0)[48]$ & $0.9393(0)[48]$ & $0.9356(2)[44]$ & $0.9542(0)[51]$ & $\underline{0.9355}(1)[44]$ & $\mathbf{0.9320}(2)[45]$\\
1 & 96 & $0.8505(0)[91]$ & $0.8443(0)[90]$ & $0.8453(1)[89]$ & $0.7678(0)[73]$ & $0.7655(0)[74]$ & $0.7524(4)[69]$ & $0.8450(0)[90]$ & $\underline{0.7523}(6)[73]$ & $\mathbf{0.7380}(17)[49]$\\
1 & 192 & $0.8635(0)[87]$ & $0.8598(0)[87]$ & $0.8605(1)[86]$ & $0.7938(0)[73]$ & $0.7915(0)[74]$ & $0.7809(3)[71]$ & $0.8603(0)[87]$ & $\underline{0.7808}(3)[73]$ & $\mathbf{0.7676}(4)[70]$\\
1 & 336 & $0.8754(0)[83]$ & $0.8728(0)[83]$ & $0.8733(1)[82]$ & $0.8109(0)[69]$ & $0.8088(0)[70]$ & $\underline{0.7989}(6)[66]$ & $0.8732(0)[83]$ & $0.7993(2)[67]$ & $\mathbf{0.7898}(3)[71]$\\
1 & 720 & $0.9069(0)[85]$ & $0.9054(0)[85]$ & $0.9058(1)[85]$ & $0.8586(0)[72]$ & $0.8571(0)[76]$ & $\underline{0.8504}(3)[74]$ & $0.9057(0)[85]$ & $\underline{0.8504}(2)[76]$ & $\mathbf{0.8442}(3)[75]$\\
\addlinespace[3pt]
\multicolumn{11}{@{}l}{\emph{PatchTST}}\\*
0 & 96 & $0.9531(0)[36]$ & $0.9502(1)[36]$ & $0.9504(1)[36]$ & $0.9486(1)[38]$ & $0.9349(1)[38]$ & $0.9320(5)[35]$ & $0.9504(1)[36]$ & $\underline{0.9314}(1)[38]$ & $\mathbf{0.9278}(1)[39]$\\
0 & 192 & $0.9775(1)[37]$ & $0.9754(0)[36]$ & $0.9757(0)[36]$ & $0.9744(0)[39]$ & $0.9644(0)[37]$ & $0.9621(1)[39]$ & $0.9756(0)[36]$ & $\underline{0.9620}(0)[38]$ & $\mathbf{0.9586}(3)[37]$\\
0 & 336 & $1.0003(0)[33]$ & $0.9989(0)[32]$ & $0.9991(0)[32]$ & $0.9983(0)[34]$ & $0.9916(0)[33]$ & $\underline{0.9901}(2)[34]$ & $0.9991(0)[32]$ & $\underline{0.9901}(0)[34]$ & $\mathbf{0.9880}(1)[31]$\\
0 & 720 & $1.0252(0)[30]$ & $1.0244(0)[30]$ & $1.0245(0)[30]$ & $1.0242(0)[31]$ & $1.0210(0)[31]$ & $\underline{1.0203}(1)[30]$ & $1.0245(0)[30]$ & $\underline{1.0203}(1)[31]$ & $\mathbf{1.0193}(2)[30]$\\
0.33 & 96 & $0.9293(1)[28]$ & $0.9254(1)[29]$ & $0.9257(2)[29]$ & $0.9170(1)[26]$ & $0.8974(1)[33]$ & $\underline{0.8918}(3)[35]$ & $0.9257(1)[29]$ & $0.8921(4)[38]$ & $\mathbf{0.8858}(3)[33]$\\
0.33 & 192 & $0.9539(1)[25]$ & $0.9507(0)[25]$ & $0.9511(1)[25]$ & $0.9445(0)[23]$ & $0.9295(1)[30]$ & $\underline{0.9256}(1)[32]$ & $0.9511(0)[25]$ & $0.9258(6)[29]$ & $\mathbf{0.9193}(1)[32]$\\
0.33 & 336 & $0.9787(0)[20]$ & $0.9764(0)[20]$ & $0.9767(1)[21]$ & $0.9721(0)[20]$ & $0.9609(0)[25]$ & $0.9581(1)[25]$ & $0.9767(0)[21]$ & $\underline{0.9580}(1)[26]$ & $\mathbf{0.9540}(1)[24]$\\
0.33 & 720 & $1.0132(0)[12]$ & $1.0119(0)[13]$ & $1.0122(1)[13]$ & $1.0098(0)[13]$ & $1.0040(0)[15]$ & $\underline{1.0025}(1)[16]$ & $1.0121(0)[13]$ & $1.0027(3)[18]$ & $\mathbf{1.0007}(1)[14]$\\
0.66 & 96 & $0.8919(1)[43]$ & $0.8869(1)[44]$ & $0.8872(2)[44]$ & $0.8584(1)[38]$ & $0.8474(0)[39]$ & $\underline{0.8371}(5)[36]$ & $0.8872(2)[44]$ & $\underline{0.8371}(3)[34]$ & $\mathbf{0.8272}(5)[33]$\\
0.66 & 192 & $0.9138(1)[39]$ & $0.9094(0)[40]$ & $0.9100(1)[40]$ & $0.8868(0)[39]$ & $0.8777(0)[34]$ & $\underline{0.8697}(3)[35]$ & $0.9099(0)[40]$ & $0.8699(10)[42]$ & $\mathbf{0.8597}(7)[32]$\\
0.66 & 336 & $0.9367(0)[41]$ & $0.9332(0)[42]$ & $0.9338(1)[42]$ & $0.9151(0)[39]$ & $0.9078(0)[36]$ & $0.9012(3)[35]$ & $0.9337(0)[42]$ & $\underline{0.9010}(1)[34]$ & $\mathbf{0.8939}(1)[38]$\\
0.66 & 720 & $0.9778(0)[47]$ & $0.9756(0)[48]$ & $0.9761(1)[47]$ & $0.9651(0)[43]$ & $0.9609(0)[43]$ & $\underline{0.9570}(2)[41]$ & $0.9760(1)[48]$ & $\underline{0.9570}(0)[40]$ & $\mathbf{0.9532}(2)[43]$\\
1 & 96 & $0.8578(1)[84]$ & $0.8518(1)[83]$ & $0.8522(1)[82]$ & $0.7727(0)[65]$ & $0.7704(1)[65]$ & $0.7577(2)[65]$ & $0.8520(2)[83]$ & $\underline{0.7572}(1)[62]$ & $\mathbf{0.7421}(5)[58]$\\
1 & 192 & $0.8724(0)[84]$ & $0.8665(0)[83]$ & $0.8673(1)[82]$ & $0.8007(0)[68]$ & $0.7983(0)[69]$ & $0.7876(5)[72]$ & $0.8671(0)[83]$ & $\underline{0.7875}(1)[66]$ & $\mathbf{0.7726}(7)[69]$\\
1 & 336 & $0.8849(1)[82]$ & $0.8791(0)[81]$ & $0.8802(1)[80]$ & $0.8181(0)[65]$ & $0.8159(0)[66]$ & $0.8062(4)[62]$ & $0.8798(1)[81]$ & $\underline{0.8059}(1)[65]$ & $\mathbf{0.7960}(10)[62]$\\
1 & 720 & $0.9185(1)[89]$ & $0.9138(0)[86]$ & $0.9149(1)[87]$ & $0.8677(1)[72]$ & $0.8661(1)[75]$ & $0.8590(2)[76]$ & $0.9143(1)[87]$ & $\underline{0.8589}(1)[74]$ & $\mathbf{0.8523}(2)[73]$\\
\addlinespace[3pt]
\multicolumn{11}{@{}l}{\emph{iTransformer}}\\*
0 & 96 & $0.9522(1)[35]$ & $0.9513(1)[35]$ & $0.9516(1)[35]$ & $0.9498(1)[37]$ & $0.9354(1)[38]$ & $\underline{0.9320}(2)[38]$ & $0.9514(1)[35]$ & $0.9329(11)[41]$ & $\mathbf{0.9293}(4)[39]$\\
0 & 192 & $0.9767(1)[36]$ & $0.9761(1)[36]$ & $0.9763(0)[36]$ & $0.9751(0)[38]$ & $0.9646(1)[37]$ & $0.9624(6)[41]$ & $0.9761(1)[36]$ & $\underline{0.9623}(4)[39]$ & $\mathbf{0.9593}(2)[39]$\\
0 & 336 & $0.9997(0)[32]$ & $0.9993(1)[32]$ & $0.9995(0)[32]$ & $0.9987(0)[34]$ & $0.9917(0)[32]$ & $\underline{0.9901}(1)[33]$ & $0.9993(0)[32]$ & $0.9904(2)[35]$ & $\mathbf{0.9886}(3)[32]$\\
0 & 720 & $1.0248(0)[29]$ & $1.0246(0)[29]$ & $1.0247(0)[30]$ & $1.0244(0)[30]$ & $1.0211(0)[30]$ & $\underline{1.0203}(2)[31]$ & $1.0246(0)[30]$ & $1.0206(2)[31]$ & $\mathbf{1.0197}(2)[28]$\\
0.33 & 96 & $0.9276(1)[29]$ & $0.9266(1)[29]$ & $0.9268(1)[28]$ & $0.9180(1)[26]$ & $0.8974(1)[33]$ & $\underline{0.8921}(3)[35]$ & $0.9267(1)[28]$ & $0.8941(26)[35]$ & $\mathbf{0.8875}(6)[34]$\\
0.33 & 192 & $0.9522(0)[25]$ & $0.9515(1)[25]$ & $0.9517(1)[25]$ & $0.9448(0)[22]$ & $0.9292(0)[29]$ & $\underline{0.9253}(3)[31]$ & $0.9516(1)[25]$ & $0.9254(2)[30]$ & $\mathbf{0.9202}(5)[30]$\\
0.33 & 336 & $0.9774(0)[20]$ & $0.9769(1)[20]$ & $0.9771(1)[20]$ & $0.9721(0)[19]$ & $0.9606(0)[24]$ & $\underline{0.9579}(2)[24]$ & $0.9770(0)[20]$ & $0.9580(6)[29]$ & $\mathbf{0.9545}(3)[27]$\\
0.33 & 720 & $1.0124(0)[13]$ & $1.0122(0)[13]$ & $1.0123(0)[13]$ & $1.0098(0)[13]$ & $1.0039(0)[15]$ & $\underline{1.0024}(0)[15]$ & $1.0122(0)[13]$ & $1.0030(8)[14]$ & $\mathbf{1.0009}(1)[15]$\\
0.66 & 96 & $0.8894(1)[45]$ & $0.8881(1)[44]$ & $0.8884(1)[44]$ & $0.8582(1)[38]$ & $0.8470(1)[40]$ & $\underline{0.8366}(2)[35]$ & $0.8883(1)[44]$ & $0.8378(9)[36]$ & $\mathbf{0.8288}(3)[31]$\\
0.66 & 192 & $0.9112(0)[41]$ & $0.9103(1)[41]$ & $0.9105(1)[41]$ & $0.8863(1)[39]$ & $0.8770(1)[35]$ & $0.8690(2)[35]$ & $0.9104(0)[41]$ & $\underline{0.8689}(4)[35]$ & $\mathbf{0.8600}(2)[36]$\\
0.66 & 336 & $0.9345(0)[43]$ & $0.9338(1)[42]$ & $0.9340(1)[42]$ & $0.9145(0)[40]$ & $0.9071(0)[37]$ & $\underline{0.9005}(1)[36]$ & $0.9339(0)[42]$ & $0.9006(2)[34]$ & $\mathbf{0.8941}(2)[38]$\\
0.66 & 720 & $0.9764(0)[48]$ & $0.9760(0)[48]$ & $0.9762(1)[48]$ & $0.9648(0)[45]$ & $0.9605(0)[45]$ & $0.9568(2)[40]$ & $0.9760(0)[48]$ & $\underline{0.9567}(1)[40]$ & $\mathbf{0.9535}(2)[42]$\\
1 & 96 & $0.8544(1)[85]$ & $0.8530(1)[85]$ & $0.8533(1)[84]$ & $0.7712(1)[65]$ & $0.7689(1)[66]$ & $\underline{0.7561}(7)[59]$ & $0.8532(1)[85]$ & $0.7573(9)[64]$ & $\mathbf{0.7450}(17)[67]$\\
1 & 192 & $0.8685(1)[83]$ & $0.8674(1)[83]$ & $0.8677(1)[82]$ & $0.7985(1)[68]$ & $0.7963(1)[68]$ & $\underline{0.7854}(4)[62]$ & $0.8675(1)[83]$ & $\underline{0.7854}(4)[67]$ & $\mathbf{0.7738}(11)[65]$\\
1 & 336 & $0.8807(0)[81]$ & $0.8798(0)[81]$ & $0.8801(1)[80]$ & $0.8160(1)[65]$ & $0.8139(1)[67]$ & $\underline{0.8041}(3)[62]$ & $0.8799(0)[81]$ & $\underline{0.8041}(2)[64]$ & $\mathbf{0.7949}(3)[68]$\\
1 & 720 & $0.9148(0)[86]$ & $0.9142(0)[87]$ & $0.9144(1)[86]$ & $0.8664(0)[72]$ & $0.8648(0)[76]$ & $\underline{0.8577}(3)[73]$ & $0.9143(0)[87]$ & $0.8578(3)[72]$ & $\mathbf{0.8521}(4)[72]$\\
\addlinespace[3pt]
\end{longtable}
\endgroup

\paragraph{Matching two active experts.}
For the endpoint comparison in Section~\ref{sec:synthetic-results},
Ridge-based routing equally combines the channel's training-fitted group
and nearest alternative; Random routing equally combines balanced random
pairs. Both share the frozen backbones, initialization, rank-16 bank,
and fitting protocol of the learned two-expert mixtures.
Ridge-based routing improves on Random in all twelve settings and all
108 seed-realization comparisons, averaging 6.73\% at $s=1,H=96$
and 4.08\% at $s=1,H=720$.
Persistent ID improves on Ridge-based routing in all twelve setting means
(Table~\ref{tab:synthetic-k2-complete}).

\begin{figure}[!ht]
\raggedright
\includegraphics[width=0.45\textwidth]{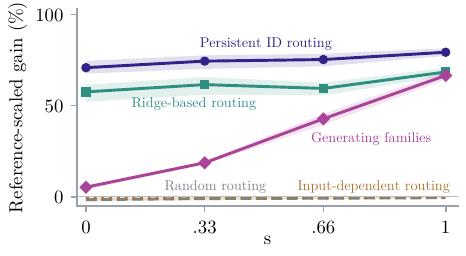}
\caption{\textbf{One-expert Ridge-based routing.} Other curves match \hyperref[fig:synthetic-allocation-gap]{Figure~\ref*{fig:synthetic-allocation-gap}}.}
\label{fig:synthetic-allocation-gap-k1}
\end{figure}

\Needspace{240pt}
\hyperref[fig:synthetic-allocation-gap-k1]{Figure~\ref*{fig:synthetic-allocation-gap-k1}} uses one-expert Ridge-based routing. At $H=96$, its reference-scaled gain over Global is higher at $s=1$ than at $s=0$, while the two-expert rule shows the reverse. This is consistent with equal-weight mixing diluting distinct forecasting rules under stronger separation, while overlap makes averaging less harmful. Random routing and generating-family controls use one expert in both figures.

\begingroup
\scriptsize
\setlength{\tabcolsep}{3pt}
\begin{longtable}{@{}rrrrrrr@{}}
\caption{Matched two-expert Random and Ridge-based controls on all twelve
endpoint settings. MSE uses $\mu(a)[b]$ notation; $G$ is the
reference-scaled gain over the same study's Global correction (\%). The last column
gives Ridge's MSE reduction over Random (\%). Bold/underline compare
the two controls.}\label{tab:synthetic-k2-complete}\\
\toprule
$s$ & $H$ & Random MSE & Ridge MSE & $G_{\rm Random}$ & $G_{\rm Ridge}$ & Ridge vs Random\\
\midrule
\endfirsthead
\multicolumn{7}{c}{\tablename\ \thetable\ continued}\\
\toprule
$s$ & $H$ & Random MSE & Ridge MSE & $G_{\rm Random}$ & $G_{\rm Ridge}$ & Ridge vs Random\\
\midrule
\endhead
\midrule
\multicolumn{7}{r}{\footnotesize Continued on the next page}\\
\endfoot
\bottomrule
\endlastfoot
\multicolumn{7}{@{}l}{\emph{DLinear}}\\*
0 & 96 & $\underline{0.9292}(0)[33]$ & $\mathbf{0.9148}(0)[35]$ & $\underline{-1.7}$ & $\mathbf{54.1}$ & 1.55\\
0 & 720 & $\underline{0.9856}(0)[24]$ & $\mathbf{0.9819}(0)[23]$ & $\underline{-1.5}$ & $\mathbf{50.5}$ & 0.38\\
1 & 96 & $\underline{0.8447}(2)[91]$ & $\mathbf{0.7879}(2)[73]$ & $\underline{-0.4}$ & $\mathbf{47.4}$ & 6.73\\
1 & 720 & $\underline{0.9056}(2)[86]$ & $\mathbf{0.8691}(1)[78]$ & $\underline{-0.2}$ & $\mathbf{41.9}$ & 4.03\\
\addlinespace[3pt]
\multicolumn{7}{@{}l}{\emph{PatchTST}}\\*
0 & 96 & $\underline{0.9503}(1)[36]$ & $\mathbf{0.9357}(1)[37]$ & $\underline{-0.5}$ & $\mathbf{56.3}$ & 1.54\\
0 & 720 & $\underline{1.0245}(0)[29]$ & $\mathbf{1.0208}(0)[29]$ & $\underline{-1.6}$ & $\mathbf{50.6}$ & 0.36\\
1 & 96 & $\underline{0.8518}(2)[84]$ & $\mathbf{0.7942}(1)[64]$ & $\underline{0.0}$ & $\mathbf{48.3}$ & 6.75\\
1 & 720 & $\underline{0.9146}(1)[87]$ & $\mathbf{0.8770}(0)[80]$ & $\underline{-0.9}$ & $\mathbf{42.4}$ & 4.11\\
\addlinespace[3pt]
\multicolumn{7}{@{}l}{\emph{iTransformer}}\\*
0 & 96 & $\underline{0.9515}(1)[35]$ & $\mathbf{0.9371}(0)[36]$ & $\underline{-0.7}$ & $\mathbf{55.0}$ & 1.51\\
0 & 720 & $\underline{1.0247}(0)[29]$ & $\mathbf{1.0210}(0)[29]$ & $\underline{-1.1}$ & $\mathbf{50.4}$ & 0.36\\
1 & 96 & $\underline{0.8530}(3)[86]$ & $\mathbf{0.7958}(1)[66]$ & $\underline{0.1}$ & $\mathbf{48.1}$ & 6.70\\
1 & 720 & $\underline{0.9143}(1)[87]$ & $\mathbf{0.8769}(0)[79]$ & $\underline{-0.1}$ & $\mathbf{43.1}$ & 4.09\\
\addlinespace[3pt]
\end{longtable}
\endgroup

\subsection{Supplementary Raw-Window Checks}
\label{app:numerical-synthetic}

\paragraph{Raw-window predictors.}
These checks replace backbone features by the channel's raw history,
using rank-32 experts and a rank-128 Global correction.
Backbones are pretrained without channel identity using
Appendix~\ref{app:backbones}, then frozen. Corrections follow
Appendix~\ref{app:correction-framework}; all variants share training
settings, initialization, minibatch order, and validation checkpoint selection.

\begin{figure}[htbp]
\centering
\includegraphics[width=\textwidth]{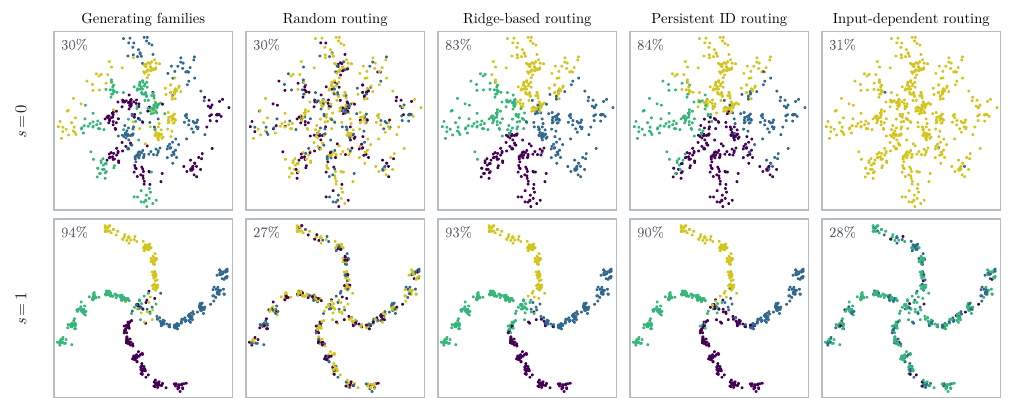}
\caption{\textbf{Raw-window allocation patterns.} The 512 channels at
$s=0$ and $s=1$, colored by family, Random, Ridge-based, Persistent ID,
and Input-dependent allocation (left to right). Random and Ridge show
one-expert partitions, not the matched two-expert controls.
Persistent ID shows the highest-weight expert; Input-dependent shows
the most frequent highest-weight expert across test windows.
These summaries omit mixture weights. Percentages give agreement with
the dominant generating echo under the displayed color alignment.}
\label{fig:synthetic-groupings}
\end{figure}

At $s=0$, Ridge groups cut across families
(Figures~\ref{fig:toy-hidden}a and~\ref{fig:synthetic-groupings}):
winding arms place similar echo mixes in different families and different
mixes within one family. One exact map per Ridge group closes 70.4\%
of the population gap at $L=H=96$, compared with 9.4\% for family groups
(Table~\ref{tab:hidden-echo-exact}). These are exact-reference gains,
not the figure's echo-agreement percentages or fitted two-expert gains.
The backbone-feature one-expert Ridge control also has lower mean test
MSE than family grouping in all twelve $s=0$ backbone-horizon settings.

\paragraph{Persistent allocation across horizons.}
Tables~\ref{tab:synthetic-full}-\ref{tab:synthetic-full-long} retain all
240 backbone-separation-horizon-predictor results, averaged over
three realizations. Persistent ID achieves reference-scaled gains
over Global of 78.7-92.5\% across horizon-separation combinations,
averaging backbones and realizations. Dense ID improves mean MSE further, while Input-dependent
routing stays close to Global correction. The dense comparison changes
the number of active experts and does not establish an advantage for sparsity.

\paragraph{Longer histories and additional seeds.}
Endpoint replications at $s\in\{0,1\}$ first average training seeds
0-2 within each realization. Additional seeds change Global-correction
MSE by at most 0.0002; Persistent ID reduces backbone MSE by
13.7-13.9\% at $s=1$. With $L=720,H=96$, the echoes fall inside
the input window: Global MSE at $s=1$ falls to about 0.437,
Persistent ID improves on the frozen backbones by 13.9-14.2\%,
and Input-dependent routing improves on Global correction by about
1.1\% (ARI 0.09-0.14). These are raw-window checks.

\begingroup
\small
\setlength{\tabcolsep}{2pt}
\begin{longtable}{@{}lrrrr@{\hspace{6pt}}rrrr@{}}
\caption{Raw-window test MSE at $L=96$, $H=96,192$, with rank-32
corrections: mean (sample SD) over three realizations.
SD is in units of $10^{-4}$, e.g.\
$0.9342(34)=0.9342\pm0.0034$.
Bold/underline mark the lowest/second distinct displayed means within
each backbone, horizon, and separation, excluding exact references.
Ties share rank. Backbone-feature results appear in
Table~\ref{tab:synthetic-native-complete}.}
\label{tab:synthetic-full}\\
\toprule
 & \multicolumn{4}{c}{$H=96$} & \multicolumn{4}{c}{$H=192$} \\
\cmidrule(lr){2-5}\cmidrule(lr){6-9}
$s$ & $0$ & $0.33$ & $0.66$ & $1$ & $0$ & $0.33$ & $0.66$ & $1$ \\
\midrule
\endfirsthead
\multicolumn{9}{c}{\tablename\ \thetable\ continued}\\
\toprule
 & \multicolumn{4}{c}{$H=96$} & \multicolumn{4}{c}{$H=192$} \\
\cmidrule(lr){2-5}\cmidrule(lr){6-9}
$s$ & $0$ & $0.33$ & $0.66$ & $1$ & $0$ & $0.33$ & $0.66$ & $1$ \\
\midrule
\endhead
\midrule
\multicolumn{9}{r}{\footnotesize Continued on the next page}\\
\endfoot
\bottomrule
\endlastfoot

Optimal shared & 0.9274(34) & 0.9076(27) & 0.8727(48) & 0.8410(87) & 0.9500(34) & 0.9316(24) & 0.8953(45) & 0.8572(85) \\
Optimal per-family & 0.9250(36) & 0.8972(25) & 0.8385(43) & 0.7493(68) & 0.9481(36) & 0.9224(22) & 0.8637(43) & 0.7672(70) \\
Optimal per-channel & 0.9015(37) & 0.8631(32) & 0.8070(37) & 0.7220(60) & 0.9283(35) & 0.8924(30) & 0.8344(38) & 0.7402(64) \\
\midrule
\multicolumn{9}{@{}l}{\emph{DLinear}} \\*
Frozen backbone & 0.9342(34) & 0.9147(27) & 0.8803(49) & 0.8505(91) & 0.9537(33) & 0.9356(24) & 0.8998(46) & 0.8635(87) \\
Global correction & 0.9288(34) & 0.9093(25) & 0.8745(49) & 0.8443(91) & 0.9509(33) & 0.9326(23) & 0.8966(45) & 0.8599(87) \\
\shortstack[l]{Input-dependent\\routing} & 0.9289(34) & 0.9093(27) & 0.8747(49) & 0.8446(91) & 0.9509(33) & 0.9327(24) & 0.8967(46) & 0.8601(85) \\
\shortstack[l]{Persistent ID\\routing} & \underline{0.9065}(38) & \underline{0.8698}(32) & \underline{0.8158}(39) & \underline{0.7341}(68) & \underline{0.9326}(34) & \underline{0.8989}(30) & \underline{0.8437}(42) & \underline{0.7539}(67) \\
\shortstack[l]{Dense ID\\routing} & \textbf{0.9048}(39) & \textbf{0.8669}(31) & \textbf{0.8118}(37) & \textbf{0.7296}(66) & \textbf{0.9312}(36) & \textbf{0.8962}(29) & \textbf{0.8393}(38) & \textbf{0.7492}(69) \\
\midrule
\multicolumn{9}{@{}l}{\emph{iTransformer}} \\*
Frozen backbone & 0.9522(35) & 0.9277(29) & 0.8895(45) & 0.8545(85) & 0.9767(36) & 0.9522(25) & 0.9112(41) & 0.8685(83) \\
Global correction & 0.9290(34) & 0.9092(27) & 0.8748(50) & 0.8446(91) & 0.9510(33) & 0.9328(24) & 0.8967(46) & 0.8600(87) \\
\shortstack[l]{Input-dependent\\routing} & 0.9291(34) & 0.9095(27) & 0.8749(49) & 0.8450(91) & 0.9511(34) & 0.9329(24) & 0.8969(46) & 0.8603(87) \\
\shortstack[l]{Persistent ID\\routing} & \underline{0.9067}(37) & \underline{0.8702}(31) & \underline{0.8166}(42) & \underline{0.7350}(67) & \underline{0.9334}(34) & \underline{0.8997}(28) & \underline{0.8444}(39) & \underline{0.7548}(69) \\
\shortstack[l]{Dense ID\\routing} & \textbf{0.9050}(39) & \textbf{0.8672}(31) & \textbf{0.8121}(38) & \textbf{0.7306}(62) & \textbf{0.9321}(36) & \textbf{0.8971}(30) & \textbf{0.8401}(39) & \textbf{0.7501}(67) \\
\midrule
\multicolumn{9}{@{}l}{\emph{PatchTST}} \\*
Frozen backbone & 0.9507(35) & 0.9260(29) & 0.8877(44) & 0.8525(84) & 0.9758(36) & 0.9511(25) & 0.9099(40) & 0.8670(83) \\
Global correction & 0.9288(34) & 0.9095(28) & 0.8746(50) & 0.8447(90) & 0.9510(34) & 0.9327(26) & 0.8966(45) & 0.8601(86) \\
\shortstack[l]{Input-dependent\\routing} & 0.9291(34) & 0.9095(28) & 0.8749(50) & 0.8449(90) & 0.9512(32) & 0.9329(25) & 0.8969(46) & 0.8602(88) \\
\shortstack[l]{Persistent ID\\routing} & \underline{0.9068}(38) & \underline{0.8701}(31) & \underline{0.8164}(34) & \underline{0.7343}(69) & \underline{0.9328}(34) & \underline{0.8988}(29) & \underline{0.8435}(39) & \underline{0.7538}(69) \\
\shortstack[l]{Dense ID\\routing} & \textbf{0.9051}(39) & \textbf{0.8673}(32) & \textbf{0.8120}(38) & \textbf{0.7298}(66) & \textbf{0.9315}(36) & \textbf{0.8963}(30) & \textbf{0.8395}(37) & \textbf{0.7493}(68) \\
\end{longtable}
\endgroup

\begingroup
\small
\setlength{\tabcolsep}{2pt}
\begin{longtable}{@{}lrrrr@{\hspace{6pt}}rrrr@{}}
\caption{Raw-window test MSE at $L=96$, $H=336,720$.
Protocol, uncertainty notation, and ranking follow
Table~\ref{tab:synthetic-full}; backbones are reported separately.}
\label{tab:synthetic-full-long}\\
\toprule
 & \multicolumn{4}{c}{$H=336$} & \multicolumn{4}{c}{$H=720$} \\
\cmidrule(lr){2-5}\cmidrule(lr){6-9}
$s$ & $0$ & $0.33$ & $0.66$ & $1$ & $0$ & $0.33$ & $0.66$ & $1$ \\
\midrule
\endfirsthead
\multicolumn{9}{c}{\tablename\ \thetable\ continued}\\
\toprule
 & \multicolumn{4}{c}{$H=336$} & \multicolumn{4}{c}{$H=720$} \\
\cmidrule(lr){2-5}\cmidrule(lr){6-9}
$s$ & $0$ & $0.33$ & $0.66$ & $1$ & $0$ & $0.33$ & $0.66$ & $1$ \\
\midrule
\endhead
\midrule
\multicolumn{9}{r}{\footnotesize Continued on the next page}\\
\endfoot
\bottomrule
\endlastfoot

Optimal shared & 0.9687(29) & 0.9540(22) & 0.9177(46) & 0.8705(82) & 0.9851(24) & 0.9808(24) & 0.9531(50) & 0.9029(85) \\
Optimal per-family & 0.9674(30) & 0.9472(21) & 0.8916(44) & 0.7851(67) & 0.9845(25) & 0.9773(24) & 0.9374(46) & 0.8363(69) \\
Optimal per-channel & 0.9541(29) & 0.9248(26) & 0.8672(40) & 0.7594(62) & 0.9780(24) & 0.9656(26) & 0.9227(43) & 0.8163(69) \\
\midrule
\multicolumn{9}{@{}l}{\emph{DLinear}} \\*
Frozen backbone & 0.9709(29) & 0.9565(23) & 0.9207(47) & 0.8754(83) & 0.9863(24) & 0.9823(23) & 0.9551(51) & 0.9069(86) \\
Global correction & 0.9693(28) & 0.9547(22) & 0.9187(46) & 0.8729(84) & 0.9856(24) & 0.9814(23) & 0.9542(52) & 0.9054(85) \\
\shortstack[l]{Input-dependent\\routing} & 0.9693(29) & 0.9548(22) & 0.9189(47) & 0.8729(83) & 0.9856(24) & 0.9815(23) & 0.9542(51) & 0.9055(85) \\
\shortstack[l]{Persistent ID\\routing} & \underline{0.9571}(29) & \underline{0.9298}(28) & \underline{0.8751}(41) & \underline{0.7729}(69) & \underline{0.9797}(24) & \underline{0.9686}(26) & \underline{0.9282}(44) & \underline{0.8283}(72) \\
\shortstack[l]{Dense ID\\routing} & \textbf{0.9563}(29) & \textbf{0.9278}(25) & \textbf{0.8715}(41) & \textbf{0.7682}(67) & \textbf{0.9794}(24) & \textbf{0.9676}(26) & \textbf{0.9263}(45) & \textbf{0.8247}(73) \\
\midrule
\multicolumn{9}{@{}l}{\emph{iTransformer}} \\*
Frozen backbone & 0.9997(32) & 0.9774(20) & 0.9345(42) & 0.8807(81) & 1.0248(30) & 1.0125(13) & 0.9765(48) & 0.9149(86) \\
Global correction & 0.9694(29) & 0.9548(23) & 0.9189(47) & 0.8731(82) & 0.9856(24) & 0.9815(23) & 0.9542(51) & 0.9054(85) \\
\shortstack[l]{Input-dependent\\routing} & 0.9695(29) & 0.9550(23) & 0.9190(47) & 0.8735(85) & 0.9857(24) & 0.9817(24) & 0.9543(51) & 0.9059(88) \\
\shortstack[l]{Persistent ID\\routing} & \underline{0.9577}(29) & \underline{0.9303}(27) & \underline{0.8756}(41) & \underline{0.7734}(68) & \underline{0.9802}(24) & \underline{0.9691}(26) & \underline{0.9285}(45) & \underline{0.8282}(73) \\
\shortstack[l]{Dense ID\\routing} & \textbf{0.9568}(29) & \textbf{0.9284}(27) & \textbf{0.8719}(40) & \textbf{0.7686}(68) & \textbf{0.9798}(24) & \textbf{0.9681}(26) & \textbf{0.9267}(45) & \textbf{0.8248}(74) \\
\midrule
\multicolumn{9}{@{}l}{\emph{PatchTST}} \\*
Frozen backbone & 0.9991(32) & 0.9766(21) & 0.9335(42) & 0.8796(81) & 1.0245(29) & 1.0120(13) & 0.9758(48) & 0.9141(87) \\
Global correction & 0.9693(29) & 0.9548(23) & 0.9188(46) & 0.8730(83) & 0.9856(24) & 0.9815(23) & 0.9542(50) & 0.9054(85) \\
\shortstack[l]{Input-dependent\\routing} & 0.9695(28) & 0.9549(23) & 0.9190(48) & 0.8732(83) & 0.9857(24) & 0.9816(23) & 0.9542(51) & 0.9057(86) \\
\shortstack[l]{Persistent ID\\routing} & \underline{0.9572}(29) & \underline{0.9297}(27) & \underline{0.8749}(40) & \underline{0.7724}(66) & \underline{0.9800}(25) & \underline{0.9688}(25) & \underline{0.9282}(44) & \underline{0.8279}(74) \\
\shortstack[l]{Dense ID\\routing} & \textbf{0.9564}(29) & \textbf{0.9279}(27) & \textbf{0.8715}(40) & \textbf{0.7680}(68) & \textbf{0.9796}(24) & \textbf{0.9678}(26) & \textbf{0.9263}(44) & \textbf{0.8243}(73) \\
\end{longtable}
\endgroup

\clearpage
\section{Persistent Closed-Form Routing}
\label{app:pcr-benchmark}

This appendix follows Section~\ref{sec:method}: fitting persistent
mixtures, specializing frozen forecasters on GalaxyTS, forecasting on
real data, and comparing accuracy and cost with learned allocation,
MoLE, and CCM.

\subsection{Fitting PCR}
\label{app:pcr-details}

\paragraph{Features and residuals.}
Let $o_{ice}=a_c(X_i)A_ez_c(X_i)\in\mathbb R^H$ be expert $e$'s
correction for training window $i$, and let
$\varepsilon_{ic}=y_{ic}-f_\theta(X_i)_c$.
Stacking $o_{ice}$ over windows and forecast steps gives column $e$
of $D_c$. Thus $G_c=D_c^\top D_c$ and
$b_c=D_c^\top\varepsilon_c$ describe the squared residual error in
training-standardized observation coordinates. The normalization offset
is already in the backbone forecast and cancels from the residual;
DLinear uses $a_c=1$. Feature widths are $F=2L$ for DLinear,
$F=d_{\mathrm{model}}\times12$ for PatchTST at $L=96$, and
$F=d_{\mathrm{model}}$ for iTransformer.

\paragraph{Exact conditional allocation.}
The ridge penalty for channel $c$ is
$\lambda_c=\lambda_0\operatorname{mean}(\operatorname{diag}G_c)$,
with a numerical floor for degenerate cases. For each
$\lambda_0\in\{10^{-6},10^{-4},10^{-2},0.1,1\}$, PCR enumerates
every nonempty support $S$ with $|S|\leq K$ and evaluates
Equation~\eqref{eq:pcr-solution} using linear solves, not explicit inverses.
Negative-weight candidates are infeasible. Among feasible candidates,
PCR minimizes
$\pi_c^\top G_c\pi_c-2b_c^\top\pi_c+\lambda_c\|\pi_c\|_2^2$
on training data, retaining smaller supports when the optimum lies on a boundary.
PCR-K2 checks four singletons and six pairs; PCR-K4 checks all 15
nonempty subsets. Support selection uses regularized training loss,
not validation error or truncation of a dense solution.

Why is this conditional solution exact? Positive ridge makes the
objective strictly convex on each simplex face. A constrained optimum
is the equality-constrained minimizer on its own positive support.
Enumerating all admissible supports, including when $K< E$, and choosing
the lowest feasible objective therefore finds the conditional global
optimum up to numerical precision. This does not establish global
optimality of alternating expert and mixture updates.

\paragraph{Initialization and updates.}
Every method starts from a fresh four-expert bank, with seeded input
factors $U_e$ and zero output factors $V_e$; no fitted router or experts
are reused. Within each setting and seed, methods share initialization,
training-window order, and frozen checkpoint. The first two epochs
train experts with uniform weights $1/4$. At each subsequent even epoch,
PCR fits mixtures from the current bank's training statistics and fixes
them for the next two expert-training epochs. Expert loss is MSE at
$\alpha=1$. Adam resets each round; the second epoch uses half the
bank learning rate, and the global plateau multiplier applies to both.
The backbone stays in evaluation mode with parameters and buffers fixed.

Learned-K4 instead trains channel logits through softmax; Learned-K2
uses the straight-through top-two estimator of
Appendix~\ref{app:allocation-arms}. Both keep one uninterrupted Adam
optimizer and router rate ten times the bank rate.
All methods use batch 16, MSE without load-balancing or entropy penalties,
and no expert-output dropout or gradual sparsification.
DLinear/PatchTST at $H=96,720$ reuse allocation-study bank rates;
other settings select them on seed-zero calibrated validation MSE.
The optimizer policies differ, so this compares fitting procedures,
not an isolated solver intervention.

\paragraph{Calibration and stopping.}
Validation jointly selects PCR's $\lambda_0$ and global correction scale
$\alpha\in\{0,0.05,\ldots,1\}$; learned allocation uses the same scale grid.
Including $\alpha=0$ admits the frozen forecast. Checkpoints are eligible
from epoch four, so K2 cannot select a dense initial checkpoint.
Test data never select learning rates, supports, scales, or checkpoints
within this fixed-rank benchmark.

Validation is evaluated every two epochs, with plateau monitoring from
epoch 14. Improvement requires
$\mathrm{MSE}_{\mathrm{val}}<\mathrm{best}\,(1-10^{-4})$.
After five evaluations without improvement, the global learning-rate
multiplier halves down to $1/16$. Five further non-improving evaluations
at that floor stop training after at least 40 epochs; the cap is 200.
Seven real-data fits reach the cap (Appendix~\ref{app:pcr-budget}).
Plateau stopping does not guarantee convergence.

\paragraph{Caching.}
PCR reuses exact frozen-feature or moment caches when memory permits,
otherwise evaluating the frozen backbone directly. Allocation statistics
are rebuilt for the current expert bank at every solve.
Learned allocation in the real-data grid uses direct backbone evaluation;
the measured fitting costs therefore include caching and stopping effects.

\subsection{Specializing Frozen Forecasters on GalaxyTS}
\label{app:pcr-synthetic}

\paragraph{Evaluation protocol.}
PCR uses the GalaxyTS realizations and frozen backbone features of
Appendix~\ref{app:synthetic}, with four rank-16 experts, three training
seeds per realization, and all 48 backbone-separation-horizon settings.
PCR and Learned-K4/K2 follow Appendix~\ref{app:pcr-details}, starting
from the corresponding allocation study's Dense ID learning rate.
Their rank-64 Global correction is fitted separately from the
Section~\ref{sec:synthetic} control, using the learned-allocation budget
and selection rule. {Consequently, $G$ values from the two studies
do not directly rank accuracy across their methods.}
Both PCR and learned allocation may cache exact
features here, unlike the real-data timing comparison.
Reference-scaled gains and uncertainty follow
Appendix~\ref{app:synthetic-protocol}.

\paragraph{Forecasting accuracy.}
Both PCR variants improve on their matched Global correction in all
48 settings (Table~\ref{tab:synthetic-final-complete}).
PCR-K2 improves on Learned-K2 throughout, averaging 0.15 percentage
points more backbone improvement. PCR-K4 and Learned-K4 are close:
PCR wins 19/48 settings and averages 0.04 percentage points less
backbone improvement. At $s=1,H=96$, PCR-K4 reduces backbone MSE by
13.55\%, with a reference-scaled gain over Global of $G=93.14\%$,
averaging backbones; at $H=720$, these values are 6.90\% and
$G=69.92\%$.

\begingroup
\scriptsize
\setlength{\tabcolsep}{2pt}
\begin{longtable}{@{}rrrrrrrrrr@{}}
\caption{PCR and matched learned-allocation test MSE on all 48 GalaxyTS
settings, alongside exact shared and channel-specific reference errors.
The exact reference predictors use only the target channel's input
history and are evaluated on the same test windows; their errors are
averaged across realizations. Fitted-model uncertainty uses
$\mu(a)[b]$ as defined in Appendix~\ref{app:synthetic-protocol}.
Global is fitted under this study's budget, separately from the
allocation-study control. Bold/underline rank fitted-model means only;
the reference columns are unranked.}\label{tab:synthetic-final-complete}\\
\toprule
$s$ & $H$ & \shortstack{Exact\\shared} &
\shortstack{Exact\\channel} & Base & Global & PCR-$K4$ & PCR-$K2$ &
Learned-$K4$ & Learned-$K2$\\
\midrule
\endfirsthead
\multicolumn{10}{c}{\tablename\ \thetable\ continued}\\
\toprule
$s$ & $H$ & \shortstack{Exact\\shared} &
\shortstack{Exact\\channel} & Base & Global & PCR-$K4$ & PCR-$K2$ &
Learned-$K4$ & Learned-$K2$\\
\midrule
\endhead
\midrule
\multicolumn{10}{r}{\footnotesize Continued on the next page}\\
\endfoot
\bottomrule
\endlastfoot
\multicolumn{10}{@{}l}{\emph{DLinear}}\\*
0 & 96 & $0.92738$ & $0.90152$ & $0.9342(0)[34]$ & $0.9281(0)[34]$ & $\mathbf{0.9047}(1)[39]$ & $0.9082(3)[38]$ & $\underline{0.9050}(2)[38]$ & $0.9087(1)[37]$\\
0 & 192 & $0.95004$ & $0.92827$ & $0.9537(0)[33]$ & $0.9505(0)[33]$ & $\mathbf{0.9335}(5)[31]$ & $\underline{0.9359}(2)[36]$ & $\mathbf{0.9335}(4)[34]$ & $0.9366(1)[36]$\\
0 & 336 & $0.96869$ & $0.95405$ & $0.9709(0)[29]$ & $0.9691(0)[29]$ & $\underline{0.9587}(6)[25]$ & $0.9595(0)[30]$ & $\mathbf{0.9581}(3)[29]$ & $0.9599(1)[30]$\\
0 & 720 & $0.98513$ & $0.97799$ & $0.9863(0)[24]$ & $0.9854(0)[24]$ & $\underline{0.9807}(4)[23]$ & $0.9809(1)[25]$ & $\mathbf{0.9802}(1)[24]$ & $0.9811(1)[25]$\\
0.33 & 96 & $0.90762$ & $0.86314$ & $0.9147(0)[27]$ & $0.9084(0)[27]$ & $\mathbf{0.8675}(2)[31]$ & $0.8737(8)[28]$ & $\underline{0.8676}(2)[32]$ & $0.8741(2)[34]$\\
0.33 & 192 & $0.93161$ & $0.89243$ & $0.9356(0)[24]$ & $0.9322(0)[24]$ & $\mathbf{0.9000}(4)[28]$ & $0.9052(2)[30]$ & $\underline{0.9001}(2)[28]$ & $0.9065(1)[28]$\\
0.33 & 336 & $0.95398$ & $0.92478$ & $0.9565(0)[23]$ & $0.9544(0)[23]$ & $\underline{0.9327}(9)[14]$ & $0.9351(4)[26]$ & $\mathbf{0.9319}(4)[29]$ & $0.9361(8)[26]$\\
0.33 & 720 & $0.98084$ & $0.96558$ & $0.9823(0)[23]$ & $0.9813(0)[23]$ & $\underline{0.9707}(10)[28]$ & $0.9714(1)[25]$ & $\mathbf{0.9702}(2)[26]$ & $0.9718(1)[25]$\\
0.66 & 96 & $0.87273$ & $0.80701$ & $0.8803(0)[49]$ & $0.8738(0)[49]$ & $\underline{0.8129}(4)[36]$ & $0.8210(7)[30]$ & $\mathbf{0.8128}(2)[37]$ & $0.8228(2)[40]$\\
0.66 & 192 & $0.89527$ & $0.83440$ & $0.8998(0)[46]$ & $0.8961(0)[46]$ & $\mathbf{0.8448}(2)[42]$ & $0.8528(6)[42]$ & $\underline{0.8449}(1)[41]$ & $0.8551(3)[39]$\\
0.66 & 336 & $0.91767$ & $0.86718$ & $0.9207(0)[47]$ & $0.9184(0)[47]$ & $\underline{0.8787}(8)[43]$ & $0.8841(3)[44]$ & $\mathbf{0.8781}(1)[42]$ & $0.8856(1)[40]$\\
0.66 & 720 & $0.95308$ & $0.92274$ & $0.9551(0)[51]$ & $0.9539(0)[51]$ & $\underline{0.9320}(11)[42]$ & $0.9342(3)[44]$ & $\mathbf{0.9310}(3)[45]$ & $0.9351(1)[45]$\\
1 & 96 & $0.84101$ & $0.72198$ & $0.8505(0)[91]$ & $0.8435(1)[90]$ & $\underline{0.7325}(4)[66]$ & $0.7459(30)[86]$ & $\mathbf{0.7321}(3)[66]$ & $0.7487(4)[70]$\\
1 & 192 & $0.85716$ & $0.74021$ & $0.8635(0)[87]$ & $0.8592(0)[87]$ & $\underline{0.7657}(20)[76]$ & $0.7763(6)[71]$ & $\mathbf{0.7636}(7)[73]$ & $0.7790(3)[70]$\\
1 & 336 & $0.87053$ & $0.75939$ & $0.8754(0)[83]$ & $0.8724(0)[83]$ & $\underline{0.7887}(10)[56]$ & $0.7965(10)[65]$ & $\mathbf{0.7872}(4)[68]$ & $0.7983(3)[62]$\\
1 & 720 & $0.90293$ & $0.81628$ & $0.9069(0)[85]$ & $0.9052(0)[85]$ & $\underline{0.8447}(2)[74]$ & $0.8486(12)[69]$ & $\mathbf{0.8430}(6)[71]$ & $0.8501(4)[72]$\\
\addlinespace[3pt]
\multicolumn{10}{@{}l}{\emph{PatchTST}}\\*
0 & 96 & $0.92738$ & $0.90152$ & $0.9531(0)[36]$ & $0.9501(1)[35]$ & $\mathbf{0.9270}(2)[39]$ & $0.9312(4)[35]$ & $\underline{0.9274}(1)[39]$ & $0.9318(3)[39]$\\
0 & 192 & $0.95004$ & $0.92827$ & $0.9775(1)[37]$ & $0.9754(0)[36]$ & $\mathbf{0.9584}(4)[35]$ & $\underline{0.9616}(3)[34]$ & $\mathbf{0.9584}(1)[38]$ & $0.9623(1)[38]$\\
0 & 336 & $0.96869$ & $0.95405$ & $1.0003(0)[33]$ & $0.9989(0)[32]$ & $\underline{0.9880}(3)[35]$ & $0.9898(3)[33]$ & $\mathbf{0.9879}(1)[32]$ & $0.9902(1)[34]$\\
0 & 720 & $0.98513$ & $0.97799$ & $1.0252(0)[30]$ & $1.0243(0)[30]$ & $\underline{1.0193}(2)[29]$ & $1.0200(1)[30]$ & $\mathbf{1.0192}(1)[29]$ & $1.0202(1)[30]$\\
0.33 & 96 & $0.90762$ & $0.86314$ & $0.9293(1)[28]$ & $0.9254(1)[29]$ & $\mathbf{0.8847}(2)[34]$ & $0.8911(9)[38]$ & $\underline{0.8850}(2)[34]$ & $0.8921(2)[34]$\\
0.33 & 192 & $0.93161$ & $0.89243$ & $0.9539(1)[25]$ & $0.9507(0)[25]$ & $\mathbf{0.9187}(1)[30]$ & $0.9245(3)[28]$ & $\underline{0.9190}(1)[30]$ & $0.9259(1)[31]$\\
0.33 & 336 & $0.95398$ & $0.92478$ & $0.9787(0)[20]$ & $0.9764(0)[21]$ & $\underline{0.9541}(6)[23]$ & $0.9575(4)[26]$ & $\mathbf{0.9539}(3)[27]$ & $0.9585(7)[23]$\\
0.33 & 720 & $0.98084$ & $0.96558$ & $1.0132(0)[12]$ & $1.0118(0)[13]$ & $\underline{1.0009}(5)[14]$ & $1.0024(4)[14]$ & $\mathbf{1.0005}(1)[14]$ & $1.0027(0)[15]$\\
0.66 & 96 & $0.87273$ & $0.80701$ & $0.8919(1)[43]$ & $0.8868(1)[44]$ & $\mathbf{0.8262}(2)[29]$ & $0.8352(7)[37]$ & $\underline{0.8264}(1)[32]$ & $0.8372(3)[33]$\\
0.66 & 192 & $0.89527$ & $0.83440$ & $0.9138(1)[39]$ & $0.9094(0)[40]$ & $\mathbf{0.8589}(2)[34]$ & $\underline{0.8674}(7)[35]$ & $\mathbf{0.8589}(1)[34]$ & $0.8696(3)[34]$\\
0.66 & 336 & $0.91767$ & $0.86718$ & $0.9367(0)[41]$ & $0.9332(0)[42]$ & $\underline{0.8942}(3)[37]$ & $0.8996(2)[36]$ & $\mathbf{0.8936}(2)[35]$ & $0.9015(2)[34]$\\
0.66 & 720 & $0.95308$ & $0.92274$ & $0.9778(0)[47]$ & $0.9756(0)[48]$ & $\underline{0.9533}(4)[41]$ & $0.9564(3)[40]$ & $\mathbf{0.9530}(3)[44]$ & $0.9572(1)[40]$\\
1 & 96 & $0.84101$ & $0.72198$ & $0.8578(1)[84]$ & $0.8519(1)[83]$ & $\mathbf{0.7412}(3)[58]$ & $0.7544(9)[64]$ & $\underline{0.7416}(5)[59]$ & $0.7579(5)[63]$\\
1 & 192 & $0.85716$ & $0.74021$ & $0.8724(0)[84]$ & $0.8666(1)[83]$ & $\underline{0.7731}(16)[71]$ & $0.7845(7)[68]$ & $\mathbf{0.7715}(1)[66]$ & $0.7879(4)[64]$\\
1 & 336 & $0.87053$ & $0.75939$ & $0.8849(1)[82]$ & $0.8791(0)[81]$ & $\underline{0.7963}(8)[69]$ & $0.8038(7)[66]$ & $\mathbf{0.7948}(6)[72]$ & $0.8064(7)[61]$\\
1 & 720 & $0.90293$ & $0.81628$ & $0.9185(1)[89]$ & $0.9138(0)[86]$ & $\underline{0.8536}(6)[74]$ & $0.8572(4)[76]$ & $\mathbf{0.8521}(5)[71]$ & $0.8593(4)[77]$\\
\addlinespace[3pt]
\multicolumn{10}{@{}l}{\emph{iTransformer}}\\*
0 & 96 & $0.92738$ & $0.90152$ & $0.9522(1)[35]$ & $0.9508(1)[35]$ & $\mathbf{0.9277}(2)[39]$ & $0.9309(3)[40]$ & $\underline{0.9281}(2)[38]$ & $0.9314(2)[37]$\\
0 & 192 & $0.95004$ & $0.92827$ & $0.9767(1)[36]$ & $0.9757(0)[36]$ & $\mathbf{0.9588}(3)[40]$ & $0.9615(4)[36]$ & $\underline{0.9589}(2)[37]$ & $0.9621(2)[39]$\\
0 & 336 & $0.96869$ & $0.95405$ & $0.9997(0)[32]$ & $0.9991(0)[32]$ & $\mathbf{0.9880}(2)[33]$ & $0.9897(2)[34]$ & $\underline{0.9882}(2)[33]$ & $0.9901(1)[34]$\\
0 & 720 & $0.98513$ & $0.97799$ & $1.0248(0)[29]$ & $1.0244(0)[30]$ & $\underline{1.0195}(4)[30]$ & $1.0200(1)[29]$ & $\mathbf{1.0193}(1)[30]$ & $1.0202(0)[30]$\\
0.33 & 96 & $0.90762$ & $0.86314$ & $0.9276(1)[29]$ & $0.9261(1)[29]$ & $\mathbf{0.8854}(1)[33]$ & $0.8912(16)[34]$ & $\underline{0.8858}(1)[33]$ & $0.8919(2)[35]$\\
0.33 & 192 & $0.93161$ & $0.89243$ & $0.9522(0)[25]$ & $0.9512(0)[25]$ & $\underline{0.9195}(7)[33]$ & $0.9243(3)[29]$ & $\mathbf{0.9192}(1)[29]$ & $0.9255(2)[32]$\\
0.33 & 336 & $0.95398$ & $0.92478$ & $0.9774(0)[20]$ & $0.9766(0)[20]$ & $\mathbf{0.9540}(6)[24]$ & $\underline{0.9572}(5)[26]$ & $\mathbf{0.9540}(1)[23]$ & $0.9580(1)[25]$\\
0.33 & 720 & $0.98084$ & $0.96558$ & $1.0124(0)[13]$ & $1.0120(0)[13]$ & $\underline{1.0011}(5)[13]$ & $1.0022(2)[15]$ & $\mathbf{1.0007}(2)[14]$ & $1.0029(3)[15]$\\
0.66 & 96 & $0.87273$ & $0.80701$ & $0.8894(1)[45]$ & $0.8876(1)[44]$ & $\mathbf{0.8271}(3)[34]$ & $0.8346(5)[36]$ & $\underline{0.8274}(1)[32]$ & $0.8367(4)[33]$\\
0.66 & 192 & $0.89527$ & $0.83440$ & $0.9112(0)[41]$ & $0.9099(0)[41]$ & $\mathbf{0.8590}(3)[34]$ & $0.8672(3)[35]$ & $\underline{0.8593}(1)[34]$ & $0.8695(12)[37]$\\
0.66 & 336 & $0.91767$ & $0.86718$ & $0.9345(0)[43]$ & $0.9336(0)[42]$ & $\underline{0.8942}(11)[38]$ & $0.8992(4)[36]$ & $\mathbf{0.8937}(1)[35]$ & $0.9005(1)[35]$\\
0.66 & 720 & $0.95308$ & $0.92274$ & $0.9764(0)[48]$ & $0.9758(0)[48]$ & $\underline{0.9536}(4)[36]$ & $0.9560(5)[44]$ & $\mathbf{0.9530}(1)[42]$ & $0.9569(1)[42]$\\
1 & 96 & $0.84101$ & $0.72198$ & $0.8544(1)[85]$ & $0.8527(1)[85]$ & $\mathbf{0.7418}(2)[59]$ & $0.7536(14)[57]$ & $\underline{0.7425}(4)[59]$ & $0.7562(4)[63]$\\
1 & 192 & $0.85716$ & $0.74021$ & $0.8685(1)[83]$ & $0.8671(0)[83]$ & $\mathbf{0.7713}(12)[68]$ & $0.7832(13)[65]$ & $\underline{0.7715}(3)[68]$ & $0.7857(6)[64]$\\
1 & 336 & $0.87053$ & $0.75939$ & $0.8807(0)[81]$ & $0.8796(0)[81]$ & $\underline{0.7956}(15)[62]$ & $0.8016(6)[59]$ & $\mathbf{0.7936}(3)[66]$ & $0.8043(6)[63]$\\
1 & 720 & $0.90293$ & $0.81628$ & $0.9148(0)[86]$ & $0.9140(0)[87]$ & $\underline{0.8528}(8)[71]$ & $0.8558(1)[73]$ & $\mathbf{0.8512}(5)[74]$ & $0.8581(4)[72]$\\
\addlinespace[3pt]
\end{longtable}
\endgroup

\subsubsection{Do Channel Assignments Matter?}
\label{app:synthetic-perm}

At $s\in\{0,1\}$ and $H\in\{96,720\}$, we shuffle whole fitted
mixture vectors across channels twenty times, keeping the experts,
backbone, and correction scale fixed without refitting or recalibration.
The four PCR/learned variants cover all backbones, realizations, and
training seeds: 432 checkpoints and 8,640 permutations.
Every permutation increases MSE (Table~\ref{tab:synthetic-permutation}).
At $s=1,H=96$, PCR-K4's 13.55\% mean backbone improvement becomes a
12.58\% degradation; error rises 30.23\% relative to the corrected
model. Thus the fitted channel assignments, not only the expert bank,
matter to forecasting accuracy.

\begingroup
\scriptsize
\setlength{\tabcolsep}{3pt}
\begin{longtable}{@{}rrrrrr@{}}
\caption{Effect of permuting fitted routing vectors on all endpoint settings: relative MSE increase (\%), followed in parentheses by mean within-checkpoint MSE SD across permutations in units of $10^{-4}$. Each entry averages nine checkpoints and twenty permutations per checkpoint; every permutation increases MSE.}\label{tab:synthetic-permutation}\\
\toprule
$s$ & $H$ & PCR-$K4$ & PCR-$K2$ & Learned-$K4$ & Learned-$K2$\\
\midrule
\endfirsthead
\multicolumn{6}{c}{\tablename\ \thetable\ continued}\\
\toprule
$s$ & $H$ & PCR-$K4$ & PCR-$K2$ & Learned-$K4$ & Learned-$K2$\\
\midrule
\endhead
\midrule
\multicolumn{6}{r}{\footnotesize Continued on the next page}\\
\endfoot
\bottomrule
\endlastfoot
\multicolumn{6}{@{}l}{\emph{DLinear}}\\*
0 & 96 & 5.24 (19.7) & 4.50 (18.1) & 5.16 (19.5) & 4.37 (17.2)\\
0 & 720 & 1.00 (4.6) & 0.96 (4.3) & 1.09 (4.6) & 0.92 (4.1)\\
1 & 96 & 30.63 (60.0) & 26.74 (54.0) & 30.91 (60.3) & 26.01 (55.1)\\
1 & 720 & 14.66 (28.7) & 13.84 (29.5) & 15.26 (30.4) & 13.46 (29.3)\\
\addlinespace[3pt]
\multicolumn{6}{@{}l}{\emph{PatchTST}}\\*
0 & 96 & 4.99 (19.2) & 4.19 (18.2) & 4.92 (19.1) & 4.13 (17.0)\\
0 & 720 & 1.01 (4.8) & 0.89 (4.2) & 1.03 (4.6) & 0.86 (3.9)\\
1 & 96 & 29.99 (59.4) & 26.61 (55.3) & 30.14 (59.6) & 25.88 (53.9)\\
1 & 720 & 14.59 (28.7) & 13.68 (29.9) & 15.14 (30.4) & 13.21 (29.0)\\
\addlinespace[3pt]
\multicolumn{6}{@{}l}{\emph{iTransformer}}\\*
0 & 96 & 5.00 (19.3) & 4.41 (18.0) & 4.88 (18.9) & 4.28 (17.4)\\
0 & 720 & 0.97 (4.6) & 0.89 (4.1) & 1.01 (4.4) & 0.86 (3.8)\\
1 & 96 & 30.07 (59.7) & 26.83 (56.4) & 30.15 (59.2) & 26.12 (53.7)\\
1 & 720 & 14.77 (29.3) & 14.01 (30.2) & 15.20 (30.5) & 13.43 (29.8)\\
\addlinespace[3pt]
\end{longtable}
\endgroup

\subsubsection{Fitting and Variability}
Eight of 864 PCR fits reach the 200-epoch cap: three PCR-K2 and five
PCR-K4. All 864 learned-allocation and 432 Global fits stop by the
plateau rule, which does not establish convergence.
PCR has lower within-realization test-MSE seed SD than its same-$K$
learned comparator in 11/48 settings for $K2$ and 7/48 for $K4$.
These results do not establish a uniform stability advantage.
Table~\ref{tab:synthetic-optimization} gives endpoint stopping summaries.

Table~\ref{tab:synthetic-timing} reports 24 standalone fits on exclusive
RTX 3090 GPUs, one realization and training seed per condition.
Method time measures synchronized computation, excluding input fetching,
feature-cache construction, and statistics; elapsed time includes the
complete run. PCR-K2 has lower method time than Learned-K2 in five of
six settings and lower elapsed time in four. PCR-K4 has lower method
time in four settings but higher elapsed time in all six.
These costs depend on whether setup is included, and the optimizer
policies differ; they do not establish a universal fitting-time advantage.

\begingroup
\scriptsize
\setlength{\tabcolsep}{3pt}
\begin{longtable}{@{}lrr@{}}
\caption{GalaxyTS stopping summaries over four endpoint settings per
backbone. Epoch ranges span setting summaries, each the median of three
within-realization medians. Cap hits count individual fits across
realizations and seeds; two of the full grid's eight PCR cap hits occur
here. Seed variability appears in Table~\ref{tab:synthetic-final-complete}.}\label{tab:synthetic-optimization}\\
\toprule
Method & Run epochs (range) & Cap hits\\
\midrule
\endfirsthead
\multicolumn{3}{c}{\tablename\ \thetable\ continued}\\
\toprule
Method & Run epochs (range) & Cap hits\\
\midrule
\endhead
\midrule
\multicolumn{3}{r}{\footnotesize Continued on the next page}\\
\endfoot
\bottomrule
\endlastfoot
\multicolumn{3}{@{}l}{\emph{DLinear}}\\*
Global correction & 68-90 & 0\\
PCR-$K4$ & 78-152 & 0\\
PCR-$K2$ & 72-120 & 1\\
Learned-$K4$ & 70-122 & 0\\
Learned-$K2$ & 76-94 & 0\\
\addlinespace[3pt]
\multicolumn{3}{@{}l}{\emph{PatchTST}}\\*
Global correction & 68-76 & 0\\
PCR-$K4$ & 80-118 & 0\\
PCR-$K2$ & 68-106 & 1\\
Learned-$K4$ & 74-116 & 0\\
Learned-$K2$ & 76-96 & 0\\
\addlinespace[3pt]
\multicolumn{3}{@{}l}{\emph{iTransformer}}\\*
Global correction & 72-84 & 0\\
PCR-$K4$ & 80-136 & 0\\
PCR-$K2$ & 70-110 & 0\\
Learned-$K4$ & 82-120 & 0\\
Learned-$K2$ & 86-96 & 0\\
\addlinespace[3pt]
\end{longtable}
\endgroup

\begingroup
\scriptsize
\setlength{\tabcolsep}{3pt}
\begin{longtable}{@{}lrrrrrr@{}}
\caption{GalaxyTS fitting costs at $s=1$, realization 1, training seed
0, on exclusive RTX 3090 GPUs. Entries give method / complete elapsed
time (s); peak memory is the range across four methods (GiB).
Each entry is one fit without a variance estimate; all stop by the
plateau rule.}\label{tab:synthetic-timing}\\
\toprule
Backbone & $H$ & PCR-$K4$ & PCR-$K2$ & Learned-$K4$ & Learned-$K2$ & Peak\\
\midrule
\endfirsthead
\multicolumn{7}{c}{\tablename\ \thetable\ continued}\\
\toprule
Backbone & $H$ & PCR-$K4$ & PCR-$K2$ & Learned-$K4$ & Learned-$K2$ & Peak\\
\midrule
\endhead
\midrule
\multicolumn{7}{r}{\footnotesize Continued on the next page}\\
\endfoot
\bottomrule
\endlastfoot
DLinear & 96 & 237.9 / 385.0 & 214.0 / 349.1 & 230.7 / 329.4 & 306.2 / 410.3 & 0.134\\
DLinear & 720 & 157.7 / 338.6 & 101.3 / 239.2 & 215.8 / 335.8 & 221.0 / 318.9 & 0.539\\
PatchTST & 96 & 149.9 / 1501.6 & 141.9 / 1531.5 & 156.5 / 637.0 & 165.9 / 1197.3 & 1.340\\
PatchTST & 720 & 114.8 / 1405.7 & 98.7 / 696.1 & 169.7 / 1345.9 & 160.1 / 696.9 & 1.663-1.664\\
iTransformer & 96 & 404.6 / 599.0 & 237.3 / 358.7 & 239.4 / 318.8 & 380.7 / 478.1 & 12.498\\
iTransformer & 720 & 157.7 / 347.9 & 158.6 / 308.6 & 208.0 / 308.6 & 139.5 / 216.3 & 10.995\\
\addlinespace[3pt]
\end{longtable}
\endgroup

\subsection{Forecasting Accuracy on Real Data}
\label{app:pcr-coverage}
\label{app:pcr-fullgrid}

The benchmark uses three backbones, eleven datasets,
$H\in\{96,192,336,720\}$, $L=96$, and three seeds:
132 settings and 1,584 fits of PCR-K4, PCR-K2, Learned-K4, and Learned-K2.
Splits follow Appendix~\ref{app:datasets}; all four methods share the
frozen checkpoint within each setting and seed.

DLinear and PatchTST bases
at $H=96,720$ are shared with the allocation study; bases at $H=192,336$
use seed-0 base-learning-rate selection. The iTransformer bases are
trained for this benchmark.

\paragraph{Reading the results.}
Backbone-relative MSE/MAE reductions are computed per seed, then
averaged over seeds, horizons within datasets, and the eleven datasets.
Setting-level tables report mean $\pm$ sample SD over three seeds.
Aggregate 95\% bootstrap intervals resample dataset means, not seeds
or horizons; they describe dataset variability rather than establishing
statistical significance.

\paragraph{Where improvements are consistent, and where they are not.}
Positive mean MSE gains for PCR-K4, PCR-K2, Learned-K4, and Learned-K2
occur in respectively 40/44, 38/44,
40/44, and 41/44 for DLinear; 42/44, 41/44, 42/44, and 43/44 for
PatchTST; and 43/44, 44/44, 43/44, and 42/44 for iTransformer.
Absolute mean gains below $10^{-6}\%$ count as numerical ties.
Requiring \emph{every seed} to improve gives 33/33/33/34 for
DLinear, 39/39/39/41 for PatchTST, and 42/42/42/41 for iTransformer,
in the same method order and each out of 44 settings.

Regressions remain in all aggregates. The largest PCR mean regression
is DLinear on ETTh2 at $H=720$: $-7.26\%$ for PCR-K4 and $-7.16\%$
for PCR-K2, with all three seeds worse than their frozen base.
The corresponding Learned-K4 and Learned-K2 gains are $-1.18\%$
and $+0.68\%$. DLinear on ETTm2 at $H=336$ recovers the frozen
forecast to numerical precision for every arm. Including $\alpha=0$
in validation calibration therefore does not guarantee nonnegative
test improvement.

\par\begingroup
\subsubsection{Comparison with Matched Shared Capacity}
\label{app:pcr-global-control}

To test whether PCR gains reflect additional shared capacity, we fit one
rank-64 Global correction on the same frozen checkpoints, matching PCR's
$4\times16(F+H)=64(F+H)$ expert-factor budget. The control follows the
same final optimization and model-selection protocol, including the
learning-rate grid and validation-selected correction scale: eleven
datasets, four horizons, and three seeds per backbone (396 completed fits).
It is separate from the allocation study's earlier Global control.

Table~\ref{tab:pcr-global-control} includes the dataset-balanced relative
MSE reductions for this comparison. PCR-K4 improves on Global by
$1.87\%$, $9.19\%$, and $9.82\%$ on DLinear, PatchTST, and iTransformer,
respectively; PCR-K2 also improves each backbone average. The averages
retain regressions: Global is better on DLinear's ETTh2 and ETTm2 at
all four horizons. The smaller DLinear average therefore does not imply
an advantage on every dataset.
\par\endgroup

\subsubsection{PCR versus Learned Allocation, and K2 versus K4}
\label{app:pcr-contrasts}

Table~\ref{tab:pcr-final-contrasts} compares methods on identical
settings and seeds. Relative MSE reduction is
$100(1-\mathrm{MSE}_A/\mathrm{MSE}_B)$, positive for an improvement
by $A$, averaged over seeds, horizons within datasets, and datasets.
PCR is stronger on average on PatchTST and iTransformer; DLinear
differences are small and of either sign. Restricting either estimator
to two active experts increases average error. This is not a uniform
ordering across settings, and PCR and learned allocation have different
optimizer policies.

\begin{table}[!htb]
\centering\small
\setlength{\tabcolsep}{5pt}
\caption{Paired relative MSE reductions (\%); positive values favor the first method. Means weight datasets equally; the upper panel's 95\% percentile intervals use 100,000 bootstrap resamples of the eleven dataset means. Its last column counts improvements in all three seeds, out of 44 settings. The lower panel reports final-protocol Global-control comparisons on the same 44 settings and three seeds per backbone.}
\label{tab:pcr-final-contrasts}\label{tab:pcr-global-control}
\begin{tabular}{@{}llrrr@{}}
\toprule
Backbone & Comparison & Mean (\%) & 95\% interval & Positive in 3/3 \\
\midrule
DLinear & PCR versus learned, $K4$ & $-0.07$ & $[-0.50, +0.32]$ & 12/44 \\
 & PCR versus learned, $K2$ & $+0.02$ & $[-0.51, +0.40]$ & 18/44 \\
 & $K2$ versus $K4$, PCR & $-0.25$ & $[-0.43, -0.03]$ & 3/44 \\
 & $K2$ versus $K4$, learned & $-0.33$ & $[-0.61, -0.04]$ & 3/44 \\
\addlinespace[2pt]
PatchTST & PCR versus learned, $K4$ & $+0.38$ & $[+0.05, +0.75]$ & 19/44 \\
 & PCR versus learned, $K2$ & $+0.20$ & $[-0.36, +0.88]$ & 13/44 \\
 & $K2$ versus $K4$, PCR & $-1.41$ & $[-2.57, -0.35]$ & 6/44 \\
 & $K2$ versus $K4$, learned & $-1.24$ & $[-2.79, +0.02]$ & 8/44 \\
\addlinespace[2pt]
iTransformer & PCR versus learned, $K4$ & $+0.22$ & $[+0.14, +0.31]$ & 15/44 \\
 & PCR versus learned, $K2$ & $+0.79$ & $[+0.30, +1.33]$ & 28/44 \\
 & $K2$ versus $K4$, PCR & $-2.24$ & $[-4.30, -0.52]$ & 3/44 \\
 & $K2$ versus $K4$, learned & $-2.86$ & $[-5.50, -0.60]$ & 3/44 \\
\addlinespace[2pt]
\bottomrule
\end{tabular}
\par\vspace{6pt}
\par\begingroup
\begin{tabular}{@{}lrrr@{}}
\toprule
Backbone & Global vs. base & PCR-K4 vs. Global & PCR-K2 vs. Global \\
\midrule
DLinear & 1.55 & 1.87 & 1.66 \\
PatchTST & 6.10 & 9.19 & 8.12 \\
iTransformer & 0.47 & 9.82 & 8.15 \\
\bottomrule
\end{tabular}
\par\endgroup
\end{table}

\subsubsection{Training Duration and Checkpoint Selection}
\label{app:pcr-budget}

Table~\ref{tab:pcr-final-budgets} separates the selected checkpoint from
total training duration. Seven fits reach the 200-epoch cap:
iTransformer PCR-K2 on PEMS04 ($H=192$, seed 0),
PEMS07 ($H=192$, seed 1), and PEMS07 ($H=96$, seed 0); PatchTST
PCR-K2 on PEMS07 ($H=192$, seed 0) and PEMS07 ($H=336$, seed 1);
and PatchTST PCR-K4 on PEMS07 ($H=192$, seed 0) and
PEMS08 ($H=96$, seed 2). Their selected results remain included;
all other fits stop by the plateau rule.

The early-checkpoint column selects only eligible epochs $\leq30$
from these same trajectories; it is not a separate fixed-budget or
equal-time experiment.

\begin{table}[!htb]
\centering\small
\setlength{\tabcolsep}{5pt}
\caption{Training diagnostics over 132 runs per backbone and arm: median epochs, cap counts, and dataset-balanced MSE reductions (\%). Bold/underline mark the best/second distinct gain within each backbone and column; ties share rank.}
\label{tab:pcr-final-budgets}
\begin{tabular}{@{}llrrrrr@{}}
\toprule
Backbone & Arm & Selected epoch & Epochs run & Caps & Gain, $\leq30$ & Final gain \\
\midrule
DLinear & PCR-$K4$ & 61 & 82 & 0 & \textbf{3.65}\% & \underline{3.86}\% \\
 & PCR-$K2$ & 47 & 80 & 0 & 3.49\% & 3.63\% \\
 & Learned-$K4$ & 48 & 80 & 0 & \underline{3.53}\% & \textbf{3.92}\% \\
 & Learned-$K2$ & 65 & 86 & 0 & 3.14\% & 3.62\% \\
\addlinespace[2pt]
PatchTST & PCR-$K4$ & 86 & 102 & 2 & \textbf{13.45}\% & \textbf{14.25}\% \\
 & PCR-$K2$ & 76 & 99 & 2 & \underline{12.54}\% & 13.31\% \\
 & Learned-$K4$ & 77 & 98 & 0 & 12.49\% & \underline{13.98}\% \\
 & Learned-$K2$ & 75 & 92 & 0 & 11.39\% & 13.20\% \\
\addlinespace[2pt]
iTransformer & PCR-$K4$ & 83 & 102 & 0 & \textbf{9.69}\% & \textbf{10.22}\% \\
 & PCR-$K2$ & 67 & 96 & 3 & 8.10\% & 8.56\% \\
 & Learned-$K4$ & 70 & 92 & 0 & \underline{8.96}\% & \underline{10.02}\% \\
 & Learned-$K2$ & 71 & 90 & 0 & 6.94\% & 7.91\% \\
\addlinespace[2pt]
\bottomrule
\end{tabular}
\end{table}

\subsubsection{Full Benchmark Results}
\label{app:pcr-full-grid-tables}

Tables~\ref{tab:pcr-full-dlinear-mse}-\ref{tab:pcr-full-itransformer-tslib-mse}
report test MSE for all 132 settings. Entries give mean $\pm$ sample SD
over three seeds. Within each row, bold and underlined means identify
the lowest and second distinct values, including the frozen base;
ties at four-decimal precision share rank.
Table~\ref{tab:pcr-mae-summary} summarizes MAE reductions.

\begingroup
\small
\setlength{\tabcolsep}{5pt}
\renewcommand{\arraystretch}{1.05}
\begin{longtable}{@{}lrccccc@{}}
\caption{DLinear: test MSE, mean $\pm$ sample SD over three seeds.}
\label{tab:pcr-full-dlinear-mse}\\
\toprule
Dataset & $H$ & Frozen base & PCR-$K4$ & PCR-$K2$ & Learned-$K4$ & Learned-$K2$ \\
\midrule
\endfirsthead
\multicolumn{7}{c}{\tablename\ \thetable\ continued: DLinear, MSE}\\
\toprule
Dataset & $H$ & Frozen base & PCR-$K4$ & PCR-$K2$ & Learned-$K4$ & Learned-$K2$ \\
\midrule
\endhead
\midrule
\multicolumn{7}{r}{\footnotesize Continued on the next page}\\
\endfoot
\bottomrule
\endlastfoot
ETTh1 & 96 & $0.3860 \pm 0.0031$ & $\mathbf{0.3793} \pm 0.0004$ & $0.3810 \pm 0.0027$ & $0.3828 \pm 0.0036$ & $\underline{0.3808} \pm 0.0010$ \\
ETTh1 & 192 & $0.4512 \pm 0.0301$ & $\underline{0.4404} \pm 0.0114$ & $\mathbf{0.4395} \pm 0.0108$ & $0.4465 \pm 0.0019$ & $0.4444 \pm 0.0035$ \\
ETTh1 & 336 & $0.4994 \pm 0.0249$ & $\mathbf{0.4814} \pm 0.0031$ & $0.4832 \pm 0.0041$ & $0.4843 \pm 0.0032$ & $\underline{0.4824} \pm 0.0025$ \\
ETTh1 & 720 & $0.5363 \pm 0.0294$ & $\mathbf{0.5081} \pm 0.0047$ & $\underline{0.5115} \pm 0.0050$ & $0.5174 \pm 0.0031$ & $0.5213 \pm 0.0056$ \\
\addlinespace[3pt]
ETTh2 & 96 & $0.3208 \pm 0.0397$ & $0.3223 \pm 0.0422$ & $0.3212 \pm 0.0404$ & $\underline{0.3190} \pm 0.0365$ & $\mathbf{0.3175} \pm 0.0340$ \\
ETTh2 & 192 & $\mathbf{0.4372} \pm 0.0065$ & $\mathbf{0.4372} \pm 0.0065$ & $\mathbf{0.4372} \pm 0.0065$ & $0.4381 \pm 0.0063$ & $\underline{0.4380} \pm 0.0063$ \\
ETTh2 & 336 & $0.5043 \pm 0.0152$ & $\underline{0.5016} \pm 0.0191$ & $0.5023 \pm 0.0180$ & $\mathbf{0.5013} \pm 0.0196$ & $0.5022 \pm 0.0181$ \\
ETTh2 & 720 & $\underline{0.7485} \pm 0.0564$ & $0.8044 \pm 0.0920$ & $0.8026 \pm 0.0721$ & $0.7561 \pm 0.0343$ & $\mathbf{0.7431} \pm 0.0528$ \\
\addlinespace[3pt]
ETTm1 & 96 & $0.3438 \pm 0.0014$ & $0.3279 \pm 0.0002$ & $\underline{0.3260} \pm 0.0023$ & $\mathbf{0.3242} \pm 0.0010$ & $0.3262 \pm 0.0024$ \\
ETTm1 & 192 & $0.3819 \pm 0.0011$ & $0.3696 \pm 0.0023$ & $\mathbf{0.3670} \pm 0.0005$ & $\underline{0.3671} \pm 0.0009$ & $0.3675 \pm 0.0014$ \\
ETTm1 & 336 & $0.4130 \pm 0.0003$ & $0.4037 \pm 0.0010$ & $\underline{0.4005} \pm 0.0010$ & $\mathbf{0.4001} \pm 0.0029$ & $0.4036 \pm 0.0015$ \\
ETTm1 & 720 & $0.4751 \pm 0.0029$ & $0.4647 \pm 0.0024$ & $\mathbf{0.4634} \pm 0.0028$ & $\underline{0.4643} \pm 0.0031$ & $0.4670 \pm 0.0051$ \\
\addlinespace[3pt]
ETTm2 & 96 & $0.1855 \pm 0.0032$ & $0.1854 \pm 0.0028$ & $0.1856 \pm 0.0038$ & $\underline{0.1847} \pm 0.0031$ & $\mathbf{0.1832} \pm 0.0018$ \\
ETTm2 & 192 & $\mathbf{0.2663} \pm 0.0053$ & $\mathbf{0.2663} \pm 0.0055$ & $0.2672 \pm 0.0056$ & $\mathbf{0.2663} \pm 0.0048$ & $\underline{0.2664} \pm 0.0042$ \\
ETTm2 & 336 & $\mathbf{0.3315} \pm 0.0229$ & $\mathbf{0.3315} \pm 0.0229$ & $\mathbf{0.3315} \pm 0.0229$ & $\mathbf{0.3315} \pm 0.0229$ & $\mathbf{0.3315} \pm 0.0229$ \\
ETTm2 & 720 & $0.4912 \pm 0.0428$ & $0.4785 \pm 0.0493$ & $0.4764 \pm 0.0468$ & $\underline{0.4751} \pm 0.0490$ & $\mathbf{0.4711} \pm 0.0429$ \\
\addlinespace[3pt]
Weather & 96 & $0.1961 \pm 0.0010$ & $\mathbf{0.1613} \pm 0.0006$ & $0.1638 \pm 0.0044$ & $\underline{0.1616} \pm 0.0010$ & $0.1662 \pm 0.0060$ \\
Weather & 192 & $0.2365 \pm 0.0011$ & $0.2050 \pm 0.0018$ & $\underline{0.2035} \pm 0.0033$ & $\mathbf{0.2031} \pm 0.0013$ & $0.2043 \pm 0.0010$ \\
Weather & 336 & $0.2840 \pm 0.0033$ & $\underline{0.2577} \pm 0.0012$ & $0.2615 \pm 0.0043$ & $\mathbf{0.2576} \pm 0.0020$ & $0.2581 \pm 0.0028$ \\
Weather & 720 & $0.3478 \pm 0.0028$ & $0.3318 \pm 0.0007$ & $\mathbf{0.3294} \pm 0.0066$ & $0.3379 \pm 0.0041$ & $\underline{0.3306} \pm 0.0036$ \\
\addlinespace[3pt]
Electricity & 96 & $0.1945 \pm 0.0000$ & $\mathbf{0.1872} \pm 0.0000$ & $0.1886 \pm 0.0000$ & $\underline{0.1876} \pm 0.0005$ & $0.1893 \pm 0.0005$ \\
Electricity & 192 & $0.1938 \pm 0.0001$ & $\underline{0.1871} \pm 0.0003$ & $0.1876 \pm 0.0001$ & $\mathbf{0.1869} \pm 0.0006$ & $0.1885 \pm 0.0003$ \\
Electricity & 336 & $0.2068 \pm 0.0002$ & $\mathbf{0.1996} \pm 0.0001$ & $0.2008 \pm 0.0005$ & $\underline{0.2000} \pm 0.0004$ & $0.2012 \pm 0.0004$ \\
Electricity & 720 & $0.2426 \pm 0.0004$ & $\underline{0.2358} \pm 0.0002$ & $0.2367 \pm 0.0006$ & $\mathbf{0.2357} \pm 0.0005$ & $0.2364 \pm 0.0000$ \\
\addlinespace[3pt]
Traffic & 96 & $0.6487 \pm 0.0002$ & $\underline{0.6396} \pm 0.0003$ & $\underline{0.6396} \pm 0.0002$ & $\mathbf{0.6391} \pm 0.0005$ & $0.6398 \pm 0.0005$ \\
Traffic & 192 & $0.5983 \pm 0.0001$ & $0.5915 \pm 0.0010$ & $0.5914 \pm 0.0009$ & $\mathbf{0.5897} \pm 0.0005$ & $\underline{0.5909} \pm 0.0007$ \\
Traffic & 336 & $0.6052 \pm 0.0001$ & $\underline{0.5973} \pm 0.0005$ & $\underline{0.5973} \pm 0.0003$ & $\mathbf{0.5969} \pm 0.0007$ & $0.5977 \pm 0.0005$ \\
Traffic & 720 & $0.6454 \pm 0.0001$ & $0.6352 \pm 0.0003$ & $0.6361 \pm 0.0010$ & $\mathbf{0.6341} \pm 0.0011$ & $\underline{0.6345} \pm 0.0005$ \\
\addlinespace[3pt]
PEMS03 & 96 & $0.4413 \pm 0.0030$ & $\underline{0.4273} \pm 0.0023$ & $0.4299 \pm 0.0026$ & $\mathbf{0.4272} \pm 0.0017$ & $0.4293 \pm 0.0014$ \\
PEMS03 & 192 & $0.4649 \pm 0.0003$ & $\mathbf{0.4523} \pm 0.0007$ & $0.4541 \pm 0.0006$ & $\underline{0.4528} \pm 0.0001$ & $0.4539 \pm 0.0003$ \\
PEMS03 & 336 & $0.3856 \pm 0.0004$ & $\underline{0.3767} \pm 0.0001$ & $0.3781 \pm 0.0004$ & $\mathbf{0.3762} \pm 0.0012$ & $0.3787 \pm 0.0002$ \\
PEMS03 & 720 & $0.4306 \pm 0.0002$ & $\mathbf{0.4227} \pm 0.0003$ & $0.4236 \pm 0.0002$ & $\underline{0.4230} \pm 0.0008$ & $0.4245 \pm 0.0005$ \\
\addlinespace[3pt]
PEMS04 & 96 & $0.4148 \pm 0.0004$ & $\mathbf{0.3945} \pm 0.0003$ & $0.3975 \pm 0.0001$ & $\underline{0.3959} \pm 0.0016$ & $0.3990 \pm 0.0010$ \\
PEMS04 & 192 & $0.4371 \pm 0.0037$ & $\mathbf{0.4165} \pm 0.0006$ & $0.4198 \pm 0.0011$ & $\underline{0.4181} \pm 0.0005$ & $0.4210 \pm 0.0007$ \\
PEMS04 & 336 & $0.3718 \pm 0.0001$ & $\mathbf{0.3592} \pm 0.0003$ & $0.3611 \pm 0.0002$ & $\underline{0.3596} \pm 0.0006$ & $0.3623 \pm 0.0005$ \\
PEMS04 & 720 & $0.4347 \pm 0.0006$ & $\mathbf{0.4188} \pm 0.0004$ & $0.4205 \pm 0.0004$ & $\underline{0.4198} \pm 0.0007$ & $0.4219 \pm 0.0003$ \\
\addlinespace[3pt]
PEMS07 & 96 & $0.5632 \pm 0.0008$ & $\underline{0.5237} \pm 0.0001$ & $0.5271 \pm 0.0015$ & $\mathbf{0.5226} \pm 0.0025$ & $0.5325 \pm 0.0005$ \\
PEMS07 & 192 & $0.5973 \pm 0.0003$ & $\mathbf{0.5556} \pm 0.0004$ & $0.5599 \pm 0.0006$ & $\underline{0.5576} \pm 0.0015$ & $0.5621 \pm 0.0021$ \\
PEMS07 & 336 & $0.4765 \pm 0.0028$ & $\underline{0.4510} \pm 0.0022$ & $0.4527 \pm 0.0003$ & $\mathbf{0.4506} \pm 0.0004$ & $0.4538 \pm 0.0009$ \\
PEMS07 & 720 & $0.5439 \pm 0.0004$ & $\underline{0.5183} \pm 0.0003$ & $0.5206 \pm 0.0007$ & $\mathbf{0.5179} \pm 0.0004$ & $0.5230 \pm 0.0023$ \\
\addlinespace[3pt]
PEMS08 & 96 & $0.5086 \pm 0.0042$ & $\mathbf{0.4442} \pm 0.0004$ & $0.4479 \pm 0.0014$ & $\underline{0.4470} \pm 0.0016$ & $0.4507 \pm 0.0052$ \\
PEMS08 & 192 & $0.5454 \pm 0.0002$ & $\mathbf{0.4909} \pm 0.0005$ & $0.4942 \pm 0.0002$ & $\underline{0.4936} \pm 0.0007$ & $0.4957 \pm 0.0015$ \\
PEMS08 & 336 & $0.4743 \pm 0.0019$ & $\mathbf{0.4429} \pm 0.0007$ & $\mathbf{0.4429} \pm 0.0018$ & $\underline{0.4440} \pm 0.0051$ & $0.4492 \pm 0.0025$ \\
PEMS08 & 720 & $0.5413 \pm 0.0010$ & $\mathbf{0.5113} \pm 0.0006$ & $0.5136 \pm 0.0021$ & $\underline{0.5122} \pm 0.0009$ & $0.5151 \pm 0.0005$ \\
\end{longtable}
\endgroup

\begingroup
\small
\setlength{\tabcolsep}{5pt}
\renewcommand{\arraystretch}{1.05}
\begin{longtable}{@{}lrccccc@{}}
\caption{PatchTST: test MSE, mean $\pm$ sample SD over three seeds.}
\label{tab:pcr-full-patchtst-mse}\\
\toprule
Dataset & $H$ & Frozen base & PCR-$K4$ & PCR-$K2$ & Learned-$K4$ & Learned-$K2$ \\
\midrule
\endfirsthead
\multicolumn{7}{c}{\tablename\ \thetable\ continued: PatchTST, MSE}\\
\toprule
Dataset & $H$ & Frozen base & PCR-$K4$ & PCR-$K2$ & Learned-$K4$ & Learned-$K2$ \\
\midrule
\endhead
\midrule
\multicolumn{7}{r}{\footnotesize Continued on the next page}\\
\endfoot
\bottomrule
\endlastfoot
ETTh1 & 96 & $0.4050 \pm 0.0004$ & $\mathbf{0.3869} \pm 0.0021$ & $0.3897 \pm 0.0005$ & $0.3887 \pm 0.0013$ & $\underline{0.3874} \pm 0.0027$ \\
ETTh1 & 192 & $0.4507 \pm 0.0008$ & $0.4426 \pm 0.0020$ & $0.4418 \pm 0.0017$ & $\underline{0.4417} \pm 0.0012$ & $\mathbf{0.4413} \pm 0.0009$ \\
ETTh1 & 336 & $0.4860 \pm 0.0024$ & $0.4843 \pm 0.0042$ & $0.4838 \pm 0.0031$ & $\underline{0.4797} \pm 0.0055$ & $\mathbf{0.4773} \pm 0.0042$ \\
ETTh1 & 720 & $0.4883 \pm 0.0092$ & $0.4894 \pm 0.0103$ & $0.4886 \pm 0.0119$ & $\underline{0.4865} \pm 0.0114$ & $\mathbf{0.4844} \pm 0.0114$ \\
\addlinespace[3pt]
ETTh2 & 96 & $0.2931 \pm 0.0015$ & $0.2908 \pm 0.0021$ & $0.2903 \pm 0.0023$ & $\underline{0.2873} \pm 0.0019$ & $\mathbf{0.2871} \pm 0.0026$ \\
ETTh2 & 192 & $0.3751 \pm 0.0048$ & $\mathbf{0.3677} \pm 0.0032$ & $\underline{0.3695} \pm 0.0032$ & $0.3705 \pm 0.0059$ & $0.3698 \pm 0.0038$ \\
ETTh2 & 336 & $\underline{0.4176} \pm 0.0023$ & $\mathbf{0.4160} \pm 0.0026$ & $0.4184 \pm 0.0038$ & $0.4217 \pm 0.0039$ & $0.4191 \pm 0.0036$ \\
ETTh2 & 720 & $\underline{0.4207} \pm 0.0015$ & $0.4211 \pm 0.0023$ & $0.4221 \pm 0.0009$ & $0.4219 \pm 0.0018$ & $\mathbf{0.4203} \pm 0.0025$ \\
\addlinespace[3pt]
ETTm1 & 96 & $0.3414 \pm 0.0010$ & $0.3195 \pm 0.0007$ & $\underline{0.3193} \pm 0.0010$ & $\mathbf{0.3186} \pm 0.0014$ & $0.3211 \pm 0.0025$ \\
ETTm1 & 192 & $0.3739 \pm 0.0025$ & $0.3658 \pm 0.0055$ & $0.3665 \pm 0.0054$ & $\underline{0.3650} \pm 0.0041$ & $\mathbf{0.3646} \pm 0.0033$ \\
ETTm1 & 336 & $0.4030 \pm 0.0018$ & $0.3966 \pm 0.0015$ & $0.3970 \pm 0.0034$ & $\underline{0.3943} \pm 0.0011$ & $\mathbf{0.3942} \pm 0.0008$ \\
ETTm1 & 720 & $0.4582 \pm 0.0012$ & $0.4538 \pm 0.0010$ & $0.4536 \pm 0.0018$ & $\mathbf{0.4522} \pm 0.0012$ & $\underline{0.4527} \pm 0.0012$ \\
\addlinespace[3pt]
ETTm2 & 96 & $0.1778 \pm 0.0002$ & $\mathbf{0.1739} \pm 0.0001$ & $\underline{0.1741} \pm 0.0008$ & $\underline{0.1741} \pm 0.0005$ & $\mathbf{0.1739} \pm 0.0004$ \\
ETTm2 & 192 & $0.2434 \pm 0.0006$ & $\mathbf{0.2397} \pm 0.0005$ & $\underline{0.2401} \pm 0.0004$ & $0.2406 \pm 0.0005$ & $0.2407 \pm 0.0007$ \\
ETTm2 & 336 & $0.3041 \pm 0.0006$ & $\mathbf{0.3003} \pm 0.0008$ & $\underline{0.3006} \pm 0.0009$ & $0.3013 \pm 0.0005$ & $\mathbf{0.3003} \pm 0.0003$ \\
ETTm2 & 720 & $0.4010 \pm 0.0008$ & $\mathbf{0.3965} \pm 0.0014$ & $\underline{0.3972} \pm 0.0013$ & $0.3978 \pm 0.0009$ & $0.3973 \pm 0.0009$ \\
\addlinespace[3pt]
Weather & 96 & $0.1718 \pm 0.0002$ & $\mathbf{0.1636} \pm 0.0004$ & $\underline{0.1648} \pm 0.0008$ & $0.1652 \pm 0.0003$ & $0.1652 \pm 0.0004$ \\
Weather & 192 & $0.2194 \pm 0.0003$ & $\mathbf{0.2106} \pm 0.0008$ & $0.2118 \pm 0.0008$ & $\underline{0.2114} \pm 0.0015$ & $0.2119 \pm 0.0007$ \\
Weather & 336 & $0.2744 \pm 0.0008$ & $\mathbf{0.2686} \pm 0.0006$ & $\underline{0.2689} \pm 0.0003$ & $\mathbf{0.2686} \pm 0.0003$ & $0.2690 \pm 0.0013$ \\
Weather & 720 & $0.3513 \pm 0.0018$ & $\mathbf{0.3490} \pm 0.0007$ & $\underline{0.3492} \pm 0.0010$ & $0.3503 \pm 0.0003$ & $0.3494 \pm 0.0006$ \\
\addlinespace[3pt]
Electricity & 96 & $0.1972 \pm 0.0005$ & $\mathbf{0.1622} \pm 0.0007$ & $\mathbf{0.1622} \pm 0.0002$ & $\underline{0.1623} \pm 0.0003$ & $0.1631 \pm 0.0002$ \\
Electricity & 192 & $0.2061 \pm 0.0004$ & $\mathbf{0.1733} \pm 0.0003$ & $\underline{0.1738} \pm 0.0002$ & $0.1741 \pm 0.0005$ & $0.1748 \pm 0.0002$ \\
Electricity & 336 & $0.2226 \pm 0.0003$ & $\mathbf{0.1896} \pm 0.0001$ & $\underline{0.1904} \pm 0.0005$ & $0.1906 \pm 0.0002$ & $0.1913 \pm 0.0003$ \\
Electricity & 720 & $0.2653 \pm 0.0021$ & $\mathbf{0.2294} \pm 0.0009$ & $\underline{0.2301} \pm 0.0006$ & $0.2313 \pm 0.0009$ & $0.2312 \pm 0.0007$ \\
\addlinespace[3pt]
Traffic & 96 & $0.4949 \pm 0.0007$ & $0.4843 \pm 0.0018$ & $\underline{0.4795} \pm 0.0041$ & $0.4808 \pm 0.0014$ & $\mathbf{0.4748} \pm 0.0026$ \\
Traffic & 192 & $0.4979 \pm 0.0003$ & $0.4846 \pm 0.0015$ & $\underline{0.4799} \pm 0.0059$ & $0.4826 \pm 0.0019$ & $\mathbf{0.4751} \pm 0.0032$ \\
Traffic & 336 & $0.5142 \pm 0.0012$ & $0.4959 \pm 0.0014$ & $\mathbf{0.4914} \pm 0.0009$ & $0.4964 \pm 0.0019$ & $\underline{0.4929} \pm 0.0053$ \\
Traffic & 720 & $0.5510 \pm 0.0010$ & $0.5320 \pm 0.0019$ & $\underline{0.5285} \pm 0.0009$ & $0.5319 \pm 0.0007$ & $\mathbf{0.5282} \pm 0.0044$ \\
\addlinespace[3pt]
PEMS03 & 96 & $0.4856 \pm 0.0090$ & $\mathbf{0.3218} \pm 0.0213$ & $0.3335 \pm 0.0191$ & $\underline{0.3220} \pm 0.0193$ & $0.3389 \pm 0.0180$ \\
PEMS03 & 192 & $0.5350 \pm 0.0021$ & $\underline{0.3595} \pm 0.0073$ & $0.3691 \pm 0.0066$ & $\mathbf{0.3582} \pm 0.0077$ & $0.3728 \pm 0.0078$ \\
PEMS03 & 336 & $0.4314 \pm 0.0021$ & $\mathbf{0.3111} \pm 0.0046$ & $0.3190 \pm 0.0037$ & $\underline{0.3130} \pm 0.0044$ & $0.3234 \pm 0.0026$ \\
PEMS03 & 720 & $0.5135 \pm 0.0040$ & $\mathbf{0.3762} \pm 0.0017$ & $0.3845 \pm 0.0027$ & $\underline{0.3791} \pm 0.0032$ & $0.3861 \pm 0.0012$ \\
\addlinespace[3pt]
PEMS04 & 96 & $0.6279 \pm 0.0022$ & $\mathbf{0.3692} \pm 0.0026$ & $0.3928 \pm 0.0013$ & $\underline{0.3707} \pm 0.0024$ & $0.4062 \pm 0.0077$ \\
PEMS04 & 192 & $0.6634 \pm 0.0067$ & $\mathbf{0.4138} \pm 0.0024$ & $0.4341 \pm 0.0028$ & $\underline{0.4175} \pm 0.0037$ & $0.4467 \pm 0.0012$ \\
PEMS04 & 336 & $0.5198 \pm 0.0022$ & $\mathbf{0.3550} \pm 0.0023$ & $0.3703 \pm 0.0039$ & $\underline{0.3575} \pm 0.0030$ & $0.3790 \pm 0.0028$ \\
PEMS04 & 720 & $0.6132 \pm 0.0028$ & $\mathbf{0.4076} \pm 0.0021$ & $0.4246 \pm 0.0014$ & $\underline{0.4140} \pm 0.0026$ & $0.4394 \pm 0.0036$ \\
\addlinespace[3pt]
PEMS07 & 96 & $0.5044 \pm 0.0033$ & $\mathbf{0.3218} \pm 0.0006$ & $0.3391 \pm 0.0050$ & $\underline{0.3235} \pm 0.0066$ & $0.3427 \pm 0.0008$ \\
PEMS07 & 192 & $0.5549 \pm 0.0062$ & $\underline{0.3795} \pm 0.0068$ & $0.3931 \pm 0.0087$ & $\mathbf{0.3777} \pm 0.0040$ & $0.3923 \pm 0.0041$ \\
PEMS07 & 336 & $0.4412 \pm 0.0014$ & $\mathbf{0.3162} \pm 0.0008$ & $0.3277 \pm 0.0046$ & $\underline{0.3238} \pm 0.0056$ & $0.3290 \pm 0.0009$ \\
PEMS07 & 720 & $0.5129 \pm 0.0053$ & $\mathbf{0.3769} \pm 0.0021$ & $0.3899 \pm 0.0003$ & $\underline{0.3782} \pm 0.0014$ & $0.3920 \pm 0.0011$ \\
\addlinespace[3pt]
PEMS08 & 96 & $0.5598 \pm 0.0075$ & $\mathbf{0.3494} \pm 0.0020$ & $0.3684 \pm 0.0026$ & $\underline{0.3577} \pm 0.0046$ & $0.3627 \pm 0.0060$ \\
PEMS08 & 192 & $0.6164 \pm 0.0111$ & $\mathbf{0.4076} \pm 0.0074$ & $0.4222 \pm 0.0033$ & $\underline{0.4138} \pm 0.0114$ & $0.4199 \pm 0.0051$ \\
PEMS08 & 336 & $0.5271 \pm 0.0065$ & $\mathbf{0.3804} \pm 0.0060$ & $0.3898 \pm 0.0033$ & $0.3901 \pm 0.0064$ & $\underline{0.3857} \pm 0.0021$ \\
PEMS08 & 720 & $0.6180 \pm 0.0119$ & $\mathbf{0.4457} \pm 0.0086$ & $0.4565 \pm 0.0082$ & $0.4500 \pm 0.0118$ & $\underline{0.4491} \pm 0.0067$ \\
\end{longtable}
\endgroup

\begingroup
\small
\setlength{\tabcolsep}{5pt}
\renewcommand{\arraystretch}{1.05}
\begin{longtable}{@{}lrccccc@{}}
\caption{iTransformer: test MSE, mean $\pm$ sample SD over three seeds.}
\label{tab:pcr-full-itransformer-tslib-mse}\\
\toprule
Dataset & $H$ & Frozen base & PCR-$K4$ & PCR-$K2$ & Learned-$K4$ & Learned-$K2$ \\
\midrule
\endfirsthead
\multicolumn{7}{c}{\tablename\ \thetable\ continued: iTransformer, MSE}\\
\toprule
Dataset & $H$ & Frozen base & PCR-$K4$ & PCR-$K2$ & Learned-$K4$ & Learned-$K2$ \\
\midrule
\endhead
\midrule
\multicolumn{7}{r}{\footnotesize Continued on the next page}\\
\endfoot
\bottomrule
\endlastfoot
ETTh1 & 96 & $0.3898 \pm 0.0007$ & $\mathbf{0.3824} \pm 0.0014$ & $0.3838 \pm 0.0027$ & $\underline{0.3835} \pm 0.0003$ & $0.3857 \pm 0.0016$ \\
ETTh1 & 192 & $0.4445 \pm 0.0006$ & $\mathbf{0.4363} \pm 0.0010$ & $\underline{0.4379} \pm 0.0032$ & $0.4415 \pm 0.0008$ & $0.4394 \pm 0.0042$ \\
ETTh1 & 336 & $0.4856 \pm 0.0017$ & $\underline{0.4805} \pm 0.0039$ & $\mathbf{0.4789} \pm 0.0055$ & $0.4821 \pm 0.0025$ & $0.4811 \pm 0.0050$ \\
ETTh1 & 720 & $0.5137 \pm 0.0089$ & $\underline{0.5082} \pm 0.0055$ & $0.5083 \pm 0.0049$ & $\mathbf{0.5080} \pm 0.0037$ & $\underline{0.5082} \pm 0.0049$ \\
\addlinespace[3pt]
ETTh2 & 96 & $0.2988 \pm 0.0009$ & $\mathbf{0.2918} \pm 0.0014$ & $0.2930 \pm 0.0022$ & $\underline{0.2927} \pm 0.0006$ & $0.2931 \pm 0.0010$ \\
ETTh2 & 192 & $0.3800 \pm 0.0009$ & $\underline{0.3738} \pm 0.0009$ & $\mathbf{0.3735} \pm 0.0010$ & $0.3756 \pm 0.0015$ & $0.3745 \pm 0.0019$ \\
ETTh2 & 336 & $0.4236 \pm 0.0019$ & $\underline{0.4201} \pm 0.0032$ & $\mathbf{0.4198} \pm 0.0034$ & $0.4231 \pm 0.0010$ & $0.4263 \pm 0.0002$ \\
ETTh2 & 720 & $0.4317 \pm 0.0043$ & $0.4324 \pm 0.0062$ & $\underline{0.4304} \pm 0.0053$ & $0.4321 \pm 0.0065$ & $\mathbf{0.4301} \pm 0.0043$ \\
\addlinespace[3pt]
ETTm1 & 96 & $0.3430 \pm 0.0011$ & $\underline{0.3291} \pm 0.0014$ & $0.3292 \pm 0.0007$ & $\mathbf{0.3286} \pm 0.0012$ & $0.3305 \pm 0.0006$ \\
ETTm1 & 192 & $0.3826 \pm 0.0016$ & $\mathbf{0.3719} \pm 0.0012$ & $0.3722 \pm 0.0020$ & $0.3726 \pm 0.0009$ & $\underline{0.3720} \pm 0.0014$ \\
ETTm1 & 336 & $0.4184 \pm 0.0004$ & $\mathbf{0.4109} \pm 0.0022$ & $\underline{0.4110} \pm 0.0021$ & $0.4114 \pm 0.0013$ & $0.4123 \pm 0.0027$ \\
ETTm1 & 720 & $0.4922 \pm 0.0023$ & $\underline{0.4834} \pm 0.0005$ & $0.4838 \pm 0.0005$ & $\mathbf{0.4830} \pm 0.0019$ & $0.4836 \pm 0.0024$ \\
\addlinespace[3pt]
ETTm2 & 96 & $0.1841 \pm 0.0009$ & $\mathbf{0.1763} \pm 0.0012$ & $\underline{0.1768} \pm 0.0013$ & $0.1769 \pm 0.0007$ & $0.1771 \pm 0.0016$ \\
ETTm2 & 192 & $0.2517 \pm 0.0005$ & $\mathbf{0.2434} \pm 0.0009$ & $0.2440 \pm 0.0006$ & $\underline{0.2435} \pm 0.0010$ & $0.2447 \pm 0.0005$ \\
ETTm2 & 336 & $0.3151 \pm 0.0021$ & $\underline{0.3080} \pm 0.0028$ & $\mathbf{0.3079} \pm 0.0023$ & $0.3081 \pm 0.0034$ & $0.3086 \pm 0.0022$ \\
ETTm2 & 720 & $0.4120 \pm 0.0004$ & $\underline{0.4045} \pm 0.0006$ & $\mathbf{0.4043} \pm 0.0001$ & $0.4061 \pm 0.0019$ & $0.4069 \pm 0.0016$ \\
\addlinespace[3pt]
Weather & 96 & $0.1768 \pm 0.0008$ & $\underline{0.1638} \pm 0.0009$ & $0.1650 \pm 0.0010$ & $\mathbf{0.1635} \pm 0.0014$ & $0.1653 \pm 0.0021$ \\
Weather & 192 & $0.2252 \pm 0.0011$ & $\underline{0.2142} \pm 0.0009$ & $0.2166 \pm 0.0007$ & $0.2153 \pm 0.0016$ & $\mathbf{0.2140} \pm 0.0013$ \\
Weather & 336 & $0.2823 \pm 0.0003$ & $\underline{0.2730} \pm 0.0014$ & $0.2747 \pm 0.0013$ & $\mathbf{0.2719} \pm 0.0012$ & $0.2736 \pm 0.0025$ \\
Weather & 720 & $0.3599 \pm 0.0007$ & $\mathbf{0.3509} \pm 0.0009$ & $0.3529 \pm 0.0007$ & $0.3543 \pm 0.0010$ & $\underline{0.3527} \pm 0.0003$ \\
\addlinespace[3pt]
Electricity & 96 & $0.1507 \pm 0.0004$ & $\mathbf{0.1433} \pm 0.0001$ & $0.1442 \pm 0.0001$ & $\underline{0.1435} \pm 0.0001$ & $0.1450 \pm 0.0002$ \\
Electricity & 192 & $0.1656 \pm 0.0008$ & $\underline{0.1604} \pm 0.0008$ & $0.1610 \pm 0.0005$ & $\mathbf{0.1603} \pm 0.0005$ & $0.1614 \pm 0.0008$ \\
Electricity & 336 & $0.1802 \pm 0.0008$ & $\mathbf{0.1755} \pm 0.0013$ & $\underline{0.1761} \pm 0.0013$ & $0.1762 \pm 0.0011$ & $0.1762 \pm 0.0012$ \\
Electricity & 720 & $0.2140 \pm 0.0021$ & $\mathbf{0.2073} \pm 0.0010$ & $0.2087 \pm 0.0012$ & $\underline{0.2080} \pm 0.0011$ & $0.2081 \pm 0.0011$ \\
\addlinespace[3pt]
Traffic & 96 & $0.4080 \pm 0.0012$ & $\mathbf{0.4044} \pm 0.0011$ & $\mathbf{0.4044} \pm 0.0021$ & $\underline{0.4046} \pm 0.0010$ & $0.4053 \pm 0.0013$ \\
Traffic & 192 & $0.4283 \pm 0.0003$ & $\underline{0.4256} \pm 0.0007$ & $\mathbf{0.4245} \pm 0.0010$ & $0.4261 \pm 0.0002$ & $0.4268 \pm 0.0001$ \\
Traffic & 336 & $0.4432 \pm 0.0004$ & $\underline{0.4404} \pm 0.0010$ & $\mathbf{0.4396} \pm 0.0005$ & $0.4406 \pm 0.0005$ & $0.4416 \pm 0.0004$ \\
Traffic & 720 & $0.4768 \pm 0.0006$ & $0.4762 \pm 0.0015$ & $\mathbf{0.4744} \pm 0.0002$ & $\underline{0.4758} \pm 0.0003$ & $0.4772 \pm 0.0002$ \\
\addlinespace[3pt]
PEMS03 & 96 & $0.2624 \pm 0.0066$ & $\mathbf{0.2074} \pm 0.0052$ & $0.2162 \pm 0.0041$ & $\underline{0.2082} \pm 0.0048$ & $0.2201 \pm 0.0049$ \\
PEMS03 & 192 & $0.3086 \pm 0.0027$ & $\underline{0.2433} \pm 0.0025$ & $0.2537 \pm 0.0030$ & $\mathbf{0.2429} \pm 0.0029$ & $0.2567 \pm 0.0043$ \\
PEMS03 & 336 & $0.2661 \pm 0.0035$ & $\underline{0.2240} \pm 0.0035$ & $0.2301 \pm 0.0040$ & $\mathbf{0.2239} \pm 0.0036$ & $0.2330 \pm 0.0035$ \\
PEMS03 & 720 & $0.3376 \pm 0.0007$ & $\mathbf{0.2872} \pm 0.0019$ & $0.2937 \pm 0.0020$ & $\underline{0.2873} \pm 0.0023$ & $0.2970 \pm 0.0023$ \\
\addlinespace[3pt]
PEMS04 & 96 & $0.2978 \pm 0.0017$ & $\mathbf{0.1934} \pm 0.0009$ & $\underline{0.2158} \pm 0.0011$ & $\mathbf{0.1934} \pm 0.0011$ & $0.2237 \pm 0.0024$ \\
PEMS04 & 192 & $0.3258 \pm 0.0060$ & $\mathbf{0.2178} \pm 0.0049$ & $\underline{0.2419} \pm 0.0049$ & $\mathbf{0.2178} \pm 0.0045$ & $0.2491 \pm 0.0068$ \\
PEMS04 & 336 & $0.2870 \pm 0.0026$ & $\mathbf{0.2120} \pm 0.0018$ & $0.2275 \pm 0.0019$ & $\underline{0.2124} \pm 0.0021$ & $0.2329 \pm 0.0023$ \\
PEMS04 & 720 & $0.3438 \pm 0.0029$ & $\mathbf{0.2493} \pm 0.0008$ & $0.2680 \pm 0.0014$ & $\underline{0.2508} \pm 0.0012$ & $0.2730 \pm 0.0009$ \\
\addlinespace[3pt]
PEMS07 & 96 & $0.2335 \pm 0.0035$ & $\mathbf{0.1583} \pm 0.0023$ & $0.1723 \pm 0.0022$ & $\underline{0.1591} \pm 0.0017$ & $0.1796 \pm 0.0018$ \\
PEMS07 & 192 & $0.2750 \pm 0.0132$ & $\mathbf{0.1892} \pm 0.0071$ & $0.2078 \pm 0.0074$ & $\underline{0.1903} \pm 0.0070$ & $0.2109 \pm 0.0074$ \\
PEMS07 & 336 & $0.2325 \pm 0.0066$ & $\mathbf{0.1779} \pm 0.0055$ & $0.1899 \pm 0.0040$ & $\underline{0.1783} \pm 0.0049$ & $0.1917 \pm 0.0056$ \\
PEMS07 & 720 & $0.2868 \pm 0.0029$ & $\mathbf{0.2181} \pm 0.0029$ & $0.2305 \pm 0.0022$ & $\underline{0.2186} \pm 0.0032$ & $0.2351 \pm 0.0026$ \\
\addlinespace[3pt]
PEMS08 & 96 & $0.3166 \pm 0.0023$ & $\underline{0.2448} \pm 0.0022$ & $0.2540 \pm 0.0032$ & $\mathbf{0.2446} \pm 0.0021$ & $0.2586 \pm 0.0037$ \\
PEMS08 & 192 & $0.3774 \pm 0.0092$ & $\mathbf{0.2973} \pm 0.0100$ & $0.3067 \pm 0.0105$ & $\underline{0.2977} \pm 0.0092$ & $0.3098 \pm 0.0087$ \\
PEMS08 & 336 & $0.3376 \pm 0.0062$ & $\mathbf{0.2830} \pm 0.0053$ & $0.2884 \pm 0.0049$ & $\underline{0.2854} \pm 0.0071$ & $0.2903 \pm 0.0065$ \\
PEMS08 & 720 & $0.4063 \pm 0.0015$ & $\mathbf{0.3349} \pm 0.0013$ & $0.3433 \pm 0.0027$ & $\underline{0.3368} \pm 0.0020$ & $0.3466 \pm 0.0021$ \\
\end{longtable}
\endgroup

\begin{table}[!htb]
\centering\small
\setlength{\tabcolsep}{8pt}
\caption{Dataset-balanced MAE reductions (\%) from the frozen base. Brackets give 95\% dataset-bootstrap intervals; higher is better.}
\label{tab:pcr-mae-summary}
\begin{tabular}{@{}lcccc@{}}
\toprule
Backbone & PCR-$K4$ & PCR-$K2$ & Learned-$K4$ & Learned-$K2$ \\
\midrule
DLinear & \shortstack{$1.87$\\$[0.96, 2.86]$} & \shortstack{$1.76$\\$[0.83, 2.76]$} & \shortstack{$1.87$\\$[1.13, 2.69]$} & \shortstack{$1.85$\\$[1.08, 2.75]$} \\
PatchTST & \shortstack{$11.17$\\$[5.41, 17.01]$} & \shortstack{$10.44$\\$[5.07, 15.81]$} & \shortstack{$10.99$\\$[5.39, 16.78]$} & \shortstack{$10.24$\\$[5.03, 15.48]$} \\
iTransformer & \shortstack{$5.87$\\$[2.59, 9.50]$} & \shortstack{$4.75$\\$[2.19, 7.57]$} & \shortstack{$5.71$\\$[2.39, 9.37]$} & \shortstack{$4.29$\\$[2.06, 6.79]$} \\
\bottomrule
\end{tabular}
\end{table}

\subsection{Comparison with Existing Methods}
\label{app:pcr-competitors}

We compare PCR with MoLE~\citep{ni2024mole}, CCM~\citep{chen2024ccm},
and VE~\citep{wang2024ve} at $L=96$, four horizons, eleven datasets,
and seeds $0,1,2$.
MoLE is evaluated on DLinear; CCM and VE are evaluated on all three
backbones, with our CCM adaptation to iTransformer described in
Appendix~\ref{app:pcr-ccm-itransformer}.
MoLE, CCM, PCR, and matched VE use four experts; original VE follows
the documented configuration for each dataset and backbone.
PCR and learned-allocation results are those in
Appendix~\ref{app:pcr-full-grid-tables}.
These expert-based competitors train jointly, whereas PCR freezes its
backbone, so the comparisons include modeling and optimization
differences beyond routing. A separate $\delta$-Adapter comparison
keeps the same pretrained backbone frozen (Appendix~\ref{app:pcr-delta}).

\paragraph{Implementations and capacity controls.}
MoLE original uses a first-step calendar router and head dropout.
CCM uses soft cluster assignments, prototypes, and a clustering loss;
our multivariate implementation updates prototypes by the batch mean
and uses a $C\times C$ channel-similarity matrix.
CCM-PatchTST shares our encoder width but retains CCM's affine
RevIN and cluster head.

Original DLinear experts have separate seasonal/trend maps and biases.
Matched experts concatenate these features and use $U_eV_e$, with
$U_e\in\mathbb R^{2L\times16}$ and $V_e\in\mathbb R^{16\times H}$,
without bias. Matched PatchTST heads use flattened prediction-head
features. These banks have $4\cdot16(F+H)$ parameters, matching PCR's
expert budget, not total capacity: backbones, routers, assigners, and
prototypes remain additional. MoLE's expert-budget control retains
calendar inputs; MoLE matched removes them
(Appendix~\ref{app:pcr-calendar}).

\paragraph{Selection and calendar features.}
Hyperparameters are selected on seed-zero validation MSE and fixed
for seeds 1 and 2. MoLE original and the expert-budget control search
learning rates $\{0.005,0.01,0.05\}$ and head dropout $\{0,0.2\}$,
with batch 8, a 40-epoch cap, and patience 6.
CCM uses batch 32, a 200-epoch cap, and patience 10; learning-rate
grids are $\{10^{-4},10^{-3},10^{-2}\}$ for DLinear and
$\{10^{-4},5\times10^{-4},10^{-3}\}$ for PatchTST and iTransformer.
Calendar-aware MoLE uses a constant date offset for Electricity/Traffic,
preserving periodicity with shifted phase. PEMS timestamps are reconstructed
from documented starts; PEMS07 ends on 6 August 2017.

\paragraph{Coverage and comparison.}
Every original/matched method covers 44 settings per evaluated backbone
and three seeds. Competitor-relative reductions use each setting's
three-seed mean error, with the competitor in the denominator, then
average horizons within datasets and datasets equally.
This differs from averaging seed-relative backbone gains in
Table~\ref{tab:pcr-competitors}.
PCR-K4 reduces MSE relative to original/matched CCM by
1.37\%/2.69\% on DLinear and 7.19\%/9.02\% on PatchTST,
winning 35/44, 36/44, 36/44, and 41/44 settings, respectively.
PCR-K2 gives reductions of 1.15\%/2.47\% and 6.07\%/7.90\%,
with 34/44, 36/44, 36/44, and 41/44 wins.
PCR-K4 reduces MSE by 1.96\% against MoLE matched, winning 40/44 settings.
MoLE original and the expert-budget control achieve lower average DLinear
MSE; PCR-K4 wins 18/44 and 19/44 settings against them, respectively.
Table~\ref{tab:pcr-ccm-dlinear-full} retains all competitor and
calendar-control results.

\subsubsection{CCM on iTransformer}
\label{app:pcr-ccm-itransformer}
\par\begingroup
CCM's official release does not include iTransformer. We therefore adapt
its cluster assigner, prototypes, clustering loss, and cluster-specific
heads to the benchmark's iTransformer. The encoder architecture,
calendar tokens, and per-window normalization are unchanged; no additional
RevIN is introduced. The assigner receives only the observed channels'
train-standardized input windows. Calendar tokens remain inside the encoder
and do not enter clustering or the prediction heads.

Each observed-channel token is passed through four heads, and their outputs
are mixed using that channel's soft cluster probabilities. The mixture is
then denormalized once. The original variant uses four full linear heads
with biases; the matched variant uses four bias-free rank-16 heads with
$4\cdot16(d_{\mathrm{model}}+H)$ parameters, exactly matching PCR's expert
bank. The encoder and assigner remain additional parameters.
We retain the validated multivariate CCM implementation's batch-mean
prototype updates and $C\times C$ channel-similarity matrix. Prototypes
update during training only. Cluster probabilities average similarity
across the current batch: evaluation uses chronological batches of 32,
whereas single-window deployment assigns clusters from that window alone.

The encoder, assigner, and heads are trained jointly from fresh
initialization using Adam, batch size 32, a 200-epoch cap, and patience 10.
The learning rate is halved after epochs 2, 4, 6, 8, and 10, then held fixed.
For each dataset, horizon, and variant, seed-zero validation MSE selects
from $\{10^{-4},5\times10^{-4},10^{-3}\}$; the chosen configuration is
then run with seeds 1 and 2. The selected seed-zero trial supplies the
third reported result. Checkpoints are selected by validation MSE, and
test evaluation follows selection. All 264 selected fits are complete:
11 datasets, $L=96$, $H\in\{96,192,336,720\}$, three seeds, and two variants.
Data splits, scaling, windows, and metrics match the PCR benchmark.

This adaptation compares complete methods: CCM jointly trains its backbone
under its own recipe, while PCR corrects a frozen pretrained backbone.
It therefore does not isolate routing, and expert-budget matching does not
equalize total parameters.
\par\endgroup

\par\begingroup
\paragraph{Accuracy and exceptions.}
Using the seed-relative convention of Table~\ref{tab:pcr-competitors},
original CCM improves backbone MSE/MAE by $4.70\%/3.36\%$;
the matched variant gives $-1.70\%/-1.91\%$.
For the competitor-relative convention above, PCR-K4 reduces MSE by
$5.66\%$ against original CCM and $10.97\%$ against matched CCM,
winning 37/44 and 42/44 settings; the corresponding MAE reductions
are $2.49\%$ and $7.26\%$.
PCR-K2 gives MSE reductions of $3.77\%$ and $9.24\%$, with
35/44 and 42/44 wins. Original CCM is better than PCR-K4 on Traffic
at every horizon and on PEMS08 at $H=192,336,720$; matched CCM is
better on ETTh2 at $H=720$ and Traffic at $H=96$.
All per-setting errors are included in Table~\ref{tab:pcr-ccm-itransformer-full}.
One original-CCM fit reaches the 200-epoch cap. The large PEMS04 margins
partly reflect a validation-to-test gap under CCM's registered recipe;
alternative training schedules were not evaluated.
\par\endgroup

\subsubsection{Separating Expert Budget and Calendar Access}
\label{app:pcr-calendar}

MoLE matched replaces the expert-budget control's calendar input by
zeros during training, validation, and test. Its four rank-16 experts
contain $4\cdot16(2L+H)$ parameters, matching PCR-DLinear, and neither
receives calendar covariates. Architecture, expert tensors, head dropout,
loss, optimizer, and selected hyperparameters are unchanged.
Matching seeds reproduce initial weights and window order; the ablation
selects its own checkpoint and stopping epoch on validation MSE.

The three MoLE variants answer different questions. Original versus
expert-budget control changes expert capacity and parameterization;
expert-budget control versus matched removes calendar conditioning
at fixed capacity and recipe. With zero input, the router learns a
persistent channel-specific mixture. Nominal parameters are unchanged,
but input-layer weights receive no data gradient and inactive hidden
units remain inactive. This is not an independently retuned model or
a full factorial experiment, and freezing and residual prediction
still differ from PCR.

For three-seed mean errors $m_{\mathrm{orig}}$, $m_{\mathrm{control}}$,
and $m_{\mathrm{matched}}$, Table~\ref{tab:pcr-calendar-contrasts}
reports the original-expert and calendar benefits:
\[
100\left(1-\frac{m_{\mathrm{orig}}}{m_{\mathrm{control}}}\right),
\qquad
100\left(1-\frac{m_{\mathrm{control}}}{m_{\mathrm{matched}}}\right).
\]
Positive values favor original experts or calendar access; aggregation
weights horizons within datasets and then datasets equally.
Calendar access reduces mean MSE by 8.20\% under this recipe, most
strongly on PEMS03/04/08 and Traffic. All 25 settings favoring the
expert-budget control over PCR-K4 reverse when calendar inputs are removed.
PCR-K2 improves on MoLE matched by 1.72\%, with 31/44 wins;
PCR-K4's comparison is given above. Calendar information therefore
contributes to MoLE's advantage under the fixed recipe, while MoLE
original remains stronger on DLinear on average.

\begin{table}[!htbp]
\centering\small
\setlength{\tabcolsep}{5pt}
\caption{MoLE capacity and calendar contrasts (\%). Each dataset averages all four horizons and each error uses three seeds. Positive values favor original experts or calendar access. The final row weights datasets equally; these contrasts are not method rankings.}
\label{tab:pcr-calendar-contrasts}
\begin{tabular}{lrrrr}
\toprule
& \multicolumn{2}{c}{Original-expert benefit} & \multicolumn{2}{c}{Calendar benefit} \\
\cmidrule(lr){2-3}\cmidrule(lr){4-5}
Dataset & MSE & MAE & MSE & MAE \\
\midrule
ETTh1 & $+0.70$ & $+1.08$ & $-2.02$ & $-1.86$ \\
ETTh2 & $-42.63$ & $-15.33$ & $-2.19$ & $-0.93$ \\
ETTm1 & $-1.59$ & $-0.96$ & $-5.93$ & $-4.99$ \\
ETTm2 & $-4.62$ & $-1.85$ & $-1.33$ & $+0.06$ \\
Electricity & $+5.81$ & $+3.25$ & $+2.46$ & $+0.11$ \\
Traffic & $+5.62$ & $+6.35$ & $+15.43$ & $+8.00$ \\
Weather & $+0.23$ & $-0.62$ & $-2.66$ & $-1.45$ \\
PEMS03 & $+26.11$ & $+13.45$ & $+25.05$ & $+19.80$ \\
PEMS04 & $+22.59$ & $+10.42$ & $+31.27$ & $+20.99$ \\
PEMS07 & $+5.94$ & $+3.56$ & $+3.41$ & $+13.22$ \\
PEMS08 & $+19.55$ & $+10.55$ & $+26.73$ & $+20.78$ \\
\midrule
Mean & $+3.43$ & $+2.72$ & $+8.20$ & $+6.70$ \\
\bottomrule
\end{tabular}
\end{table}
\subsubsection{Full Competitor Results}
\begingroup\scriptsize\setlength{\tabcolsep}{2.5pt}
\begin{longtable}{llrccccc}
\caption{Competitor and calendar-ablation results for all 44 dataset-horizon settings per backbone. Entries stack MSE above MAE, with mean $\pm$ sample SD over three seeds. The corresponding frozen-base, PCR, and learned-mixture results appear in Appendix~\ref{app:pcr-full-grid-tables}. iTransformer evaluates our CCM adaptation; dashes indicate methods not evaluated on that backbone.}
\label{tab:pcr-ccm-dlinear-full}\label{tab:pcr-ccm-patchtst-full}\label{tab:pcr-calendar-full}\label{tab:pcr-ccm-itransformer-full}\\
\toprule
Backbone & Dataset & $H$ & \shortstack{CCM\\original} & \shortstack{CCM\\matched} & \shortstack{MoLE\\original} & \shortstack{MoLE expert-\\budget control} & \shortstack{MoLE\\matched} \\
\midrule
\endfirsthead
\multicolumn{8}{c}{\tablename\ \thetable{} - continued}\\
\toprule
Backbone & Dataset & $H$ & \shortstack{CCM\\original} & \shortstack{CCM\\matched} & \shortstack{MoLE\\original} & \shortstack{MoLE expert-\\budget control} & \shortstack{MoLE\\matched} \\
\midrule
\endhead
\midrule\multicolumn{8}{r}{Continued on next page}\\
\endfoot
\bottomrule
\endlastfoot
DLinear & ETTh1 & 96 & \shortstack{$0.385\!\pm\!0.002$\\$0.396\!\pm\!0.001$} & \shortstack{$0.383\!\pm\!0.001$\\$0.399\!\pm\!0.001$} & \shortstack{$0.391\!\pm\!0.003$\\$0.401\!\pm\!0.005$} & \shortstack{$0.400\!\pm\!0.009$\\$0.411\!\pm\!0.012$} & \shortstack{$0.381\!\pm\!0.003$\\$0.395\!\pm\!0.004$} \\
DLinear & ETTh1 & 192 & \shortstack{$0.436\!\pm\!0.003$\\$0.431\!\pm\!0.002$} & \shortstack{$0.438\!\pm\!0.003$\\$0.436\!\pm\!0.004$} & \shortstack{$0.438\!\pm\!0.003$\\$0.429\!\pm\!0.006$} & \shortstack{$0.441\!\pm\!0.007$\\$0.436\!\pm\!0.010$} & \shortstack{$0.436\!\pm\!0.003$\\$0.433\!\pm\!0.003$} \\
DLinear & ETTh1 & 336 & \shortstack{$0.483\!\pm\!0.002$\\$0.459\!\pm\!0.004$} & \shortstack{$0.500\!\pm\!0.011$\\$0.478\!\pm\!0.012$} & \shortstack{$0.489\!\pm\!0.016$\\$0.459\!\pm\!0.015$} & \shortstack{$0.488\!\pm\!0.008$\\$0.457\!\pm\!0.009$} & \shortstack{$0.477\!\pm\!0.003$\\$0.446\!\pm\!0.001$} \\
DLinear & ETTh1 & 720 & \shortstack{$0.524\!\pm\!0.010$\\$0.515\!\pm\!0.008$} & \shortstack{$0.515\!\pm\!0.011$\\$0.509\!\pm\!0.006$} & \shortstack{$0.507\!\pm\!0.005$\\$0.495\!\pm\!0.005$} & \shortstack{$0.508\!\pm\!0.024$\\$0.498\!\pm\!0.021$} & \shortstack{$0.509\!\pm\!0.024$\\$0.497\!\pm\!0.023$} \\
\addlinespace[2pt]
DLinear & ETTh2 & 96 & \shortstack{$0.301\!\pm\!0.009$\\$0.357\!\pm\!0.008$} & \shortstack{$0.311\!\pm\!0.012$\\$0.367\!\pm\!0.010$} & \shortstack{$0.433\!\pm\!0.052$\\$0.426\!\pm\!0.024$} & \shortstack{$0.341\!\pm\!0.005$\\$0.391\!\pm\!0.004$} & \shortstack{$0.328\!\pm\!0.012$\\$0.383\!\pm\!0.010$} \\
DLinear & ETTh2 & 192 & \shortstack{$0.397\!\pm\!0.019$\\$0.418\!\pm\!0.015$} & \shortstack{$0.423\!\pm\!0.019$\\$0.439\!\pm\!0.017$} & \shortstack{$0.492\!\pm\!0.049$\\$0.464\!\pm\!0.020$} & \shortstack{$0.430\!\pm\!0.018$\\$0.443\!\pm\!0.011$} & \shortstack{$0.462\!\pm\!0.051$\\$0.456\!\pm\!0.026$} \\
\end{longtable}
\endgroup

\subsubsection{Variable Embedding}
\label{app:pcr-ve}

VE~\citep{wang2024ve} replaces the prediction head with low-rank
experts mixed by persistent channel embeddings and jointly trains the
complete model from initialization. DLinear uses seasonal and trend
banks sharing one embedding; PatchTST replaces its flattened-feature
projection; iTransformer projects the observed-channel tokens while
retaining calendar tokens in its encoder. Benchmark backbone
normalization, dropout, splits, and input information are retained.
Each expert includes a bias row. For per-bank input width $d$, original
VE uses rank $\lfloor p(d+1)H/[k(d+1+H)]\rfloor$, with $(k,p)$ in
Table~\ref{tab:pcr-ve-recipes}. Documented settings are used where
available; remaining settings inherit a documented recipe with the
same benchmark architecture.

\begin{table}[!htbp]
\centering\small
\setlength{\tabcolsep}{5pt}
\caption{Original VE capacity recipes. Entries are (number of experts $k$, capacity ratio $p$); ``remaining'' covers the other datasets within each backbone. Here $d=L$ for each DLinear bank and $d=F$ for the transformers.}
\label{tab:pcr-ve-recipes}
\begin{tabular}{llcc}
\toprule
Backbone & Datasets & $H=96$ & $H\in\{192,336,720\}$ \\
\midrule
DLinear & ETTh1, ETTm1 & $(16,1)$ & $(2,1)$ \\
 & Weather & $(8,4)$ & $(8,4)$ \\
 & Remaining & $(2,4)$ & $(2,4)$ \\
\addlinespace
PatchTST & ETTh1 & $(4,4)$ & $(16,1)$ \\
 & ETTh2 & $(16,1)$ & $(128,4)$ \\
 & Remaining & $(4,1)$ & $(2,1)$ \\
\addlinespace
iTransformer & ETTh2, ETTm1, ETTm2 & $(32,4)$ & $(32,4)$ \\
 & Remaining & $(8,4)$ & $(8,4)$ \\
\bottomrule
\end{tabular}
\end{table}

Matched VE uses four experts, choosing the integer rank nearest
PCR-K4's expert-factor budget $P=4\cdot16(F+H)$, including VE's bias
parameters. All 132 settings are within $5\%$ of $P$. This matches
the expert head; VE additionally gradient-trains its backbone and
channel embeddings.

Both VE variants search learning rates
$\{10^{-4},5\times10^{-4},10^{-3},5\times10^{-3}\}$ and batch sizes
$\{16,32\}$ with Adam. The eight candidates are selected by seed-zero
validation MSE; seeds one and two independently fit the selected
configuration. Training lasts at most ten epochs with patience three;
the initial learning rate is used for two epochs and then halved each
epoch. The lowest-validation-MSE checkpoint is tested once.
All three seeds are reported.

PCR-K4 has lower average MSE and MAE than both VE variants on every
backbone. Its MSE wins span 31--43 of 44 settings
(Table~\ref{tab:pcr-ve-delta-summary}); these are comparisons of complete
methods, rather than isolated tests of the allocation solver.

\subsubsection{Broader Frozen-Forecaster Adaptation}
\label{app:pcr-delta}

We evaluate additive Ada-X+Y from
$\delta$-Adapter~\citep{liang2026deltaadapter} on the same frozen
checkpoint as PCR for each setting and seed. Two three-layer MLPs
apply bounded additive corrections to the flattened input and forecast,
with input/output dimensions $LC$ and $HC$, respectively, and ReLU and
InstanceNorm between linear layers. Both adapters are trained jointly;
the backbone stays fixed and no test-time updates are performed.
Original adapters use hidden width 512. The reduced-width variant
chooses the shared width $w\geq3$ nearest $P$; the lower bound avoids
degenerate narrow InstanceNorm representations.

Across 132 settings, original adapters use 18--12,662 times $P$
(median 752); reduced-width adapters use 0.918--86.4 times $P$
(median 5.1), with only 28 settings within $5\%$. We therefore report
the latter as reduced-width, rather than uniformly budget-matched.
Trainable counts cover both adapters; deployed counts additionally
include the frozen backbone.

Each variant searches learning rates $\{10^{-5},10^{-4},10^{-3},10^{-2}\}$
and correction scales $\{0.01,0.1\}$ at batch size 16, using one Adam
optimizer per adapter. Selection, seeds, epoch limit, stopping,
and learning-rate schedule follow the VE procedure above.

Original $\delta$-Adapter achieves higher average MSE accuracy on all
three backbones with substantially greater capacity and higher median
inference latency (Appendix~\ref{app:pcr-cost}). Against reduced-width
adapters, PCR-K4 has lower average MSE and MAE on both transformers;
DLinear favors the adapter on average, while PCR wins 24 of 44 MSE
settings. Table~\ref{tab:pcr-ve-delta-summary} reports the complete
aggregate comparisons and uncertainty.

\begin{table}[!htbp]
\centering\small
\setlength{\tabcolsep}{3pt}
\caption{VE and $\delta$-Adapter across 44 settings per backbone. Backbone-relative gains (\%) are means $\pm$ sample SD over three seed aggregates, averaging horizons within datasets and weighting datasets equally. $R=100(m_{\mathrm{method}}-m_{\mathrm{PCR}})/m_{\mathrm{method}}$ uses each setting's three-seed mean error before averaging; positive $R$ favors PCR-K4. Parentheses count PCR-K4 wins out of 44. Comparisons are descriptive.}
\label{tab:pcr-ve-delta-summary}
\begin{tabular}{llrrrr}
\toprule
Backbone & Method & \shortstack{MSE gain\\(\%)} & \shortstack{MAE gain\\(\%)} & \shortstack{MSE $R$\\(wins)} & \shortstack{MAE $R$\\(wins)} \\
\midrule
DLinear & PCR-K4 & $3.86\pm0.21$ & $1.87\pm0.21$ & --- & --- \\
 & VE original & $-2.49\pm5.47$ & $-1.11\pm2.38$ & $+5.12\ (43)$ & $+2.67\ (40)$ \\
 & VE matched & $-1.65\pm1.97$ & $-1.46\pm1.08$ & $+4.33\ (39)$ & $+2.87\ (38)$ \\
 & $\delta$-Adapter original & $11.01\pm0.10$ & $7.85\pm0.09$ & $-9.97\ (15)$ & $-7.57\ (16)$ \\
 & $\delta$-Adapter reduced-width & $4.40\pm0.56$ & $2.84\pm0.47$ & $-0.92\ (24)$ & $-1.22\ (22)$ \\
\addlinespace
PatchTST & PCR-K4 & $14.25\pm0.10$ & $11.17\pm0.09$ & --- & --- \\
 & VE original & $8.97\pm2.46$ & $6.75\pm1.94$ & $+5.63\ (36)$ & $+4.62\ (38)$ \\
 & VE matched & $11.29\pm0.40$ & $7.14\pm0.52$ & $+3.56\ (36)$ & $+4.50\ (42)$ \\
 & $\delta$-Adapter original & $18.93\pm0.08$ & $12.88\pm0.06$ & $-8.26\ (13)$ & $-2.96\ (22)$ \\
 & $\delta$-Adapter reduced-width & $12.89\pm0.22$ & $8.03\pm0.15$ & $+1.58\ (33)$ & $+3.44\ (44)$ \\
\addlinespace
iTransformer & PCR-K4 & $10.22\pm0.06$ & $5.87\pm0.08$ & --- & --- \\
 & VE original & $7.32\pm1.35$ & $4.10\pm0.88$ & $+3.18\ (31)$ & $+1.83\ (30)$ \\
 & VE matched & $3.57\pm4.21$ & $0.93\pm2.16$ & $+6.37\ (40)$ & $+4.80\ (42)$ \\
 & $\delta$-Adapter original & $11.53\pm0.04$ & $6.31\pm0.04$ & $-2.06\ (17)$ & $-0.84\ (23)$ \\
 & $\delta$-Adapter reduced-width & $6.01\pm0.21$ & $2.45\pm0.18$ & $+4.78\ (42)$ & $+3.55\ (44)$ \\
\bottomrule
\end{tabular}
\end{table}


\begingroup
\small
\setlength{\tabcolsep}{6pt}
\begin{longtable}{@{}llrrrr@{}}
\caption{VE on DLinear: test MSE and MAE, mean $\pm$ sample SD over three seeds, with $L=96$.}\label{tab:ve-dlinear-full}\\
\toprule
 & & \multicolumn{2}{c}{Original} & \multicolumn{2}{c}{Expert-budget matched}\\
\cmidrule(lr){3-4}\cmidrule(lr){5-6}
Dataset & $H$ & MSE & MAE & MSE & MAE\\
\midrule
\endfirsthead
\multicolumn{6}{c}{\tablename\ \thetable\ continued}\\
\toprule
 & & \multicolumn{2}{c}{Original} & \multicolumn{2}{c}{Expert-budget matched}\\
\cmidrule(lr){3-4}\cmidrule(lr){5-6}
Dataset & $H$ & MSE & MAE & MSE & MAE\\
\midrule
\endhead
\midrule
\multicolumn{6}{r}{Continued on the next page}\\
\endfoot
\bottomrule
\endlastfoot
ETTh1 & 96 & $0.385\pm0.003$ & $0.397\pm0.003$ & $0.384\pm0.005$ & $0.396\pm0.005$\\
 & 192 & $0.442\pm0.004$ & $0.431\pm0.004$ & $0.440\pm0.004$ & $0.429\pm0.004$\\
 & 336 & $0.482\pm0.001$ & $0.449\pm0.001$ & $0.482\pm0.000$ & $0.451\pm0.002$\\
 & 720 & $0.520\pm0.009$ & $0.502\pm0.008$ & $0.512\pm0.008$ & $0.499\pm0.006$\\
\addlinespace[3pt]
ETTh2 & 96 & $0.384\pm0.099$ & $0.406\pm0.058$ & $0.404\pm0.062$ & $0.426\pm0.032$\\
 & 192 & $0.540\pm0.115$ & $0.492\pm0.047$ & $0.525\pm0.011$ & $0.494\pm0.005$\\
 & 336 & $0.724\pm0.253$ & $0.571\pm0.091$ & $0.652\pm0.078$ & $0.563\pm0.040$\\
 & 720 & $0.934\pm0.199$ & $0.651\pm0.055$ & $0.855\pm0.050$ & $0.650\pm0.011$\\
\end{longtable}
\endgroup

\begingroup
\small
\setlength{\tabcolsep}{6pt}
\begin{longtable}{@{}llrrrr@{}}
\caption{$\delta$-Adapter on DLinear: test MSE and MAE, mean $\pm$ sample SD over three seeds, with $L=96$. The reduced-width variant uses the closest feasible adapter budget.}\label{tab:delta-dlinear-full}\\
\toprule
 & & \multicolumn{2}{c}{Original} & \multicolumn{2}{c}{Reduced-width}\\
\cmidrule(lr){3-4}\cmidrule(lr){5-6}
Dataset & $H$ & MSE & MAE & MSE & MAE\\
\midrule
\endfirsthead
\multicolumn{6}{c}{\tablename\ \thetable\ continued}\\
\toprule
 & & \multicolumn{2}{c}{Original} & \multicolumn{2}{c}{Reduced-width}\\
\cmidrule(lr){3-4}\cmidrule(lr){5-6}
Dataset & $H$ & MSE & MAE & MSE & MAE\\
\midrule
\endhead
\midrule
\multicolumn{6}{r}{Continued on the next page}\\
\endfoot
\bottomrule
\endlastfoot
ETTh1 & 96 & $0.388\pm0.003$ & $0.402\pm0.004$ & $0.386\pm0.003$ & $0.400\pm0.004$\\
 & 192 & $0.453\pm0.029$ & $0.445\pm0.028$ & $0.452\pm0.030$ & $0.444\pm0.028$\\
 & 336 & $0.525\pm0.023$ & $0.496\pm0.021$ & $0.533\pm0.020$ & $0.502\pm0.016$\\
 & 720 & $0.586\pm0.022$ & $0.563\pm0.014$ & $0.555\pm0.039$ & $0.539\pm0.027$\\
\addlinespace[3pt]
ETTh2 & 96 & $0.323\pm0.042$ & $0.376\pm0.037$ & $0.321\pm0.040$ & $0.374\pm0.036$\\
 & 192 & $0.443\pm0.007$ & $0.452\pm0.007$ & $0.437\pm0.007$ & $0.449\pm0.007$\\
 & 336 & $0.511\pm0.017$ & $0.496\pm0.011$ & $0.504\pm0.015$ & $0.492\pm0.010$\\
 & 720 & $0.788\pm0.051$ & $0.659\pm0.022$ & $0.805\pm0.052$ & $0.666\pm0.021$\\
\end{longtable}
\endgroup

\begingroup
\small
\setlength{\tabcolsep}{6pt}
\begin{longtable}{@{}llrrrr@{}}
\caption{VE on PatchTST: test MSE and MAE, mean $\pm$ sample SD over three seeds, with $L=96$.}\label{tab:ve-patchtst-full}\\
\toprule
 & & \multicolumn{2}{c}{Original} & \multicolumn{2}{c}{Expert-budget matched}\\
\cmidrule(lr){3-4}\cmidrule(lr){5-6}
Dataset & $H$ & MSE & MAE & MSE & MAE\\
\midrule
\endfirsthead
\multicolumn{6}{c}{\tablename\ \thetable\ continued}\\
\toprule
 & & \multicolumn{2}{c}{Original} & \multicolumn{2}{c}{Expert-budget matched}\\
\cmidrule(lr){3-4}\cmidrule(lr){5-6}
Dataset & $H$ & MSE & MAE & MSE & MAE\\
\midrule
\endhead
\midrule
\multicolumn{6}{r}{Continued on the next page}\\
\endfoot
\bottomrule
\endlastfoot
ETTh1 & 96 & $0.397\pm0.002$ & $0.411\pm0.001$ & $0.395\pm0.002$ & $0.412\pm0.003$\\
 & 192 & $0.442\pm0.004$ & $0.439\pm0.002$ & $0.445\pm0.003$ & $0.441\pm0.000$\\
 & 336 & $0.485\pm0.004$ & $0.464\pm0.002$ & $0.488\pm0.004$ & $0.466\pm0.001$\\
 & 720 & $0.478\pm0.014$ & $0.481\pm0.011$ & $0.475\pm0.004$ & $0.481\pm0.002$\\
\addlinespace[3pt]
ETTh2 & 96 & $0.296\pm0.006$ & $0.345\pm0.004$ & $0.294\pm0.003$ & $0.344\pm0.002$\\
 & 192 & $0.377\pm0.001$ & $0.393\pm0.000$ & $0.376\pm0.002$ & $0.394\pm0.001$\\
 & 336 & $0.415\pm0.001$ & $0.426\pm0.001$ & $0.424\pm0.005$ & $0.435\pm0.003$\\
 & 720 & $0.427\pm0.001$ & $0.443\pm0.000$ & $0.441\pm0.008$ & $0.449\pm0.004$\\
\end{longtable}
\endgroup

\begingroup
\small
\setlength{\tabcolsep}{6pt}
\begin{longtable}{@{}llrrrr@{}}
\caption{$\delta$-Adapter on PatchTST: test MSE and MAE, mean $\pm$ sample SD over three seeds, with $L=96$. The reduced-width variant uses the closest feasible adapter budget.}\label{tab:delta-patchtst-full}\\
\toprule
 & & \multicolumn{2}{c}{Original} & \multicolumn{2}{c}{Reduced-width}\\
\cmidrule(lr){3-4}\cmidrule(lr){5-6}
Dataset & $H$ & MSE & MAE & MSE & MAE\\
\midrule
\endfirsthead
\multicolumn{6}{c}{\tablename\ \thetable\ continued}\\
\toprule
 & & \multicolumn{2}{c}{Original} & \multicolumn{2}{c}{Reduced-width}\\
\cmidrule(lr){3-4}\cmidrule(lr){5-6}
Dataset & $H$ & MSE & MAE & MSE & MAE\\
\midrule
\endhead
\midrule
\multicolumn{6}{r}{Continued on the next page}\\
\endfoot
\bottomrule
\endlastfoot
ETTh1 & 96 & $0.395\pm0.001$ & $0.411\pm0.001$ & $0.403\pm0.002$ & $0.416\pm0.003$\\
 & 192 & $0.439\pm0.002$ & $0.437\pm0.001$ & $0.451\pm0.002$ & $0.444\pm0.003$\\
 & 336 & $0.480\pm0.004$ & $0.462\pm0.003$ & $0.488\pm0.003$ & $0.467\pm0.001$\\
 & 720 & $0.488\pm0.015$ & $0.492\pm0.012$ & $0.496\pm0.024$ & $0.495\pm0.017$\\
\addlinespace[3pt]
ETTh2 & 96 & $0.293\pm0.002$ & $0.342\pm0.002$ & $0.293\pm0.002$ & $0.342\pm0.002$\\
 & 192 & $0.375\pm0.005$ & $0.392\pm0.001$ & $0.375\pm0.005$ & $0.392\pm0.001$\\
 & 336 & $0.417\pm0.002$ & $0.427\pm0.001$ & $0.418\pm0.002$ & $0.428\pm0.001$\\
 & 720 & $0.443\pm0.001$ & $0.466\pm0.002$ & $0.421\pm0.001$ & $0.440\pm0.001$\\
\end{longtable}
\endgroup

\begingroup
\small
\setlength{\tabcolsep}{6pt}
\begin{longtable}{@{}llrrrr@{}}
\caption{VE on iTransformer: test MSE and MAE, mean $\pm$ sample SD over three seeds, with $L=96$.}\label{tab:ve-itransformer-full}\\
\toprule
 & & \multicolumn{2}{c}{Original} & \multicolumn{2}{c}{Expert-budget matched}\\
\cmidrule(lr){3-4}\cmidrule(lr){5-6}
Dataset & $H$ & MSE & MAE & MSE & MAE\\
\midrule
\endfirsthead
\multicolumn{6}{c}{\tablename\ \thetable\ continued}\\
\toprule
 & & \multicolumn{2}{c}{Original} & \multicolumn{2}{c}{Expert-budget matched}\\
\cmidrule(lr){3-4}\cmidrule(lr){5-6}
Dataset & $H$ & MSE & MAE & MSE & MAE\\
\midrule
\endhead
\midrule
\multicolumn{6}{r}{Continued on the next page}\\
\endfoot
\bottomrule
\endlastfoot
ETTh1 & 96 & $0.390\pm0.002$ & $0.408\pm0.002$ & $0.396\pm0.004$ & $0.415\pm0.003$\\
 & 192 & $0.443\pm0.004$ & $0.440\pm0.002$ & $0.454\pm0.006$ & $0.449\pm0.002$\\
 & 336 & $0.487\pm0.003$ & $0.461\pm0.001$ & $0.491\pm0.004$ & $0.466\pm0.001$\\
 & 720 & $0.505\pm0.004$ & $0.491\pm0.004$ & $0.504\pm0.006$ & $0.490\pm0.003$\\
\addlinespace[3pt]
ETTh2 & 96 & $0.299\pm0.002$ & $0.348\pm0.001$ & $0.302\pm0.004$ & $0.353\pm0.003$\\
 & 192 & $0.382\pm0.001$ & $0.400\pm0.001$ & $0.388\pm0.005$ & $0.405\pm0.003$\\
 & 336 & $0.426\pm0.009$ & $0.432\pm0.003$ & $0.427\pm0.006$ & $0.435\pm0.003$\\
 & 720 & $0.426\pm0.006$ & $0.443\pm0.003$ & $0.430\pm0.005$ & $0.447\pm0.003$\\
\end{longtable}
\endgroup

\begingroup
\small
\setlength{\tabcolsep}{6pt}
\begin{longtable}{@{}llrrrr@{}}
\caption{$\delta$-Adapter on iTransformer: test MSE and MAE, mean $\pm$ sample SD over three seeds, with $L=96$. The reduced-width variant uses the closest feasible adapter budget.}\label{tab:delta-itransformer-full}\\
\toprule
 & & \multicolumn{2}{c}{Original} & \multicolumn{2}{c}{Reduced-width}\\
\cmidrule(lr){3-4}\cmidrule(lr){5-6}
Dataset & $H$ & MSE & MAE & MSE & MAE\\
\midrule
\endfirsthead
\multicolumn{6}{c}{\tablename\ \thetable\ continued}\\
\toprule
 & & \multicolumn{2}{c}{Original} & \multicolumn{2}{c}{Reduced-width}\\
\cmidrule(lr){3-4}\cmidrule(lr){5-6}
Dataset & $H$ & MSE & MAE & MSE & MAE\\
\midrule
\endhead
\midrule
\multicolumn{6}{r}{Continued on the next page}\\
\endfoot
\bottomrule
\endlastfoot
ETTh1 & 96 & $0.394\pm0.001$ & $0.409\pm0.001$ & $0.394\pm0.002$ & $0.410\pm0.002$\\
 & 192 & $0.445\pm0.003$ & $0.445\pm0.002$ & $0.457\pm0.006$ & $0.450\pm0.005$\\
 & 336 & $0.494\pm0.005$ & $0.471\pm0.005$ & $0.498\pm0.003$ & $0.470\pm0.003$\\
 & 720 & $0.528\pm0.005$ & $0.515\pm0.003$ & $0.526\pm0.011$ & $0.508\pm0.004$\\
\addlinespace[3pt]
ETTh2 & 96 & $0.298\pm0.001$ & $0.349\pm0.000$ & $0.299\pm0.001$ & $0.349\pm0.000$\\
 & 192 & $0.380\pm0.001$ & $0.398\pm0.001$ & $0.380\pm0.001$ & $0.398\pm0.001$\\
 & 336 & $0.424\pm0.002$ & $0.432\pm0.001$ & $0.424\pm0.002$ & $0.432\pm0.001$\\
 & 720 & $0.454\pm0.004$ & $0.472\pm0.002$ & $0.437\pm0.003$ & $0.452\pm0.002$\\
\end{longtable}
\endgroup

\subsection{Fitting and Deployment Costs}
\label{app:pcr-cost}

\paragraph{What is trained and what is deployed?}
PCR gradient-trains $Er(F+H)$ expert parameters and analytically fits
$EC$ mixture weights and their supports. Learned allocation additionally
gradient-trains $EC$ logits. Deployment retains the frozen backbone and
head, experts, mixtures, and calibration; PCR also records its ridge
selection. Table~\ref{tab:pcr-resource-ledger} distinguishes trainable
parameters, stored model parameters, and forecast computation.

PCR variants have fewer nominal trainable parameters than every
corresponding external expert-based configuration. This does not imply the smallest
deployed model: matched CCM stores fewer parameters and has fewer
analytical FLOPs in all 88 DLinear/PatchTST settings; MoLE matched also has fewer FLOPs
in all 44 DLinear settings. K2 evaluates all four experts and has the
same FLOPs as K4. FLOP counts omit some framework operations and do
not substitute for measured latency.

\begin{table}[!htbp]\centering\small
\setlength{\tabcolsep}{5pt}
\caption{Median parameters and FLOPs per window over 44 settings.
Trainable counts include retained router weights with zero data gradient.
MoLE matched and its expert-budget control have identical counts.
PCR stores fitted mixtures/supports in addition to model parameters.
iTransformer FLOPs use a PyTorch counter; other rows are analytical
estimates, both omitting uncounted overhead. Bold/underline mark the
lowest/second distinct reported values per backbone and metric; ties share rank.}
\label{tab:pcr-resource-ledger}
\begin{tabular}{llrrr}
\toprule
Backbone & Method & Stored params. & Trainable params. & FLOPs (M) \\
\midrule
DLinear & PCR-K4/K2 & 80,400 & \textbf{29,184} & 21.8 \\
 & Learned-K4/K2 & 82,180 & \underline{30,964} & 21.8 \\
 & CCM original & 212,236 & 212,236 & 50.8 \\
 & CCM matched & \underline{36,992} & 36,992 & \underline{12.1} \\
 & MoLE original & 671,344 & 671,344 & 49.7 \\
 & MoLE matched / expert-budget control & 495,664 & 495,664 & \textbf{10.9} \\
 & VE original & 148,596 & 148,596 & --- \\
 & VE matched & \textbf{31,816} & 31,816 & --- \\
 & $\delta$-Adapter original & 55,855,072 & 55,827,136 & --- \\
 & $\delta$-Adapter reduced-width & 405,630 & 377,694 & --- \\
\addlinespace
PatchTST & PCR-K4/K2 & 805,440 & \textbf{110,592} & \underline{1,827.3} \\
 & Learned-K4/K2 & 806,696 & \underline{111,848} & \underline{1,827.3} \\
 & CCM original & 1,593,204 & 1,593,204 & 2,206.1 \\
 & CCM matched & \underline{523,380} & 523,380 & \textbf{1,693.8} \\
 & VE original & 694,302 & 694,302 & --- \\
 & VE matched & \textbf{511,656} & 511,656 & --- \\
 & $\delta$-Adapter original & 56,448,208 & 55,827,136 & --- \\
 & $\delta$-Adapter reduced-width & 1,175,694 & 377,694 & --- \\
\addlinespace
iTransformer & PCR-K4/K2 & \underline{4,872,800} & \textbf{45,056} & \underline{1,909.1} \\
 & Learned-K4/K2 & 4,876,290 & \underline{46,414} & \underline{1,909.1} \\
 & CCM original & 5,031,296 & 5,031,296 & 2,046.4 \\
 & CCM matched & 4,873,216 & 4,873,216 & \textbf{1,879.5} \\
 & VE original & 4,986,500 & 4,986,500 & --- \\
 & VE matched & \textbf{4,827,106} & 4,827,106 & --- \\
 & $\delta$-Adapter original & 60,685,648 & 55,827,136 & --- \\
 & $\delta$-Adapter reduced-width & 5,255,934 & 377,694 & --- \\
\bottomrule
\end{tabular}
\end{table}
\begin{figure}[!htbp]
\centering
\includegraphics[width=\textwidth]{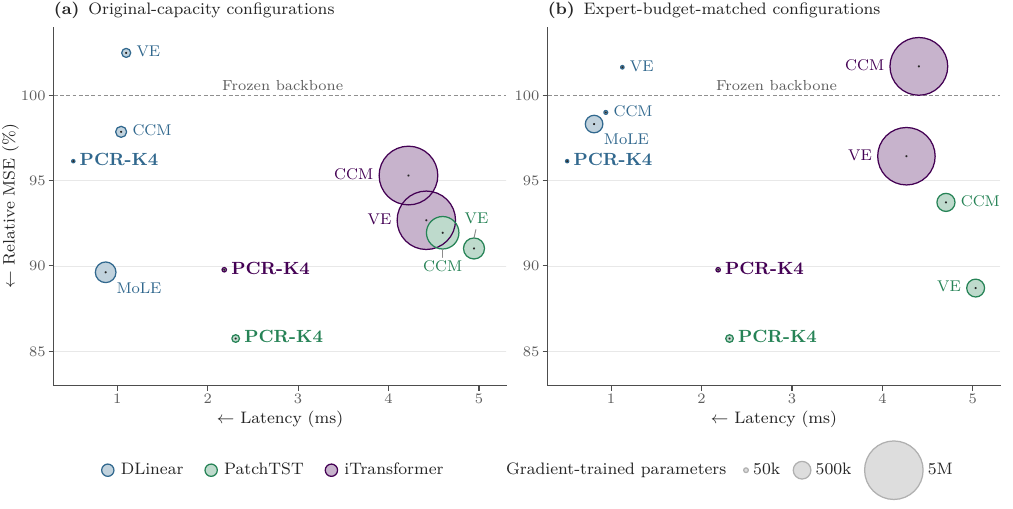}
\caption{\textbf{Original and expert-budget-matched specialization.}
PCR-K4 is repeated in both panels. Relative MSE is dataset-balanced
across eleven datasets and four horizons, averaged over three seeds;
latency and gradient-trained parameter counts are medians over the
44 settings per backbone. Colours identify backbones and bubble area
is proportional to parameter count, with a common scale across panels.
The dashed line denotes the frozen reference. Expert-budget matching
does not match total trainable parameters; MoLE matched also removes
calendar inputs. Lower MSE and latency are better.}
\label{fig:pcr-accuracy-latency-full}
\end{figure}

\paragraph{Fitting time and memory.}
Fitting time includes feature-cache construction where used, expert
training, allocation statistics and solves, and validation selection,
but excludes backbone pretraining. Table~\ref{tab:pcr-full-cost}
also records time excluding reusable cache setup and peak GPU memory.
PCR's caches reduce repeated backbone evaluation at the cost of memory.
MoLE and CCM fit the full model; their non-selected tuning trials
are summed once per setting. MoLE matched inherits the expert-budget
control's hyperparameters and is not charged a second search.
Complete search costs for reused backbones are unavailable.
Different training GPUs and concurrent workloads prevent interpreting
these observed times as controlled speed ratios.

Across six backbone-sparsity configurations, median PCR mixture-solving
time is 48.9-107.7 seconds and statistics accumulation is 1.3-16.0
seconds. Component medians need not sum to the median total; evaluating
many training checkpoints is distinct from deployment latency.

\begin{table}[!htbp]\centering\small
\setlength{\tabcolsep}{4pt}
\caption{Median fitting costs. PCR/learned rows summarize 132 fits per
method; external rows give selected joint-fit time and non-selected tuning
time per setting. Memory is median / maximum, retaining recorded units
(PCR/learned: GB; external: GiB). Dashes mark unreported quantities or
inherited tuning for MoLE matched. CCM-iTransformer fitting medians use all 132 selected trajectories per variant; tuning sums non-selected trials per setting. Training hardware and concurrency differ.}
\label{tab:pcr-full-cost}\label{tab:pcr-external-cost}\label{tab:pcr-efficiency}
\begin{tabular}{llrrrr}
\toprule
Backbone & Method & Fit (s) & \shortstack{Excl. cache\\setup (s)} & Tuning (s) & Peak memory \\
\midrule
DLinear & PCR-K4 & 562.9 & 560.0 & - & 0.54 / 5.72 \\
 & PCR-K2 & 546.3 & 538.0 & - & 0.54 / 5.72 \\
 & Learned-K4 & 527.6 & 527.6 & - & 0.08 / 0.97 \\
 & Learned-K2 & 639.6 & 639.6 & - & 0.08 / 0.97 \\
PatchTST & PCR-K4 & 1,101.9 & 1,068.6 & - & 1.95 / 17.65 \\
 & PCR-K2 & 968.2 & 929.5 & - & 1.95 / 17.65 \\
 & Learned-K4 & 4,567.9 & 4,567.9 & - & 0.42 / 2.57 \\
 & Learned-K2 & 3,911.3 & 3,911.3 & - & 0.42 / 2.57 \\
iTransformer & PCR-K4 & 701.7 & 682.0 & - & 6.13 / 17.12 \\
 & PCR-K2 & 672.5 & 656.1 & - & 6.13 / 17.12 \\
 & Learned-K4 & 1,488.4 & 1,488.4 & - & 0.24 / 3.32 \\
 & Learned-K2 & 1,730.0 & 1,730.0 & - & 0.24 / 3.32 \\
\midrule
DLinear & CCM original & 391.6 & - & 1,169.0 & 0.164 / 1.715 \\
DLinear & CCM matched & 325.3 & - & 1,062.4 & 0.120 / 0.905 \\
DLinear & MoLE original & 451.6 & - & 2,535.6 & 0.565 / 3.428 \\
DLinear & MoLE matched & 412.9 & - & - & 0.295 / 0.645 \\
DLinear & MoLE expert-budget control & 315.5 & - & 2,098.9 & 0.308 / 2.892 \\
PatchTST & CCM original & 2,426.6 & - & 8,801.3 & 2.452 / 10.242 \\
PatchTST & CCM matched & 2,249.8 & - & 4,364.5 & 2.430 / 10.320 \\
iTransformer & CCM original & 1,280.1 & - & 3,453.2 & - \\
iTransformer & CCM matched & 826.6 & - & 2,275.7 & - \\
\bottomrule
\end{tabular}
\end{table}

Table~\ref{tab:pcr-common-cost} adds selected backbone pretraining once
per PatchTST setting. It takes the median of three-seed setting means,
rather than the fit-level medians above; peak memory excludes separate
pretraining. Backbone search takes 85.67 fitting hours over 352 candidates,
and calibration 12.59 hours over 88 calibrations. These totals exclude
earlier training of reused checkpoints and are not repeated per variant.

\begingroup\small
\setlength{\tabcolsep}{3pt}
\begin{table}[!htbp]\centering
\caption{PatchTST costs with and without selected pretraining:
medians of three-seed means over 44 settings. Time is in seconds and
peak memory in GiB. Pretraining is included once; external joint fits
already train the full model. Search is reported separately.}\label{tab:pcr-common-cost}
\begin{tabular}{lrrr}
\toprule
Method & Adaptation / joint fit & Including base & Peak GiB \\
\midrule
PCR-K4 & 976.7 & 1,978.8 & 1.948 \\
PCR-K2 & 927.5 & 1,906.0 & 1.948 \\
Learned-K4 & 4,570.2 & 5,792.8 & 0.424 \\
Learned-K2 & 4,014.1 & 4,998.9 & 0.424 \\
CCM original & 2,586.9 & 2,586.9 & 2.452 \\
CCM matched & 2,361.3 & 2,361.3 & 2.430 \\
\bottomrule
\end{tabular}
\end{table}
\endgroup
\paragraph{Deployment measurements.}
We profile seed-zero selected checkpoints for the backbone and four
PCR/learned methods in all 132 settings (660 configurations), plus
396 MoLE/CCM configurations: 352 main comparisons and 44 MoLE
expert-budget controls. Timing includes the complete forecast, correction,
calibration, and inverse normalization, without training caches.
All use FP32, no gradients, GPU-resident windows, synchronized timing,
and exclusive RTX 2080 Ti devices. Batch-one latency uses 30 warm-up
and 300 timed windows; batch-32 throughput uses ten warm-up and 100
timed forwards. These repetitions are timing samples, not training seeds.
Full-test evaluation uses batch 32 for PCR/learned methods and recorded
evaluation batches for external methods.

CCM computes memberships from the current window at batch one and the
current batch at batch 32. All 44 MoLE matched checkpoints reproduce
stored validation errors before timing; their routers receive zero inputs.
Median latency is 0.812 ms and throughput 37,839 windows/s.

PCR-K4 has lower batch-one latency in all 396 MoLE/CCM comparisons.
Across 44 settings, latency ranges are 0.468-0.656 ms for DLinear,
2.097-5.608 ms for PatchTST, and 1.514-3.023 ms for iTransformer
(Table~\ref{tab:pcr-inference-summary}). Median added latencies are
0.191/0.180/0.188 ms; median setting-wise overheads are
58.9\%/8.1\%/9.8\%, and throughput ratios to the backbone are
0.733/0.974/0.967, in the same backbone order.

PCR has the highest aggregate DLinear throughput, but matched competitors
win individual settings. PatchTST CCM original/matched medians are
1,120/1,177 windows/s, versus 1,071 for PCR-K4. Aggregate medians and
medians of setting-wise ratios answer different questions.
All four PCR/learned methods use the same static-mixture computation at
deployment; small timing differences do not establish a K2 speedup.

\par\begingroup
On iTransformer, original/matched CCM have median batch-one latencies
of 4.220/4.403 ms, versus 2.183 ms for PCR-K4. Median setting-wise
CCM/PCR-K4 ratios are 1.97/2.05. Batch-32 throughputs are 3,141/3,413
windows/s versus 3,433 for PCR-K4, so the latency advantage does not
imply a comparable throughput advantage. All 88 CCM-iTransformer
profiles use the same host, GPU model, and software as the PCR profiles.
\par\endgroup

Incremental inference memory subtracts memory already resident before
measurement; model state includes deployed parameters, buffers, mixtures,
and calibration, not serialization overhead. The evaluation covers 1,056 PCR/learned and MoLE/CCM profiles and
associated memory, state, FLOP, and full-test measurements.

VE and $\delta$-Adapter add 528 seed-zero profiles, covering both
configurations on all 132 settings under the same batch-one timing
protocol. PCR-K4 has lower median latency than each configuration on
each backbone. These profiles report latency and parameter counts;
batch-32 throughput and FLOPs are not reported for these methods.

\begingroup\small
\setlength{\tabcolsep}{3pt}
\begin{table}[!htbp]\centering
\caption{Deployment medians over 44 settings per backbone. Latency: batch-one ms; throughput: batch-32 windows/s; peak/incremental memory: batch-32 GiB. State includes deployed parameters, buffers, gates, and calibration; FLOPs are per window. All profiles use seed-zero selected checkpoints. External rows show the available latency and throughput measurements; dashes denote quantities not reported here.}\label{tab:pcr-inference-summary}
\begin{tabular}{llrrrrr}
\toprule
Backbone & Configuration & Latency & Throughput & Peak / incremental & FLOPs (M) & State (MiB) \\
\midrule
DLinear & Backbone & 0.325 & 65,559 & 0.108 / 0.017 & 11.6 & 0.195 \\
DLinear & PCR-K4 & 0.514 & 59,052 & 0.111 / 0.021 & 21.8 & 0.313 \\
DLinear & PCR-K2 & 0.510 & 58,791 & 0.121 / 0.021 & 21.8 & 0.313 \\
DLinear & Learned-K4 & 0.511 & 58,536 & 0.121 / 0.021 & 21.8 & 0.313 \\
DLinear & Learned-K2 & 0.508 & 59,597 & 0.121 / 0.021 & 21.8 & 0.313 \\
PatchTST & Backbone & 2.138 & 1,086 & 0.419 / 0.316 & 1,788.1 & 5.093 \\
PatchTST & PCR-K4 & 2.310 & 1,071 & 0.416 / 0.316 & 1,827.3 & 5.520 \\
PatchTST & PCR-K2 & 2.304 & 1,069 & 0.454 / 0.316 & 1,827.3 & 5.520 \\
PatchTST & Learned-K4 & 2.305 & 1,065 & 0.454 / 0.316 & 1,827.3 & 5.520 \\
PatchTST & Learned-K2 & 2.304 & 1,063 & 0.454 / 0.316 & 1,827.3 & 5.520 \\
iTransformer & Backbone & 2.002 & 3,499 & 0.259 / 0.144 & 1,892.2 & 18.440 \\
iTransformer & PCR-K4 & 2.183 & 3,433 & 0.257 / 0.143 & 1,909.1 & 18.602 \\
iTransformer & PCR-K2 & 2.181 & 3,422 & 0.273 / 0.143 & 1,909.1 & 18.602 \\
iTransformer & Learned-K4 & 2.186 & 3,414 & 0.273 / 0.143 & 1,909.1 & 18.602 \\
iTransformer & Learned-K2 & 2.188 & 3,406 & 0.273 / 0.143 & 1,909.1 & 18.602 \\
\midrule
DLinear & CCM original & 1.043 & 23,493 & - & - & - \\
DLinear & CCM matched & 0.941 & 32,532 & - & - & - \\
DLinear & MoLE original & 0.872 & 25,957 & - & - & - \\
DLinear & MoLE matched & 0.812 & 37,839 & - & - & - \\
PatchTST & CCM original & 4.599 & 1,120 & - & - & - \\
PatchTST & CCM matched & 4.703 & 1,177 & - & - & - \\
iTransformer & CCM original & 4.220 & 3,141 & - & - & - \\
iTransformer & CCM matched & 4.403 & 3,413 & - & - & - \\
\midrule
DLinear & VE original & 1.099 & - & - & - & - \\
DLinear & VE matched & 1.124 & - & - & - & - \\
DLinear & $\delta$-Adapter original & 1.891 & - & - & - & - \\
DLinear & $\delta$-Adapter reduced-width & 1.756 & - & - & - & - \\
PatchTST & VE original & 4.945 & - & - & - & - \\
PatchTST & VE matched & 5.031 & - & - & - & - \\
PatchTST & $\delta$-Adapter original & 5.467 & - & - & - & - \\
PatchTST & $\delta$-Adapter reduced-width & 5.251 & - & - & - & - \\
iTransformer & VE original & 4.418 & - & - & - & - \\
iTransformer & VE matched & 4.265 & - & - & - & - \\
iTransformer & $\delta$-Adapter original & 5.263 & - & - & - & - \\
iTransformer & $\delta$-Adapter reduced-width & 4.967 & - & - & - & - \\
\bottomrule
\end{tabular}
\end{table}
\endgroup
\subsubsection{Measurement Hardware}
\label{app:pcr-hardware}

PCR/learned training uses NVIDIA RTX 2080 Ti and RTX 3090 GPUs;
competitor and calendar-control training uses RTX 2080 Ti, with CCM-iTransformer also using RTX 3090.
Training may share devices. Controlled adaptation and deployment
measurements use exclusive RTX 2080 Ti devices, though PCR and competitor
profiles for the earlier DLinear/PatchTST comparisons run on different hosts; CCM-iTransformer uses the same host as PCR. Input-dependent lookback profiling uses
an exclusive RTX 3090. Deployment profiles use FP32, PyTorch 2.4.1+cu121,
and CUDA 12.1.

\clearpage
\section{Supporting Deployment Results and Derivations}
\label{app:method-benchmark}\label{app:residuals}

This appendix complements the allocation and lookback comparisons:
deployment costs for Appendix~\ref{app:routing}, additional lookback
profiles for Appendix~\ref{app:lookback-protocol}, and a supporting
low-rank ridge identity. PCR's costs are reported in
Appendix~\ref{app:pcr-cost}.

\subsection{Deployment Costs of the Allocation Study}
\label{app:resource-costs}

Table~\ref{tab:resource-all} compares corrections as fitted with
equivalent static deployment. Parameter counts cover 22 settings per
backbone, excluding calibration and fixed assignments.
Deployment covers 22 DLinear, 22 iTransformer, and 20 PatchTST
settings; PatchTST PEMS07 and Traffic at $H=96$ are excluded.
Ratios are computed per setting before taking medians.
Persistent ID and Dense ID store $4C$ logits; Identity + input stores
$4C+4F$ weights. The corresponding median router-only shares of backbone
parameters are $0.683/3.676\%$ for DLinear, $0.014/0.089\%$ for
iTransformer, and $0.081/0.885\%$ for PatchTST.

Seed-zero validation-selected checkpoints are timed on an exclusive
RTX 2080 Ti using FP32, evaluation mode, no gradients, and batch 32.
Ten warm-up forwards precede 50 synchronized calls on GPU-resident windows.
Timing includes the backbone, heads, correction, and inverse normalization,
not data transfer. Activation memory is peak allocation above
pre-loop resident memory, in decimal MB; it excludes resident model
storage and is distinct from the training memory in
Appendix~\ref{app:adaptation-resources}.

Top-two routing evaluates all expert input projections, without sparse
dispatch. Static deployment merges heads when their estimated
storage is below 1~GB; otherwise, it retains fixed low-rank factors. Maximum relative output differences from the unmerged predictor
are $1.27\times10^{-6}$, $2.68\times10^{-6}$, and $3.03\times10^{-6}$
for DLinear, iTransformer, and PatchTST. A merged Global correction
discards its fitting factors and adds no deployed weights.

\begin{table}[!htb]
\centering\small
\setlength{\tabcolsep}{6pt}
\caption{Median allocation-study deployment costs. Added parameters are
in thousands; latency is batch-32 ms; overhead is the setting-wise
increase over the backbone (\%); activation memory is incremental MB.
Static counts exclude discarded fitting factors and routers.
Dashes denote unreported memory.}
\label{tab:resource-all}\label{tab:compiled-resources}
\begin{tabular}{lrrrr}
\toprule
Correction / deployment & Added (k) & Latency (ms) & Overhead (\%) & Activation (MB) \\
\midrule
\multicolumn{5}{l}{\textit{DLinear}} \\
Global (unmerged) & 38.40 & 0.56 & 25.94 & 17.65 \\
Random (unmerged) & 38.40 & 0.58 & 29.40 & 17.65 \\
Persistent ID (unmerged) & 40.18 & 0.84 & 83.91 & 17.74 \\
Identity + input (unmerged) & 40.95 & 0.92 & 98.43 & 17.74 \\
Dense ID (unmerged) & 40.18 & 0.58 & 30.41 & 17.65 \\
Global (merged head) & 0.00 & 0.36 & -27.63 & - \\
Persistent ID (static factors) & 58.82 & 0.43 & -6.78 & - \\
Dense ID (static factors) & 58.82 & 0.43 & -6.13 & - \\
\midrule
\multicolumn{5}{l}{\textit{iTransformer}} \\
Global (unmerged) & 46.59 & 7.77 & 3.00 & 74.22 \\
Random (unmerged) & 46.59 & 7.84 & 4.53 & 74.86 \\
Persistent ID (unmerged) & 48.37 & 7.96 & 6.45 & 74.26 \\
Identity + input (unmerged) & 49.65 & 8.03 & 7.67 & 74.79 \\
Dense ID (unmerged) & 48.37 & 7.91 & 5.10 & 74.79 \\
Global (merged head) & 0.00 & 7.54 & 0.29 & - \\
Persistent ID (static factors) & 60.64 & 7.84 & 3.63 & - \\
Dense ID (static factors) & 60.64 & 7.87 & 4.47 & - \\
\midrule
\multicolumn{5}{l}{\textit{PatchTST}} \\
Global (unmerged) & 104.45 & 16.95 & 2.19 & 190.92 \\
Random (unmerged) & 104.45 & 17.08 & 3.05 & 190.89 \\
Persistent ID (unmerged) & 106.89 & 17.26 & 4.00 & 190.95 \\
Identity + input (unmerged) & 113.03 & 17.26 & 3.92 & 190.95 \\
Dense ID (unmerged) & 106.89 & 17.14 & 3.20 & 190.89 \\
Global (merged head) & 0.00 & 16.88 & 1.33 & - \\
Persistent ID (static factors) & 126.18 & 17.28 & 3.08 & - \\
Dense ID (static factors) & 126.18 & 17.27 & 2.90 & - \\

\bottomrule
\end{tabular}
\end{table}

\subsection{Lookback Deployment Costs}
\label{app:lookback-cost}
Table~\ref{tab:input-lookback-resources} gives 72 iTransformer
profiles across twelve settings. Frozen backbone and Input-dependent
timings use trained seed-zero models; Global, Persistent ID,
Identity + input, and Dense ID are untrained architecture probes with
nonzero initial weights. Their timings describe execution costs,
not forecasting accuracy.

The batch-32 procedure above is repeated on an exclusive RTX 3090,
including calibration. Input-dependent corrections add 40,964 parameters
at $H=96$ and 80,900 at $H=720$, excluding the frozen backbone,
calibration, and fixed labels. Activation memory here uses MiB
($2^{20}$ bytes), rather than decimal MB.

\begin{table}[!htb]
\centering\small
\setlength{\tabcolsep}{5pt}
\caption{iTransformer lookback profiles: median batch-32 latency (ms) over 50 synchronized forwards. Base and Input use trained checkpoints; the four remaining columns are untrained architecture probes. ID denotes persistent routing.}
\label{tab:input-lookback-resources}
\begin{tabular}{lrrrrrrrr}
\toprule
Dataset & $L$ & $H$ & Base & Input & Global & ID & ID + input & Dense ID \\
\midrule
Electricity & 96 & 96 & 13.275 & 13.277 & 13.012 & 13.121 & 13.131 & 12.865 \\
Electricity & 96 & 720 & 13.428 & 13.729 & 13.474 & 13.691 & 13.700 & 13.443 \\
Electricity & 720 & 96 & 13.812 & 13.836 & 13.540 & 13.699 & 13.747 & 13.503 \\
Electricity & 720 & 720 & 14.460 & 14.637 & 14.271 & 14.450 & 14.530 & 14.264 \\
PEMS04 & 96 & 96 & 12.981 & 13.194 & 12.895 & 12.900 & 12.868 & 12.659 \\
PEMS04 & 96 & 720 & 12.591 & 13.198 & 12.885 & 13.108 & 13.170 & 12.921 \\
PEMS04 & 720 & 96 & 13.552 & 13.550 & 13.368 & 13.285 & 13.266 & 13.026 \\
PEMS04 & 720 & 720 & 13.318 & 13.761 & 13.462 & 13.540 & 13.763 & 13.461 \\
Traffic & 96 & 96 & 46.082 & 47.323 & 46.938 & 47.448 & 47.533 & 47.655 \\
Traffic & 96 & 720 & 46.463 & 47.875 & 47.531 & 47.885 & 48.172 & 47.769 \\
Traffic & 720 & 96 & 46.916 & 48.190 & 47.898 & 48.258 & 48.657 & 48.273 \\
Traffic & 720 & 720 & 47.776 & 49.212 & 48.783 & 49.173 & 49.476 & 49.123 \\

\bottomrule
\end{tabular}
\end{table}

\subsection{Reduced-Rank Ridge Identity and Calibration}
\label{app:solver}

For fixed feature and residual matrices $Z,R$ in evaluation coordinates,
consider $\min_{\operatorname{rank}(W)\le r}
\|R-ZW\|_F^2+\lambda\|W\|_F^2$ with $\lambda>0$ and no intercept.
Let $A=Z^\top Z+\lambda I$ and $T=A^{-1/2}Z^\top R$. Completing the
square gives
\begin{equation}
 \|R-ZW\|_F^2+\lambda\|W\|_F^2
 =\|A^{1/2}W-T\|_F^2+\|R\|_F^2-\|T\|_F^2.
\end{equation}
Because $A^{1/2}$ is invertible, multiplication preserves rank.
The best rank-$r$ approximation $[T]_r$ therefore yields
\begin{equation}
 W^\star=A^{-1/2}[A^{-1/2}Z^\top R]_r.
 \label{eq:rrr-solution}
\end{equation}
This is a supporting direct-fitting identity, not the PCR allocation
estimator or a reported empirical comparison.

For a fixed correction, expanding the squared validation residuals
gives Equation~\eqref{eq:global-calibration}. Selecting the minimum over
coefficients that include zero cannot worsen validation MSE relative to
the uncorrected prediction; this gives no test-error guarantee.

\end{document}